\pdfoutput=1

\documentclass[11pt]{article}

\usepackage[]{ACL2023}
\usepackage{times}
\usepackage{latexsym}

\usepackage[T1]{fontenc}

\usepackage[utf8]{inputenc}

\usepackage{microtype}

\usepackage{inconsolata}

\usepackage{enumitem,graphicx,xspace,inconsolata,subcaption}

\newcommand{\idspecificfeature}[0]{ID-specific pattern\xspace}
\newcommand{\idspecificfeatures}[0]{ID-specific patterns\xspace}

\newcommand{\bertlarge}[0]{BERT\textsubscript{LARGE}\xspace}
\newcommand{\bertbase}[0]{BERT\textsubscript{BASE}\xspace}

\newcommand{\mnli}[0]{MultiNLI\xspace}
\newcommand{\snli}[0]{SNLI\xspace}

\newif\ifcomments
\commentstrue

\ifcomments
    \providecommand{\nfl}[1]{{\protect\color{Green}{[NFL: #1]}}}
    \providecommand{\ak}[1]{{\protect\color{Magenta}{[AK: #1]}}}
    \providecommand{\rj}[1]{{\protect\color{ProcessBlue}{[RJ: #1]}}}
    \providecommand{\pl}[1]{{\protect\color{Red}{[PL: #1]}}}
\else
    \providecommand{\nfl}[1]{}
    \providecommand{\ak}[1]{}
    \providecommand{\rj}[1]{}
    \providecommand{\pl}[1]{}
\fi

\title{Are Sample-Efficient NLP Models More Robust?}

\newcommand{\emldisplay}[2]{\texttt{\href{mailto:#1}{#2}}}
\newcommand{\eml}[1]{\emldisplay{#1}{#1}}

\author{
    Nelson F. Liu$^{\spadesuit}$ \quad
	Ananya Kumar$^{\spadesuit}$ \quad
	{\bf Percy Liang}$^{\spadesuit}$ \quad
  {\bf Robin Jia}$^{\heartsuit}$\\
    $^\spadesuit$Computer Science Department, Stanford University, Stanford, CA\\
	$^\heartsuit$Department of Computer Science, University of Southern California, Los Angeles, CA\\
	\{\emldisplay{nfliu@cs.stanford.edu}{nfliu}, \emldisplay{ananya@cs.stanford.edu}{ananya}, \emldisplay{pliang@cs.stanford.edu}{pliang}\}\texttt{@cs.stanford.edu}\\
  \eml{robinjia@usc.edu}}

\begin{document}
\maketitle
\begin{abstract}
  Recent results in image classification and extractive question answering have observed that pre-trained models trained on less in-distribution data have better out-of-distribution performance.
  However, it is unclear how broadly these trends hold.
  We conduct a large empirical study across three tasks, three broadly-applicable modeling interventions (increasing model size, using a different adaptation method, and pre-training on more data), and 14 diverse datasets to investigate the relationship between sample efficiency (amount of data needed to reach a given ID accuracy) and robustness (how models fare on OOD evaluation).
  We find that higher sample efficiency is only correlated with better average OOD robustness on some modeling interventions and tasks, but not others.
  On individual datasets, models with lower sample efficiency can even be \emph{more} robust.
  These results suggest that general-purpose methods for improving sample efficiency are unlikely to yield universal OOD robustness improvements, since such improvements are highly dataset- and task-dependent. Even in an era of large, multi-purpose pre-trained models, task-specific decisions may often be necessary for OOD generalization.
\end{abstract}

\section{Introduction}

NLP models perform well when evaluated on data drawn from their training distribution (in-distribution / ID), but they typically suffer large drops in performance when evaluated on data distributions unseen during training (out-of-distribution / OOD; \citealp{blitzer2008domain}).

How does exposure to ID training examples affect the ID-OOD gap? If two models have the same ID performance, will models trained on fewer ID examples (higher \emph{sample efficiency}) also have higher OOD performance (higher \emph{robustness})?
At one extreme, zero-shot models will not learn \idspecificfeatures because they are not exposed to \emph{any} labeled ID examples. Similarly, few-shot models trained on very few ID examples may also rely less on \idspecificfeatures; if a model never sees the token \emph{``cat''} while training on \snli, then it will not learn that its presence is spuriously predictive of the contradiction label \citep{gururangan-etal-2018-annotation,utama2021avoiding}. Supporting this intuition, recent work in image classification \citep{radford2021learning} and extractive question answering \citep{awadalla2022exploring} show that zero-shot inference and few-shot fine-tuning improve \emph{average} robustness across a range of OOD test sets.
However, it is unclear how universal these trends are across various tasks and methods for reducing exposure to ID examples, or how predictive they are for any individual test set of interest. Figure~\ref{fig:example_plot} illustrates this central question.

We conduct a broad empirical study over 14 datasets across three tasks to investigate the relationship between exposure to ID training examples (sample efficiency) and robustness. We experiment with three modeling interventions that improve sample efficiency: (1)~using natural language prompts for zero-shot prediction and during fine-tuning \citep{NEURIPS2020_1457c0d6,schick-schutze-2021-exploiting,gao2021making}; (2)~fine-tuning models of increasing size; (3)~fine-tuning models pre-trained on increasing amounts of data.

We find that higher sample efficiency is only sometimes correlated with better robustness, and the effect of specific modeling interventions varies by task.
For example, increasing pre-trained model size substantially improves sample efficiency and results in higher average robustness in sentiment experiments, but these sample efficiency gains do not translate to higher average robustness in NLI and extractive QA experiments. On individual datasets, models with better sample efficiency can even be \emph{less} robust (e.g., increasing model size when training on SST-2 and evaluating OOD on IMDb).

Overall, these results indicate that general-purpose methods for improving sample efficiency are far from guaranteed to yield significant OOD robustness improvements---their success is highly dataset- and task-dependent. Furthermore, even in this era of large, multi-purpose pre-trained language models, task-specific decisions are often necessary to achieve OOD generalization.

\section{Measuring Sample Efficiency and Robustness.}

Consider two data distributions $\mathcal{D}_{iid}$ and $\mathcal{D}_{ood}$. Let $M$ be a model trained on examples drawn from $\mathcal{D}_{iid}$ (i.e., the ID training data). We study the relationship between three properties of $M$: (1)~the number of ID examples it was trained on; (2)~$M$'s performance on held-out examples from $\mathcal{D}_{iid}$ (i.e., the ID performance); (3)~$M$'s performance on examples from $\mathcal{D}_{ood}$ (i.e., the OOD performance).

\begin{figure}[t]
  \centering
  \includegraphics[width=0.8\columnwidth]{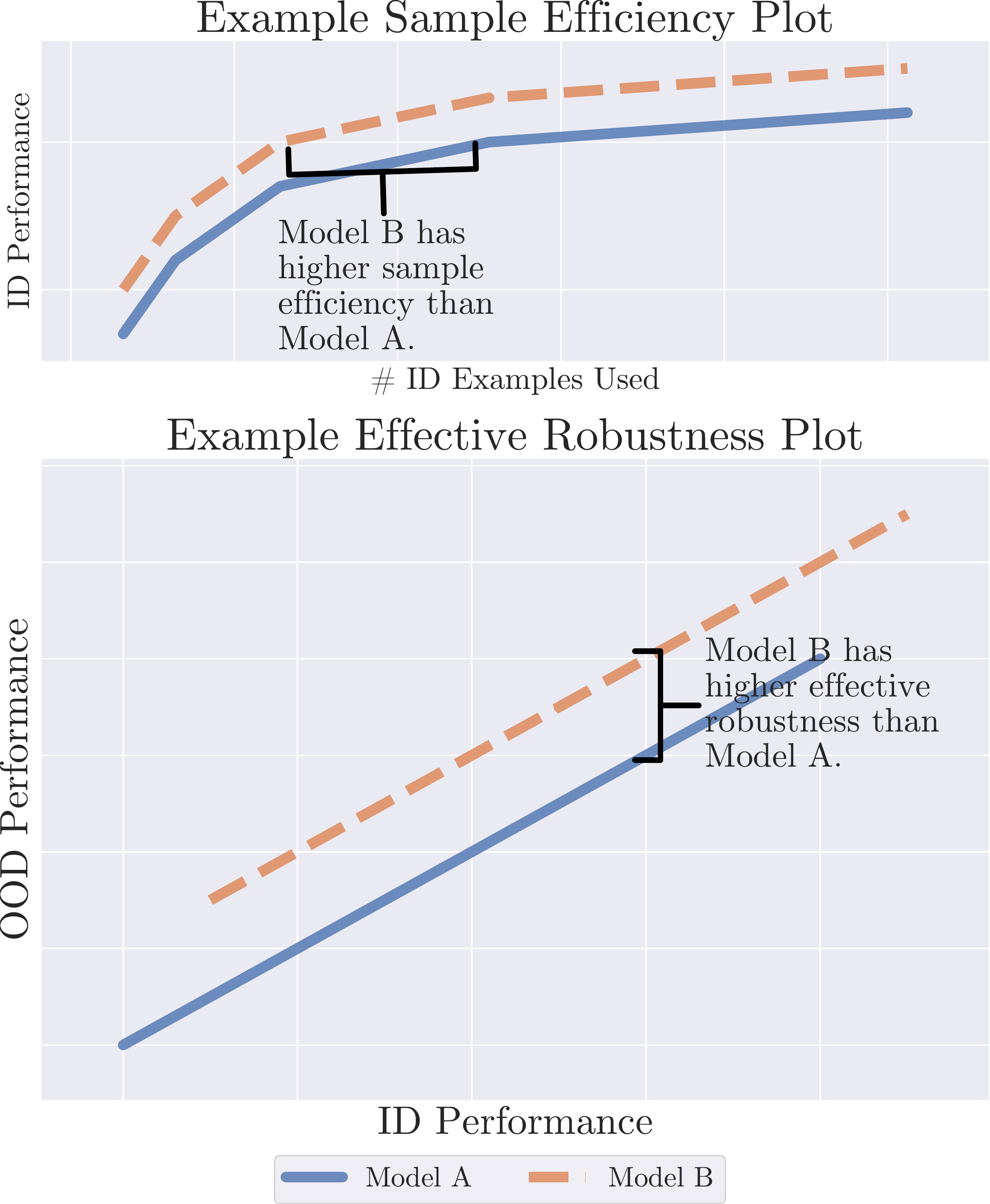}
  \caption{In this example, model B has higher sample efficiency than model A, since model B requires less ID training data to reach a given ID performance threshold (top). In this particular example, model B is also more robust than model A (bottom), since it has higher OOD performance for a given ID performance threshold.
  }\label{fig:example_plot}
\end{figure}

Let $M_1$ and $M_2$ be two models with equivalent performance on held-out ID data. If $M_1$ was trained on fewer ID examples than $M_2$, then it has higher \emph{sample efficiency}.
If $M_1$ has higher OOD performance than $M_2$, it has higher \emph{effective robustness} (henceforth ``robustness''; \citealp{Taori2020MeasuringRT}). Comparing models with equivalent ID performance controls for its effect on OOD performance, since improving ID performance usually yields commensurate improvements on OOD performance---in this study, we focus on OOD performance improvements \emph{beyond what is expected} from ID gains.

Satisfying this equivalent-ID constraint is often difficult in practice; given an arbitrary model $M_1$ and its corresponding ID performance, it is difficult to produce a different model $M_2$ with identical ID performance.
Rather than explicitly training models to identical ID performance, we train models on varying-size subsamples of a given ID dataset and interpolate between the results to estimate (1)~the number of labeled ID training examples necessary to achieve a particular ID performance (sample efficiency) and (2)~OOD performance, given ID performance (robustness).
These interpolated curves approximate the ideal setting of training a model for every possible ID value.
Figure~\ref{fig:example_plot} provides a schematized example, with model $B$ having better sample efficiency and robustness than model $A$.

\begin{figure*}[t]
  \centering
        \begin{subfigure}{0.3\linewidth}
            \centering \includegraphics[width=\linewidth]{{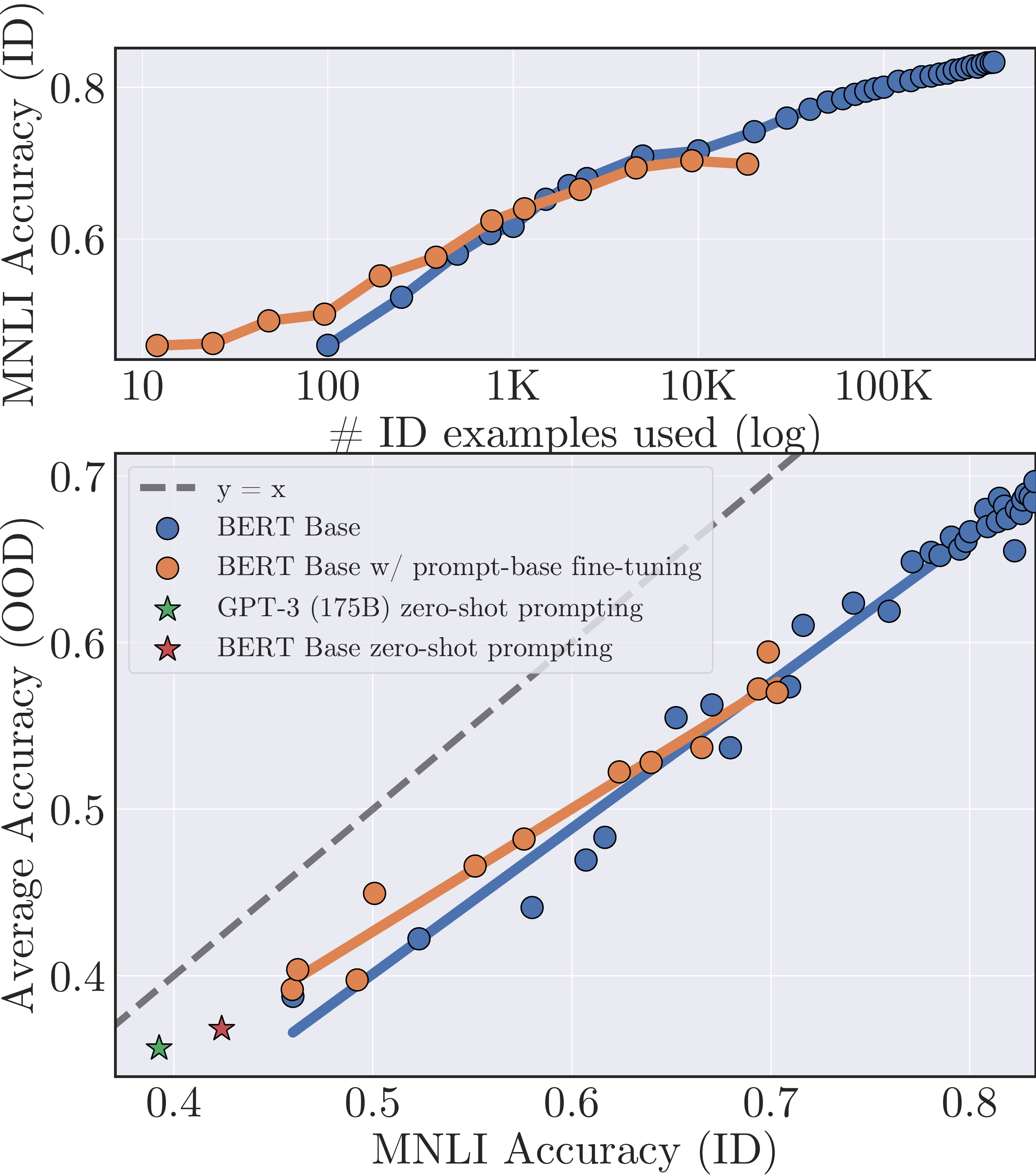}}
            \caption{}\label{fig:finetuning_method_mnli}
          \end{subfigure}
          \hfill
        \begin{subfigure}{0.3\linewidth}
            \centering \includegraphics[width=\linewidth]{{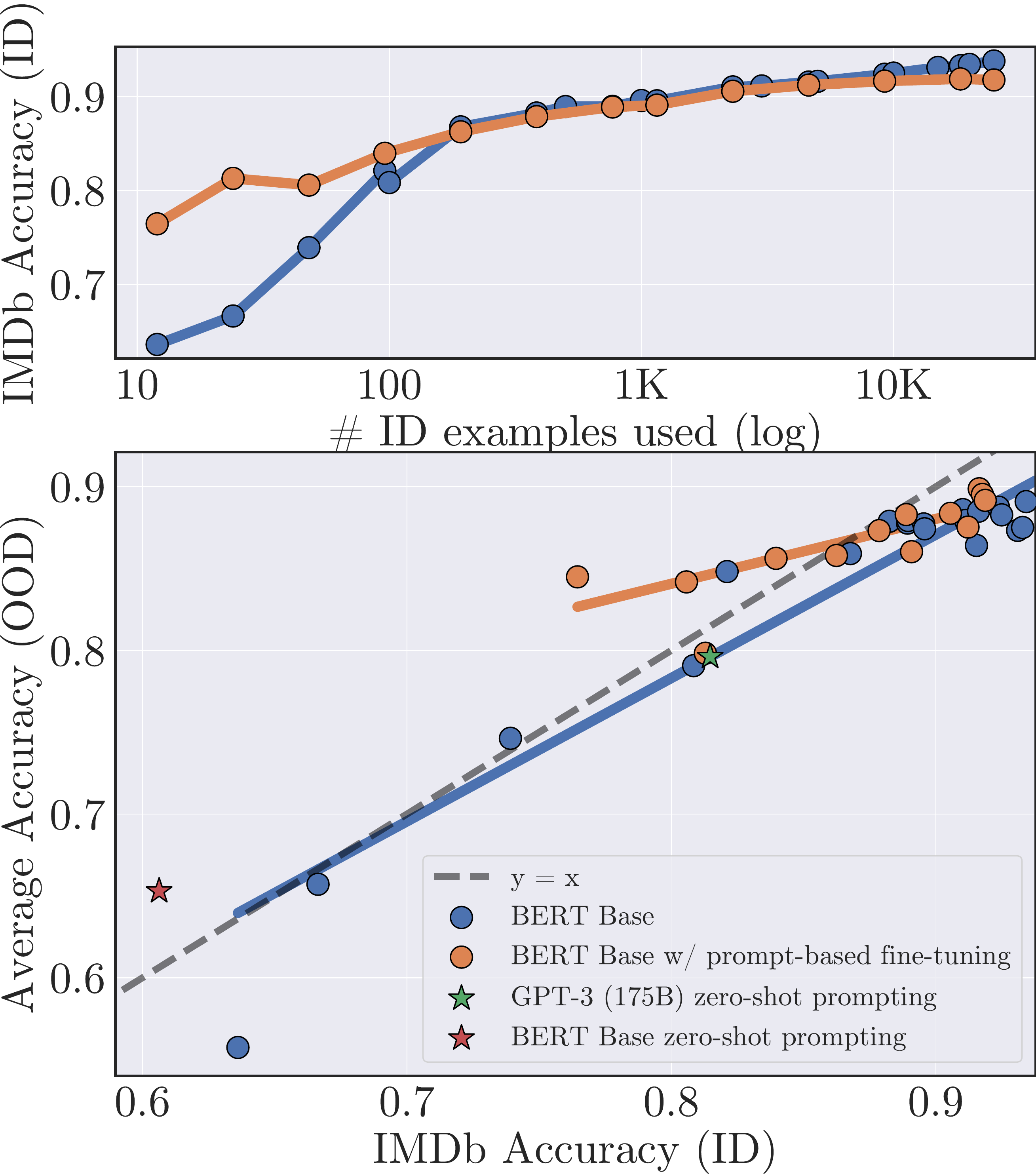}}
            \caption{}\label{fig:finetuning_method_imdb}
          \end{subfigure}
          \hfill
        \begin{subfigure}{0.3\linewidth}
            \centering \includegraphics[width=\linewidth]{{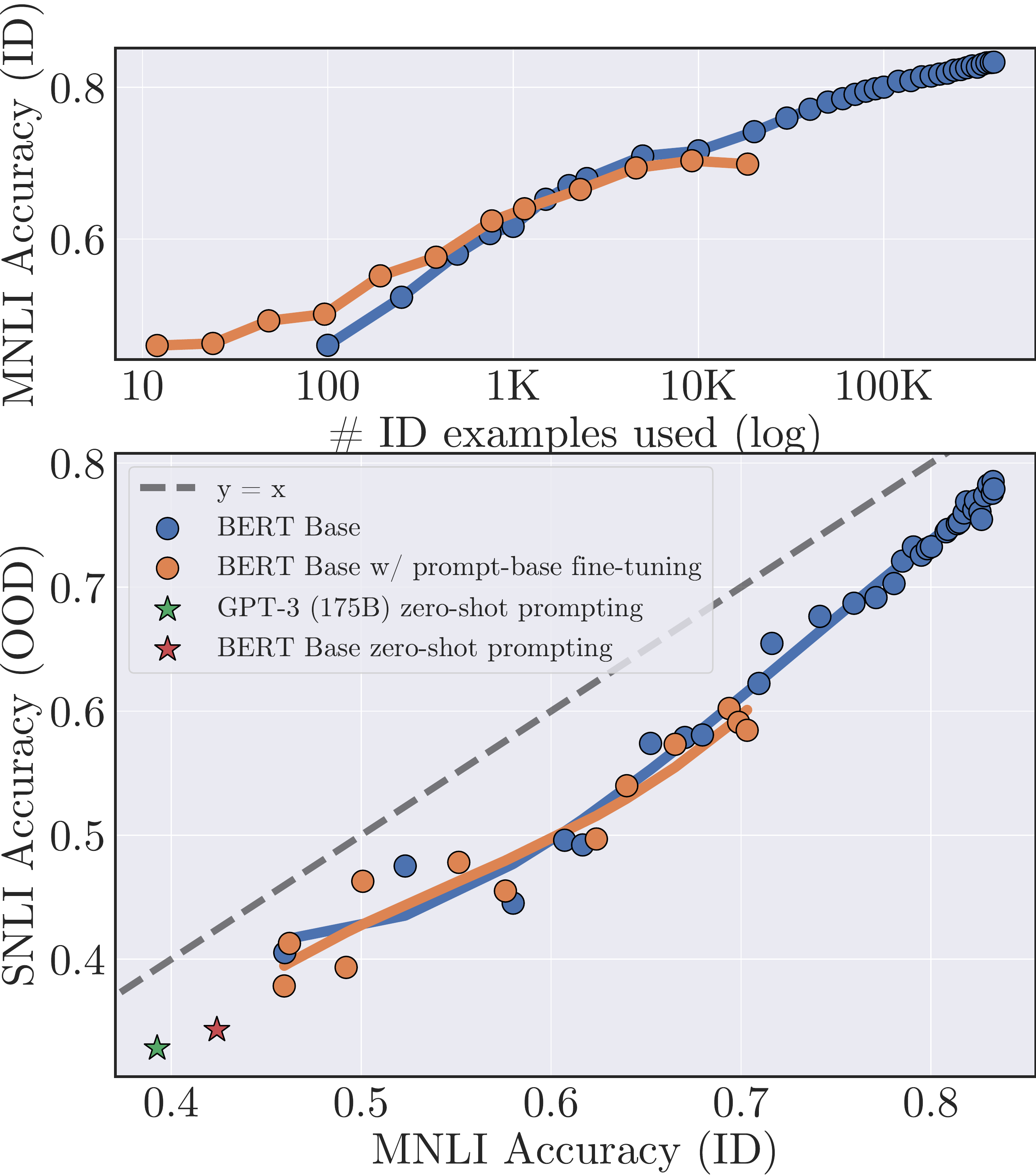}}
            \caption{}\label{fig:finetuning_method_mnli_snli}
          \end{subfigure}
  \caption{
  Prompt-based fine-tuning improves sample efficiency (orange series above blue series) and \emph{average} robustness (orange series about blue series) across experimental settings (\subref{fig:finetuning_method_mnli},\subref{fig:finetuning_method_imdb}). However, it can have no effect on robustness on \emph{individual} OOD settings (e.g., MNLI $\rightarrow$ SNLI; \subref{fig:finetuning_method_mnli_snli}).
  }\label{fig:finetuning_method}
\end{figure*}

\section{Experimental Setup}

We study three modeling interventions---using natural language prompts, increasing pre-trained model size, and pre-training on more data---on 14 total datasets spanning natural language inference (NLI), sentiment analysis, and extractive question answering (QA). See Appendix~\ref{app:experimental_setup_details} for further details about experimental settings.

\paragraph{Tasks and Datasets.} In our natural language inference (NLI) experiments, we use \mnli \citep{williams2018broad}, \snli \citep{bowman2015large}, and MedNLI \citep{romanov-shivade-2018-lessons}. For sentiment analysis, we use IMDb reviews \citet{maas-EtAl:2011:ACL-HLT2011}, SST-2 \citep{socher2013recursive}, and reviews from the ``Movies and TV'' subsection of the Amazon Reviews corpus \citep{ni2019justifying}. Lastly, for extractive question answering, we use SQuAD \citep{rajpurkar-etal-2016-squad}, NaturalQuestions \citep{kwiatkowski-etal-2019-natural}, TriviaQA, BioASQ \citep{tsatsaronis2015overview}, and the four SQuADShifts test sets \citep{miller2020effect}.

\paragraph{Modeling Interventions.} To understand the effect of a particular modeling intervention on sample efficiency and robustness, we evaluate pre-trained models that differ \emph{only} along the axis of interest (e.g., model size or fine-tuning method).
Since the optimal fine-tuning hyperparameters depend on the ID training dataset size, we separately tune hyperparameters for each model on each training dataset subsample size, taking the models that achieve the best held-out ID performance for each setting. See Appendix~\ref{app:hyperparameter_optimization_details} for details about hyperparameter optimization.

\section{Results and Discussion}

\begin{figure*}[!ht]
        \centering
        \begin{subfigure}{0.32\linewidth}
            \centering   \includegraphics[width=\linewidth]{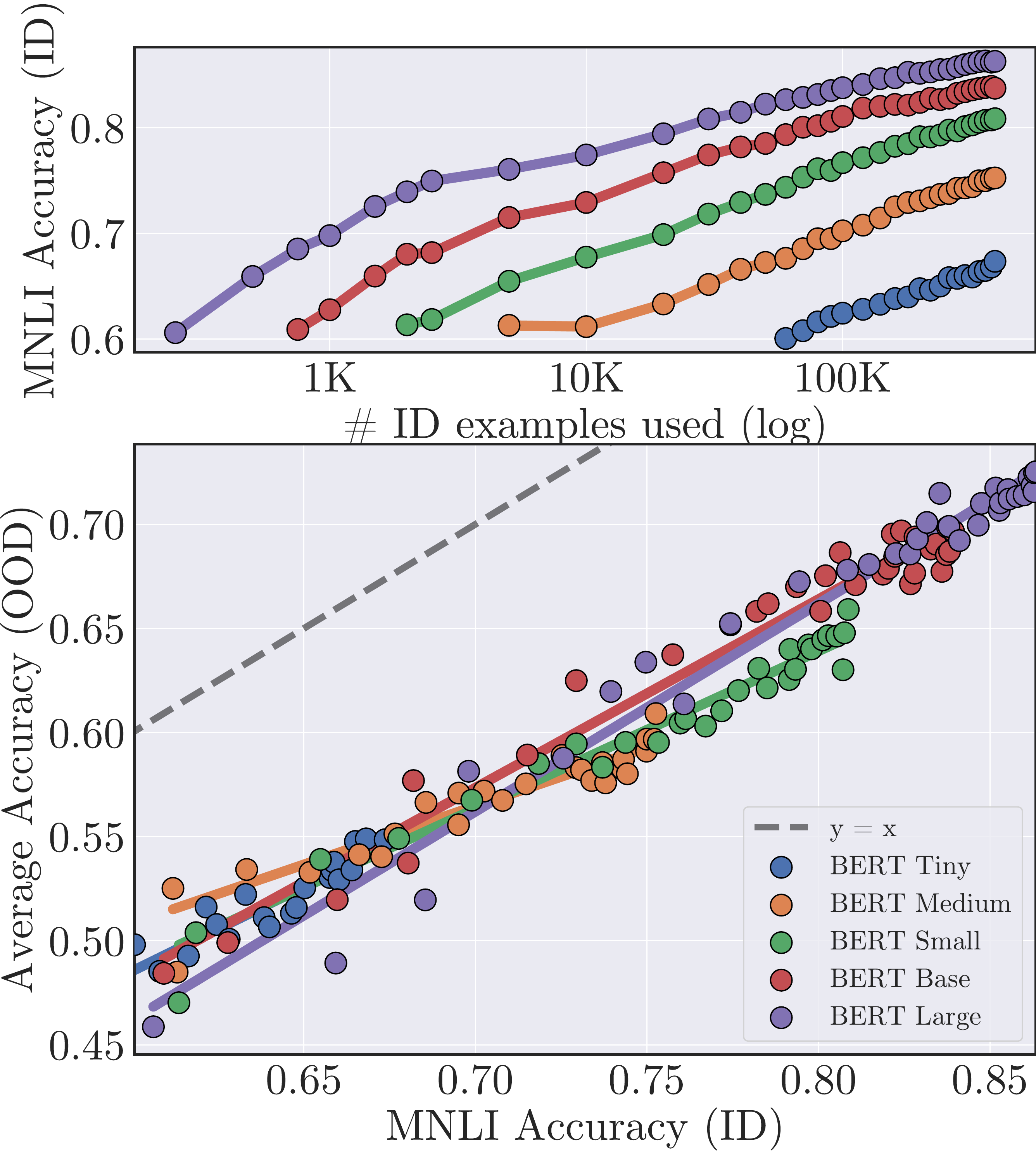}
            \caption{}\label{fig:model_size_mnli}
          \end{subfigure}
          \hfill
        \begin{subfigure}{0.32\linewidth}
            \centering
            \includegraphics[width=\linewidth]{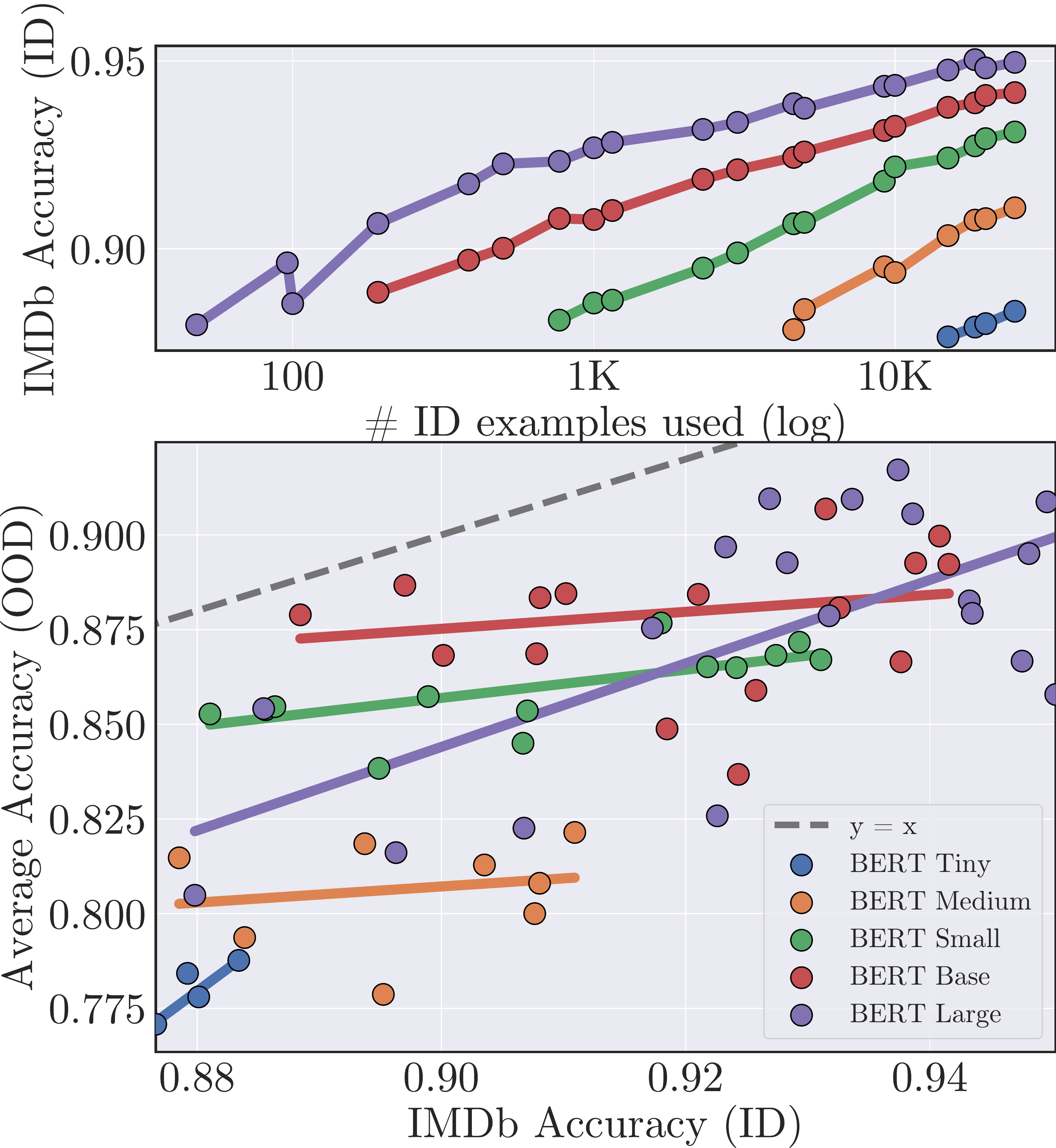}
            \caption{}\label{fig:model_size_imdb}
        \end{subfigure}
        \hfill
        \begin{subfigure}{0.32\linewidth}
            \centering
            \includegraphics[width=\linewidth]{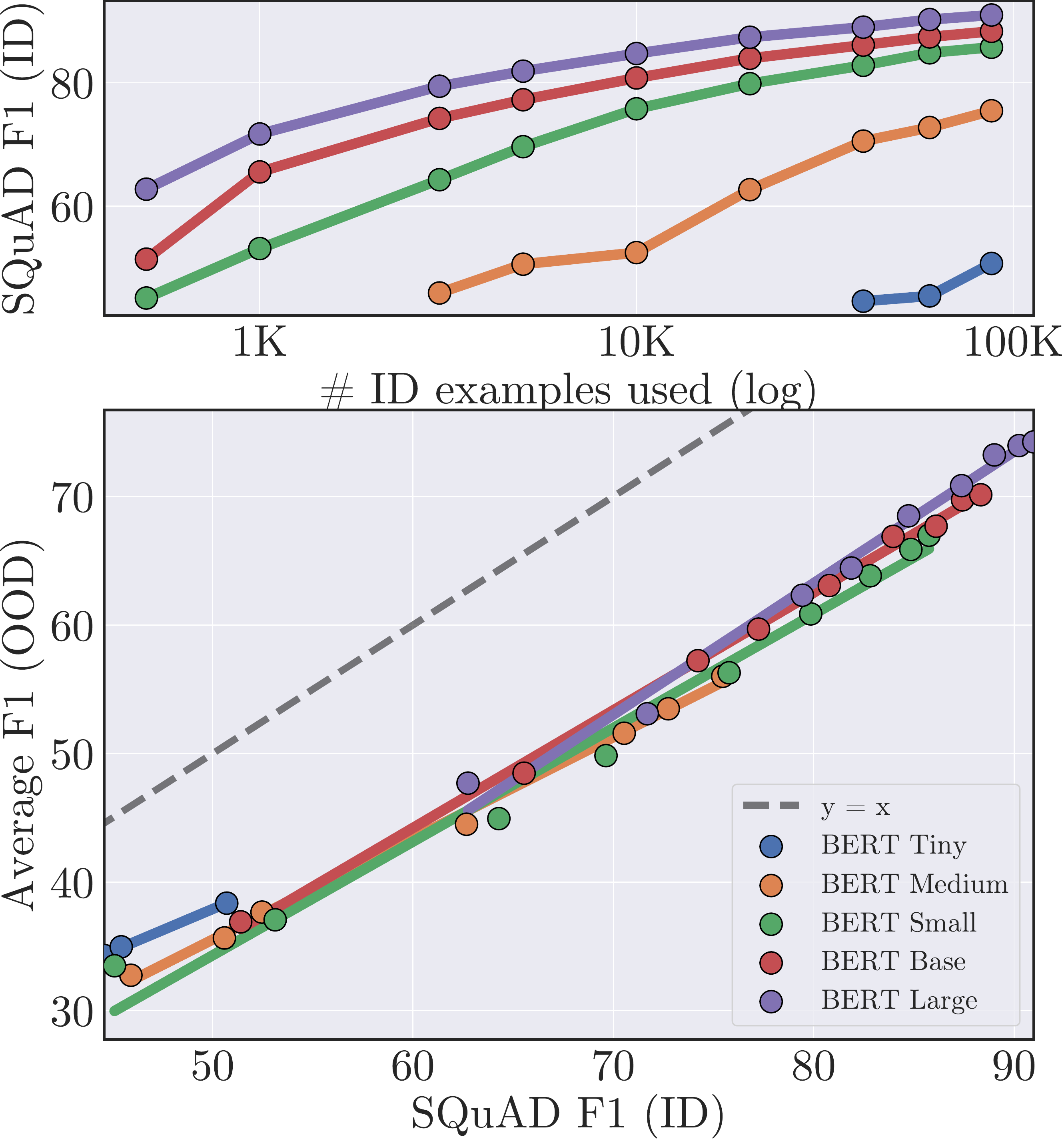}
            \caption{}\label{fig:model_size_squad}
          \end{subfigure}
          \hfill
        \begin{subfigure}{0.32\linewidth}
            \centering
            \includegraphics[width=\linewidth]{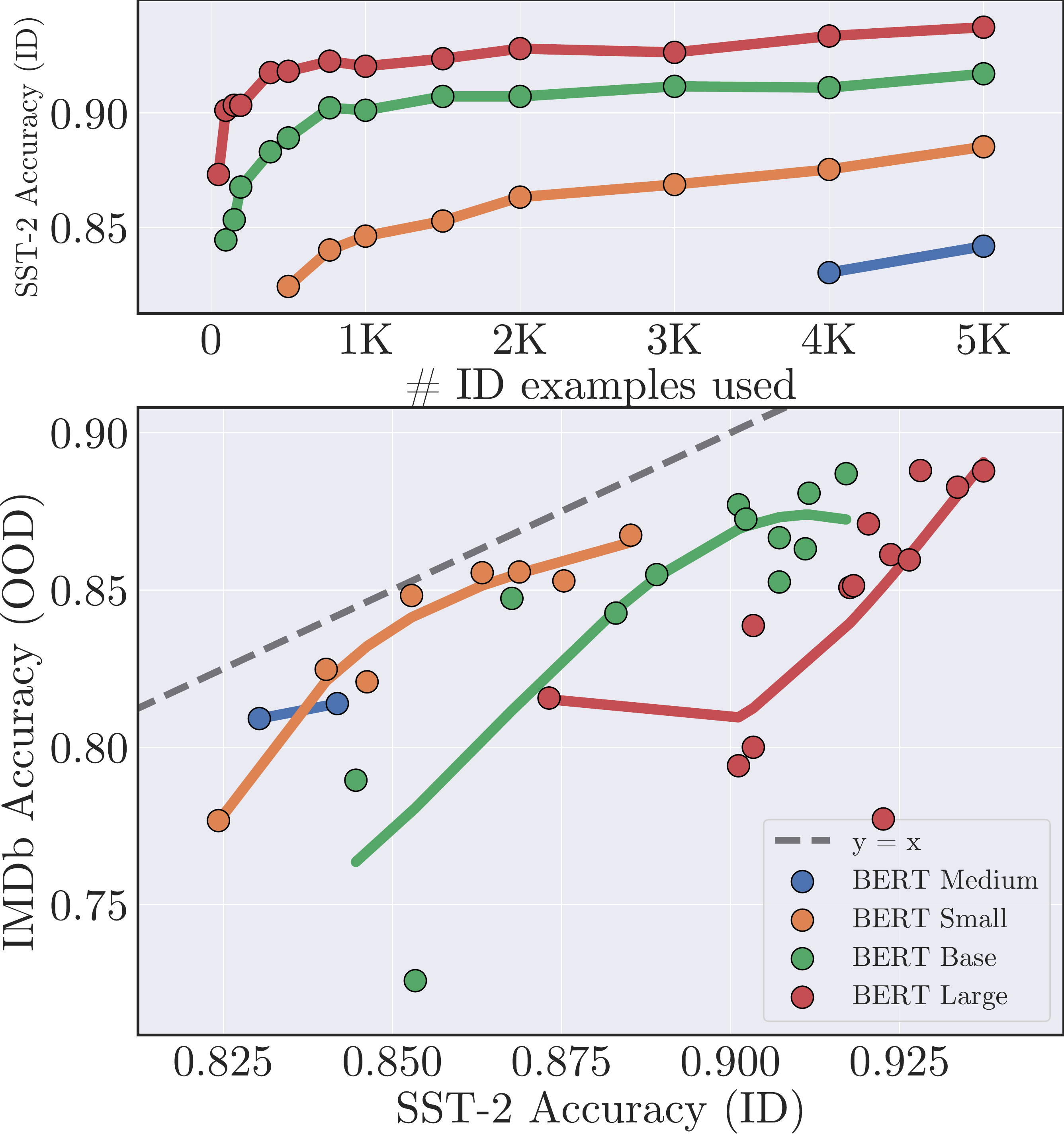}
            \caption{}\label{fig:model_size_sst_imdb}
          \end{subfigure}
        \hfill
        \begin{minipage}[b]{0.63\linewidth}
            \centering
          \caption{Although increasing pre-trained model size improves sample efficiency in all settings, these sample efficiency improvements only translate to better average robustness in sentiment analysis experiments (\subref{fig:model_size_imdb}). In NLI and extractive QA, average robustness is unchanged (\subref{fig:model_size_mnli},\subref{fig:model_size_squad}). Although increased model size improves averaged OOD performance on IMDb, these conclusions do not apply to any ID-OOD pair. For example, increasing pre-trained model size can \emph{decrease} robustness when training on SST-2 and evaluating on IMDb (\subref{fig:model_size_sst_imdb}).}\label{fig:model_size}
          \vspace{4em}
        \end{minipage}
    \end{figure*}

\begin{figure*}[!ht]
        \centering
        \begin{subfigure}{0.32\linewidth}
            \centering
            \includegraphics[width=\linewidth]{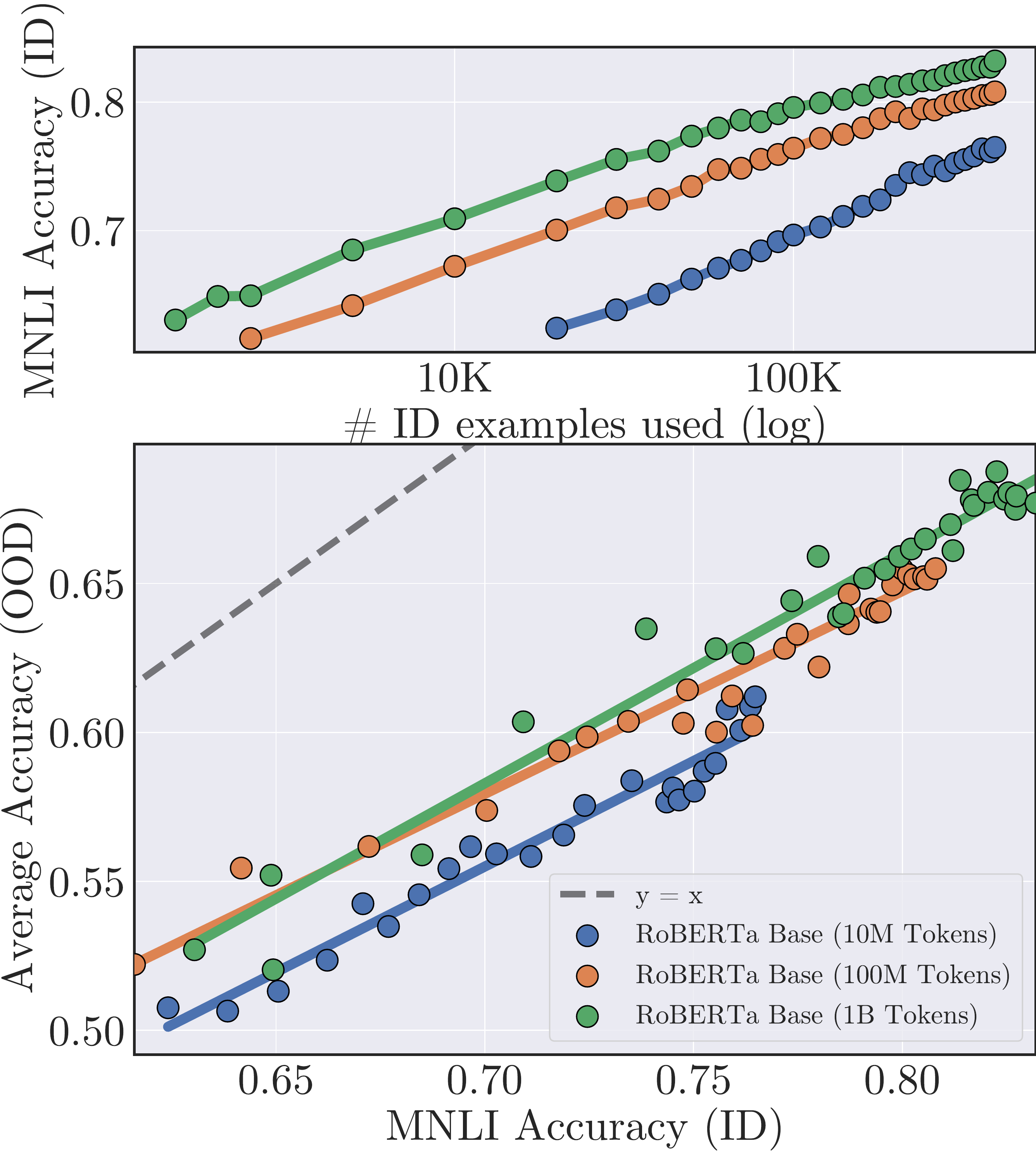}
            \caption{}\label{fig:pretraining_data_mnli}
          \end{subfigure}
          \hfill
        \begin{subfigure}{0.32\linewidth}
            \centering
            \includegraphics[width=\linewidth]{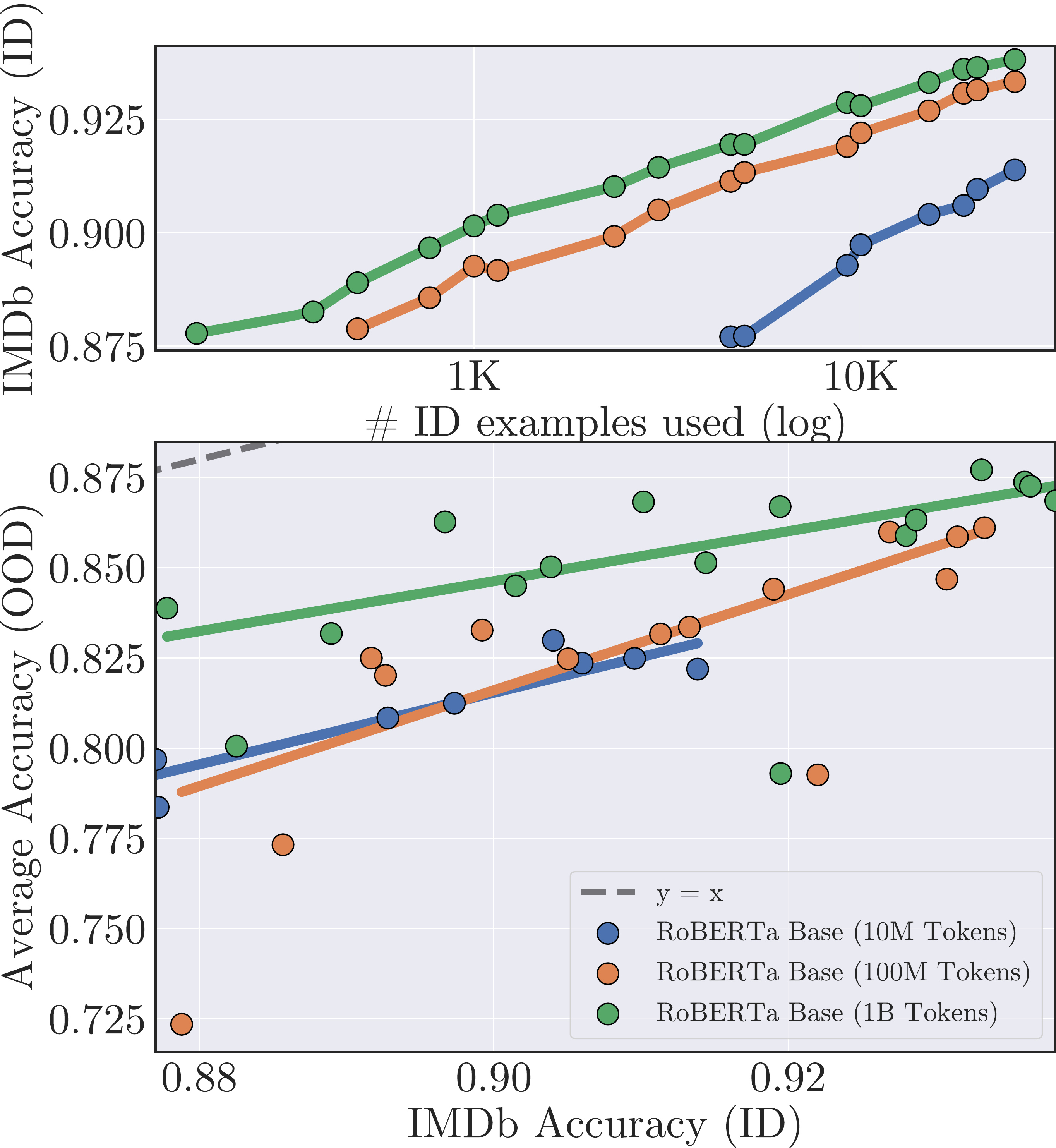}
            \caption{}\label{fig:pretraining_data_imdb}
        \end{subfigure}
        \hfill
        \begin{subfigure}{0.32\linewidth}
            \centering
            \includegraphics[width=\linewidth]{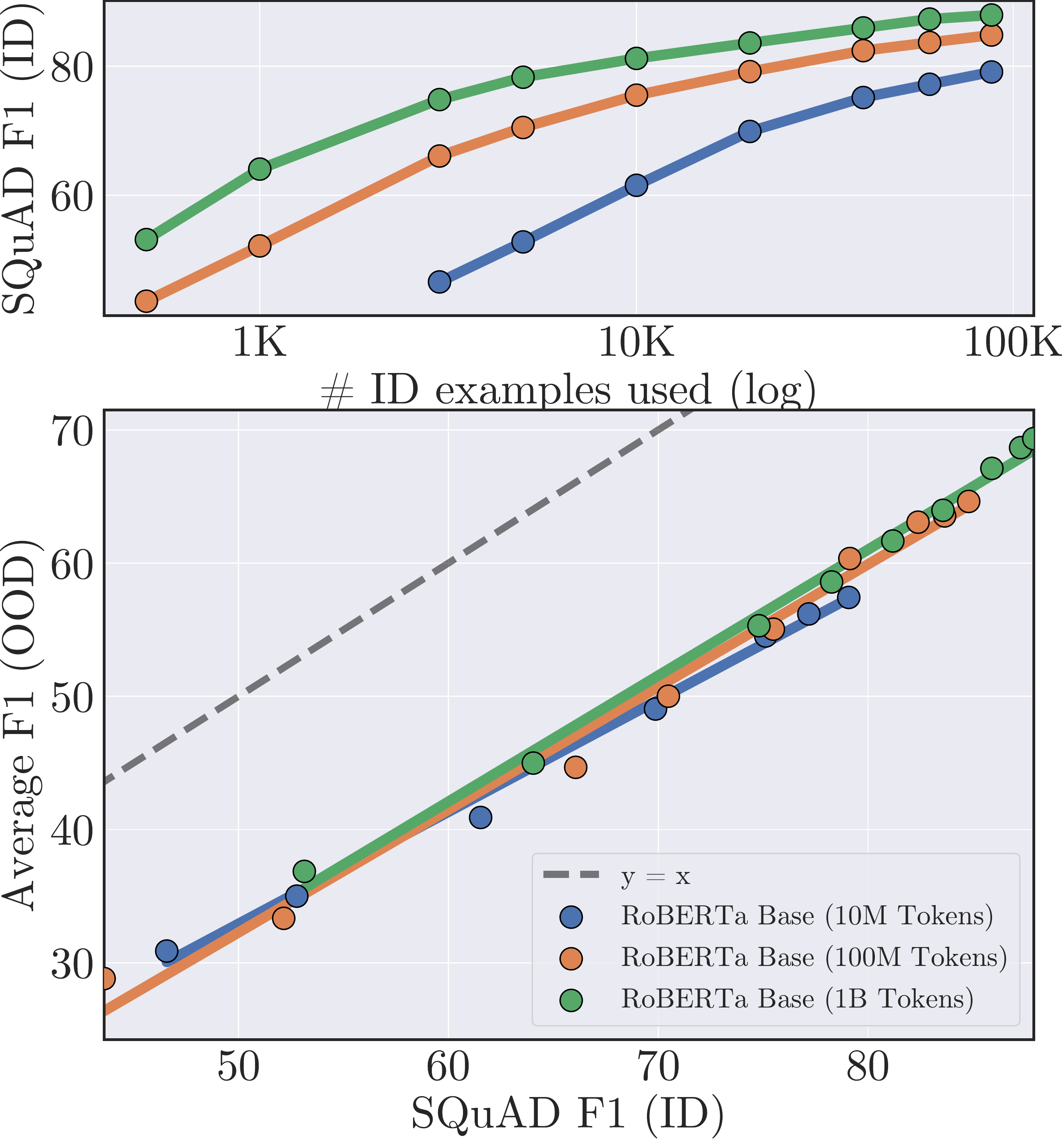}
            \caption{}\label{fig:pretraining_data_squad}
          \end{subfigure}
          \hfill
        \begin{subfigure}{0.32\linewidth}
            \centering
            \includegraphics[width=\linewidth]{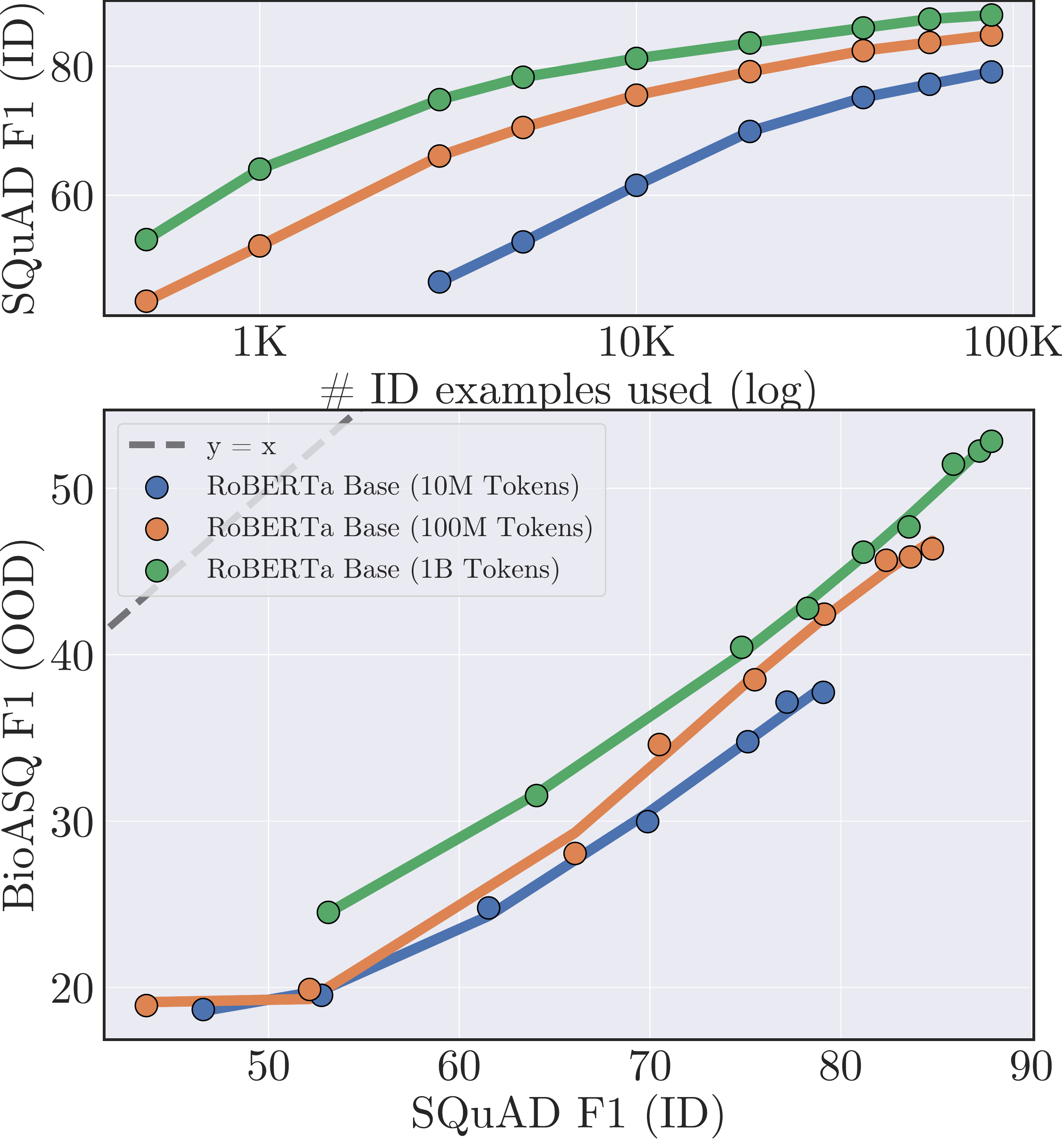}
            \caption{}\label{fig:pretraining_data_squad_bioasq}
          \end{subfigure}
          \hfill
        \begin{minipage}[b]{0.63\linewidth}
        \caption{Pre-training on more data is an effective method for improving sample efficiency, but these sample efficiency improvements are not always accompanied by robustness improvements. In NLI and sentiment analysis experiments, these sample efficiency gains correlate with improved average robustness (\subref{fig:pretraining_data_mnli},\subref{fig:pretraining_data_imdb}). However, there are no average robustness gains in extractive QA (\subref{fig:pretraining_data_squad}). Despite no average robustness improvement in extractive QA, pre-training on more data can still improve robustness on particular test sets (e.g., BioASQ; \subref{fig:pretraining_data_squad_bioasq}).}\label{fig:pretraining_data}
        \vspace{4em}
        \end{minipage}
    \end{figure*}

Our results show that models with higher sample efficiency may not necessarily have higher average OOD robustness---different tasks and modeling interventions affect robustness in different ways (Figures~\ref{fig:finetuning_method}-\ref{fig:pretraining_data}). For example, prompt-based fine-tuning consistently improves both sample efficiency \emph{and} average robustness, but only in low-data settings (Figure~\ref{fig:finetuning_method}). In contrast, increasing model size improves sample efficiency across the range of training dataset sizes and tasks, but only improves average robustness on sentiment analysis (Figure~\ref{fig:model_size}). On individual datasets, we even observe cases where models with \emph{lower} sample efficiency have higher robustness (Figure~\ref{fig:model_size_sst_imdb}). See Appendix~\ref{app:full_results} for full results on every ID-OOD setting.

\paragraph{Natural Language Prompting.}
We compare \bertbase models using (1)~standard fine-tuning, (2)~prompt-based fine-tuning, and (3)~zero-shot prompting. We also compare these results with zero-shot prompting of \texttt{text-davinci-001}, a much larger model trained on substantially more data. We run experiments on NLI and sentiment analysis, since extractive QA is not amenable to prompt-based fine-tuning with masked language models.

Figures~\ref{fig:finetuning_method_mnli} and \ref{fig:finetuning_method_imdb} plot the average performance on all OOD datasets as a function of ID performance and the ID performance as a function of the number of labeled training examples. Sample efficiency improvements from prompt-based fine-tuning also translate to higher average robustness. However these improvements only apply in the few-shot setting. As the size of the training dataset increases, the improvements in sample efficiency and average robustness steadily diminish. When using sufficiently large training datasets, models trained with prompt-based fine-tuning yield essentially the same sample efficiency and robustness results as standard fine-tuning ($\sim$1K examples for NLI, $\sim$130 examples for sentiment).

However, results on individual OOD test sets can significantly differ from averaged-OOD trends. For example, Figure~\ref{fig:finetuning_method_mnli_snli} shows that prompt-based fine-tuning on MNLI and evaluating on SNLI improves sample efficiency in the few-shot setting but without any robustness improvements.

Surprisingly, we also find that zero-shot inference does not necessarily improve average robustness over prompt-based fine-tuning---zero-shot performance lies on or below the trend line formed by prompt-based fine-tuning, despite not using any ID-specific data at all.
See Appendix~\ref{app:full_results_natural_language_prompting} for full results of increasing pre-trained model size for every ID-OOD setting.

\paragraph{Increasing Pre-Trained Model Size.} We run experiments with the checkpoints of \citet{turc2019well}, who pre-train BERT models with various numbers of transformer layers (L) and hidden embedding sizes (H).
We run experiments on NLI, sentiment analysis, and extractive QA to compare pre-trained models of five sizes: (1)~Large (L=24, H=1024), (2)~Base (L=12, H=768), (3)~Medium (L=8, H=512), (4)~Small (L=4, H=512), and (5)~Tiny (L=2, H=128).
Although increasing the pre-trained model size improves sample efficiency on every task, it does not always improve average robustness (Figure~\ref{fig:model_size}). In particular, increasing model size minimally affects average robustness in NLI and extractive QA (Figure~\ref{fig:model_size_mnli},\ref{fig:model_size_squad}), but substantially improves average robustness on sentiment analysis (Figure~\ref{fig:model_size_imdb}).\footnote{Note that moving from \bertbase to \bertlarge does not improve effective robustness until $\sim$92\% IMDb ID accuracy. We hypothesize this occurs because these \bertlarge datapoints are fine-tuned on small amounts of data (fewer than 1K examples), potentially leading to instability and reduced effective robustness.} However, results on individual ID-OOD pairs can again significantly differ from average OOD performance trends. For example, when training on SST-2 and evaluating on IMDb, larger models actually have \emph{lower} OOD performance. This occurs because SST-2 examples (single sentences) are significantly shorter than IMDb examples (paragraphs). As a result, models trained on the shorter SST-2 examples struggle when evaluated on IMDb because this particular ID-OOD pair requires length extrapolation, and increasing pre-trained model size does not help models generalize to longer input sequences. As a result, effective robustness decreases because larger models have higher ID (SST-2) performance but unchanged OOD (IMDb) performance.
See Appendix~\ref{app:full_results_increasing_model_size} for full results of natural language prompting for every ID-OOD setting.

\paragraph{Pre-Training on More Data.} We conduct NLI, sentiment, and QA experiments with RoBERTa models pre-trained on 10M, 100M, and 1B tokens of web text \citep{zhang-etal-2021-need}.

Pre-training on more data consistently improves sample efficiency, but only yields average robustness improvements in NLI and sentiment analysis (Figure~\ref{fig:pretraining_data_mnli},\subref{fig:pretraining_data_imdb}). In extractive QA experiments, varying the amount of pre-training data does not significantly change average robustness (Figure~\ref{fig:pretraining_data_squad}). Again, we find that results on average OOD performance are not predictive of results on individual test sets---despite unchanged average OOD robustness when pre-training on more data, OOD performance can be higher on individual extractive QA test sets (e.g., SQuAD $\rightarrow$ BioASQ; Figure~\ref{fig:pretraining_data_squad_bioasq}).
See Appendix~\ref{app:full_results_pretraining_on_more_data} for full results of pre-training on more data for every ID-OOD setting.

\section{Conclusion}

We study the relationship between sample efficiency and robustness across three tasks and three modeling interventions, finding that sample efficiency improvements often fail to translate to improved robustness. As larger models quickly become more sample efficient, our results caution that sample efficiency and robustness are different axes of improvement and that optimizing for sample efficiency will not necessarily always yield robustness gains.

\section*{Acknowledgments}

We thank the anonymous reviewers for their feedback and comments that helped improve this work.
We also thank Kevin Lin and Eric Wallace for their feedback and useful discussions. NL was supported by an NSF Graduate Research Fellowship under grant number DGE-1656518. Other funding was provided by a PECASE Award and the Open Philantropy Project.

\section*{Limitations}

Our study focuses on natural language understanding tasks, though it may also be interesting to study whether these trends apply in natural language generation tasks (e.g., summarization). In particular, it's possible that zero- or few-shot pre-trained models may do better on generation tasks because these tasks are more similar to the models' original pre-training objective (e.g., language modeling).

Furthermore, we compared few-shot prompt-based fine-tuning, zero-shot inference, and standard fine-tuning. However, other methods of adapting models to labeled ID data can have very different sample efficiency properties (e.g., in-context learning). Future work could explore whether these results hold with few-shot in-context learning or parameter-efficient fine-tuning tuning (e.g., adapaters; \citealp{conf/icml/HoulsbyGJMLGAG19}).

\bibliography{custom}
\bibliographystyle{acl_natbib}

\appendix

\section{Experimental Setup Details}\label{app:experimental_setup_details}

\paragraph{Natural Language Inference.} We use \mnli \citep{williams2018broad} and \snli \citep{bowman2015large} as ID datasets. We use \mnli, \snli and MedNLI \citep{romanov-shivade-2018-lessons} as OOD test sets. All of our ID datasets have three labels (\emph{entailment}, \emph{contradiction}, \emph{neutral}).

We also evaluate OOD on HANS \citep{mccoy2019right}, a diagnostic dataset targeting lexical overlap, an \idspecificfeature{} in \snli and \mnli. In \mnli and \snli, the majority of examples with high lexical overlap between the NLI premise and hypothesis have the ``entailment'' label. In HANS, 50\% of examples support this heuristic, and 50\% contradict it, so a model that exclusivly relies on the word overlap heuristic would have an accuracy of 50\%.but HANS has two labels (\emph{entailment}, \emph{non-entailment}). To evaluate our 3-class models on 2-class HANS, we follow \citet{mccoy2019right} and translate \emph{contradiction} or \emph{neutral} model predictions to \emph{non-entailment}.

We train on the \mnli and \snli training sets. We evaluate on the \mnli matched development set, the \snli test set, and the HANS evaluation split. When evaluating OOD on MedNLI, we evaluate on the \emph{training set} ($\sim$11K examples) because the development and test sets are quite small ($\sim$1.5K examples each).

\paragraph{Sentiment Analysis.} We use the IMDb reviews dataset of \citep{maas-EtAl:2011:ACL-HLT2011}, SST-2 \citep{socher2013recursive} as ID datasets. We use IMDb, SST-2, and reviews from the ``Movies and TV'' subsection of the Amazon Reviews corpus \citep{ni2019justifying} as OOD datasets.

These datasets are all binary classification, where reviews are labeled as \emph{positive} or \emph{negative} sentiment. To construct the ``Movies and TV'' Amazon review sentiment dataset, we randomly select one- or two-star (negative) reviews and four- or five-star (positive) reviews from the full Amazon Reviews corpus, using 25,000 examples for training, 10,000 examples for development, and 10,000 examples for testing. Each of these splits is balanced.

We train on the IMDb, SST, and Amazon Reviews training splits, and use the corresponding evaluation splits to measure ID performance. When evaluating OOD on SST, we use the concatenation of the train and test sets (8471 examples in total), since the original test set is quite small (1821 examples). Beyond this exception, we use each dataset's evaluation split for OOD evaluation.

\paragraph{Extractive Question Answering.}
We use SQuAD \citep{rajpurkar-etal-2016-squad} and NaturalQuestions \citep{kwiatkowski-etal-2019-natural} as ID datasets.
We use SQuAD, NaturalQuestions, TriviaQA, BioASQ \citep{tsatsaronis2015overview}, and the SQuADShifts test sets of \citet{miller2020effect} as OOD datasets.

The SQuADShifts test sets were constructed following the original SQuAD crowdsourcing procedure, but with passages drawn from both the original Wikipedia domain, as well as the New York Times (NYT), Amazon reviews, and Reddit. For NaturalQuestions, we only consider questions over paragraphs (as opposed to those over tables and lists). We use the MRQA 2019 Shared Task versions of TriviaQA and BioASQ \citep{fisch-etal-2019-mrqa}. We also use the MRQA 2019 Shared Task version of NaturalQuetsions, but only include examples questions over paragraphs (removing those with questions over tables or lists). In all of these extractive QA datasets, models are given a passage and a question and tasked with identifying a substring of the passage that answers the question.

We train on the SQuAD and NaturalQuestions training splits, and use the corresponding evaluation splits to measure ID performance.  When evaluating OOD on BioASQ, we use the concatenation of the train, development, and test sets (3977 examples in total), since the original test set is quite small (1518 examples). Beyond this exception, we use each dataset's evaluation split for OOD evaluation.

\section{Hyperparameter Optimization Details}\label{app:hyperparameter_optimization_details}

We conduct extensive hyperparameter optimization when training models on a particular ID dataset (or a subsample thereof). We re-tune hyperparameters for each subsample size, since the optimal value of certain hyperparameters may depend on number of available training examples (e.g., batch size and learning rate). For each experimental setting, we use a combination of (1)~previously-reported hyperparameters (taken from prior work) and (2)~random search (10 samples) over a pre-defined grid of reasonable hyperparameter values. For each experiment, we take the checkpoint with the best ID performance.

\paragraph{Natural Language Inference.} For every NLI ID-OOD setting, we run experiments with the cross-product of learning rates in \{1e-5, 2e-5, 3e-5\} with batch sizes of \{16, 32\}. We also sample additional runs from the following grid:

\begin{itemize}[noitemsep,topsep=0pt]
    \item Random seed: [0, 100000]
    \item Learning rate: \{1e-5, 2e-5, 3e-5\}
    \item Batch size: \{16, 32\}
    \item Number of training epochs: \{10\}
\end{itemize}

\paragraph{Sentiment Analysis.} For every sentiment analysis ID-OOD setting, we run experiments with the cross-product of learning rates in \{1e-5, 2e-5, 3e-5, 5e-5\} with batch sizes of \{16, 32\} and training for \{20, 50\} epochs. We also sample additional runs from the following grid:

\begin{itemize}[noitemsep,topsep=0pt]
    \item Random seed: [0, 100000]
    \item Learning rate: \{1e-5, 2e-5, 3e-5, 5e-5\}
    \item Batch size: \{16, 32\}
    \item Number of training epochs: \{20, 50\}
\end{itemize}

\paragraph{Extractive Question Answering.} For every extractive question answering ID-OOD setting, we run experiments with the cross-product of learning rates in \{2e-5, 3e-5, 5e-5\} with batch sizes of \{16, 32\}. We also sample additional runs from the following grid:

\begin{itemize}[noitemsep,topsep=0pt]
    \item Random seed: [0, 100000]
    \item Learning rate: \{2e-5, 3e-5, 5e-5\}
    \item Batch size: \{16, 32\}
    \item Number of training epochs: \{4\}
\end{itemize}

\clearpage
\onecolumn

\section{Results of All Methods on All ID-OOD Settings}\label{app:full_results}

\subsection{Natural Language Prompting}\label{app:full_results_natural_language_prompting}

\begin{figure*}[!h]
        \centering
        \begin{subfigure}{0.32\linewidth}
            \centering
            \includegraphics[width=\linewidth]{figures/finetuning_method_mnli_dev_vs_snli_test.pdf}
            \caption{}
          \end{subfigure}
          \hfill
        \begin{subfigure}{0.32\linewidth}
            \centering
            \includegraphics[width=\linewidth]{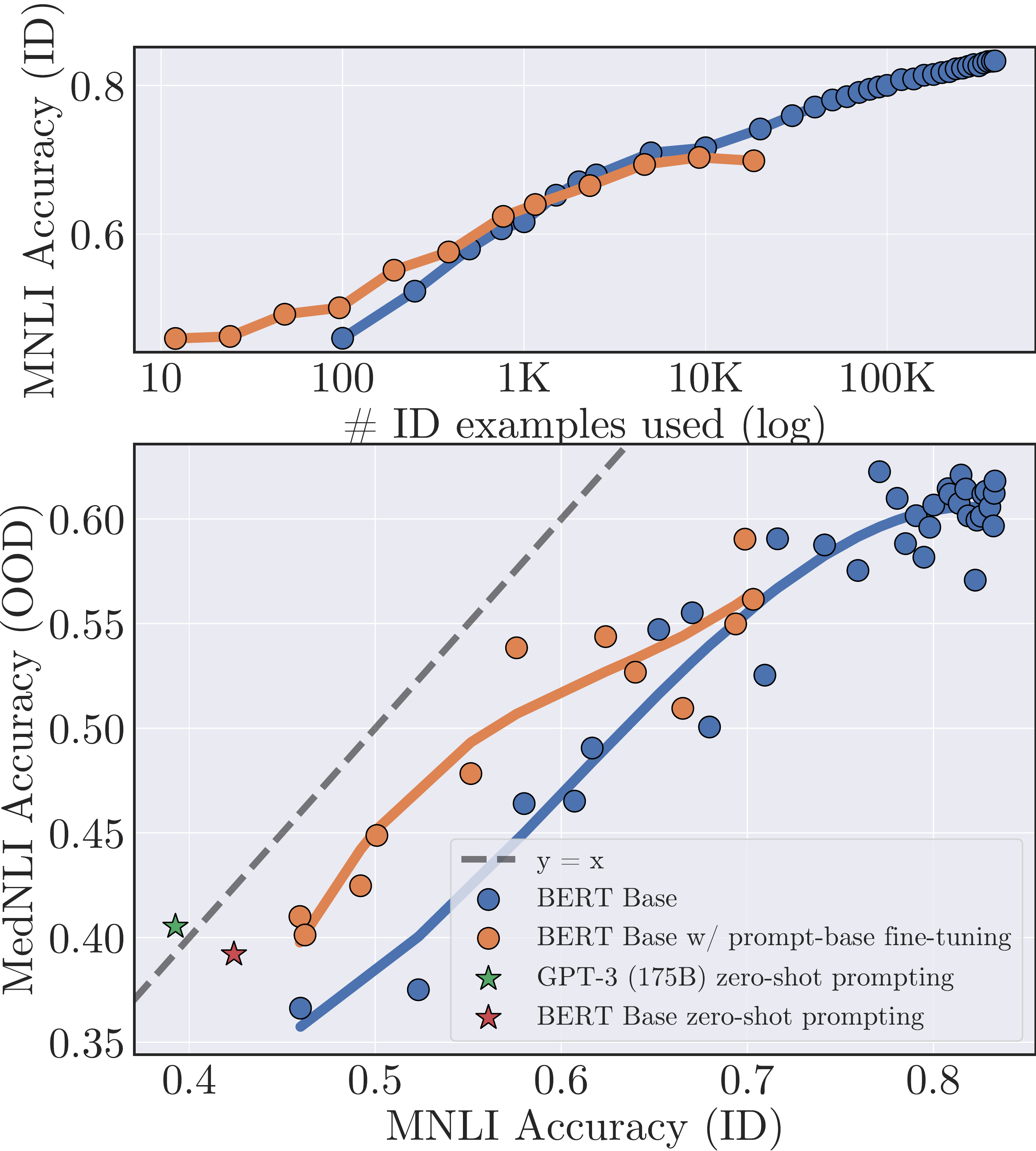}
            \caption{}
        \end{subfigure}
        \hfill
        \begin{subfigure}{0.32\linewidth}
            \centering
            \includegraphics[width=\linewidth]{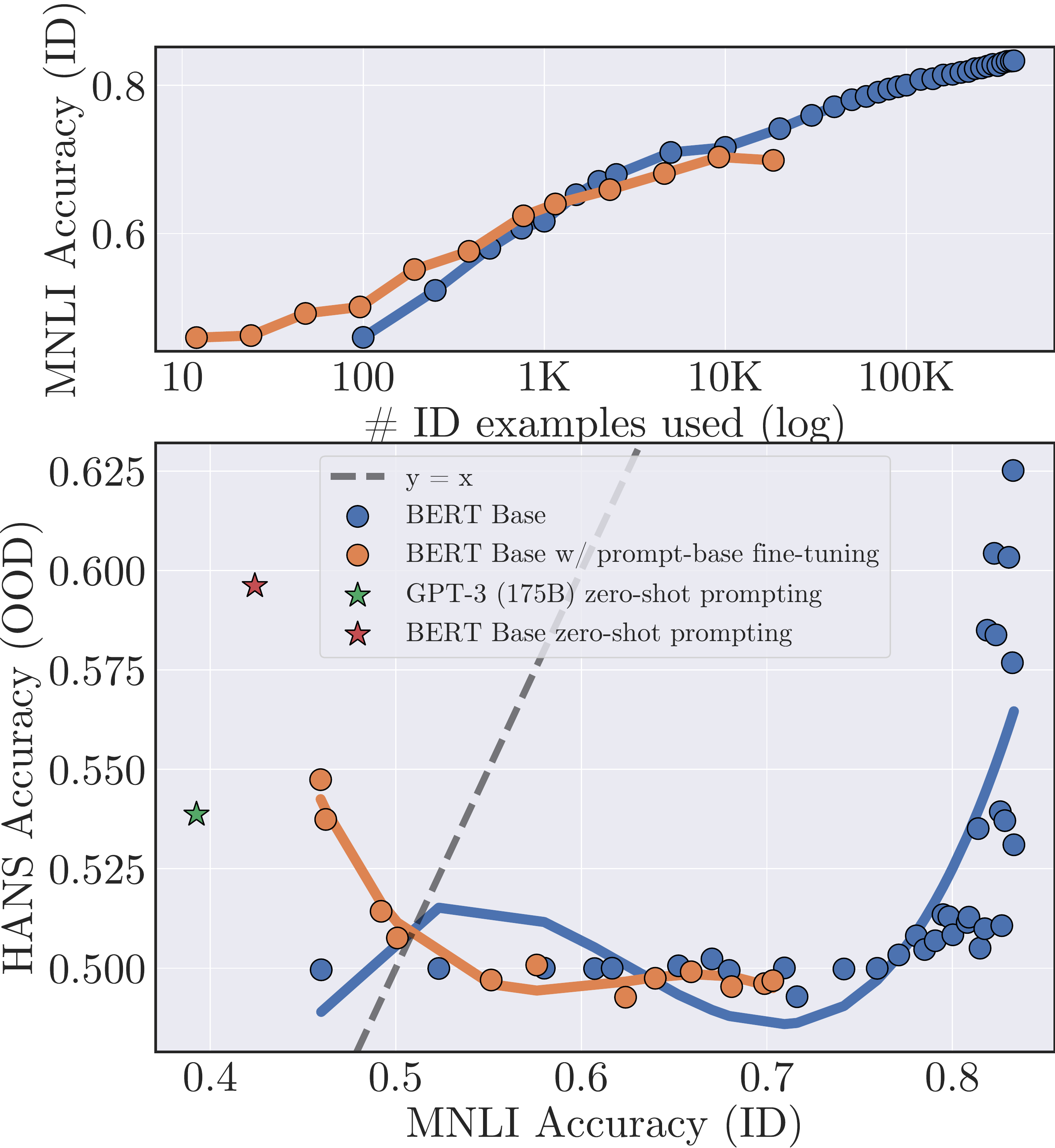}
            \caption{}
          \end{subfigure}
        \begin{subfigure}{0.32\linewidth}
            \centering
            \includegraphics[width=\linewidth]{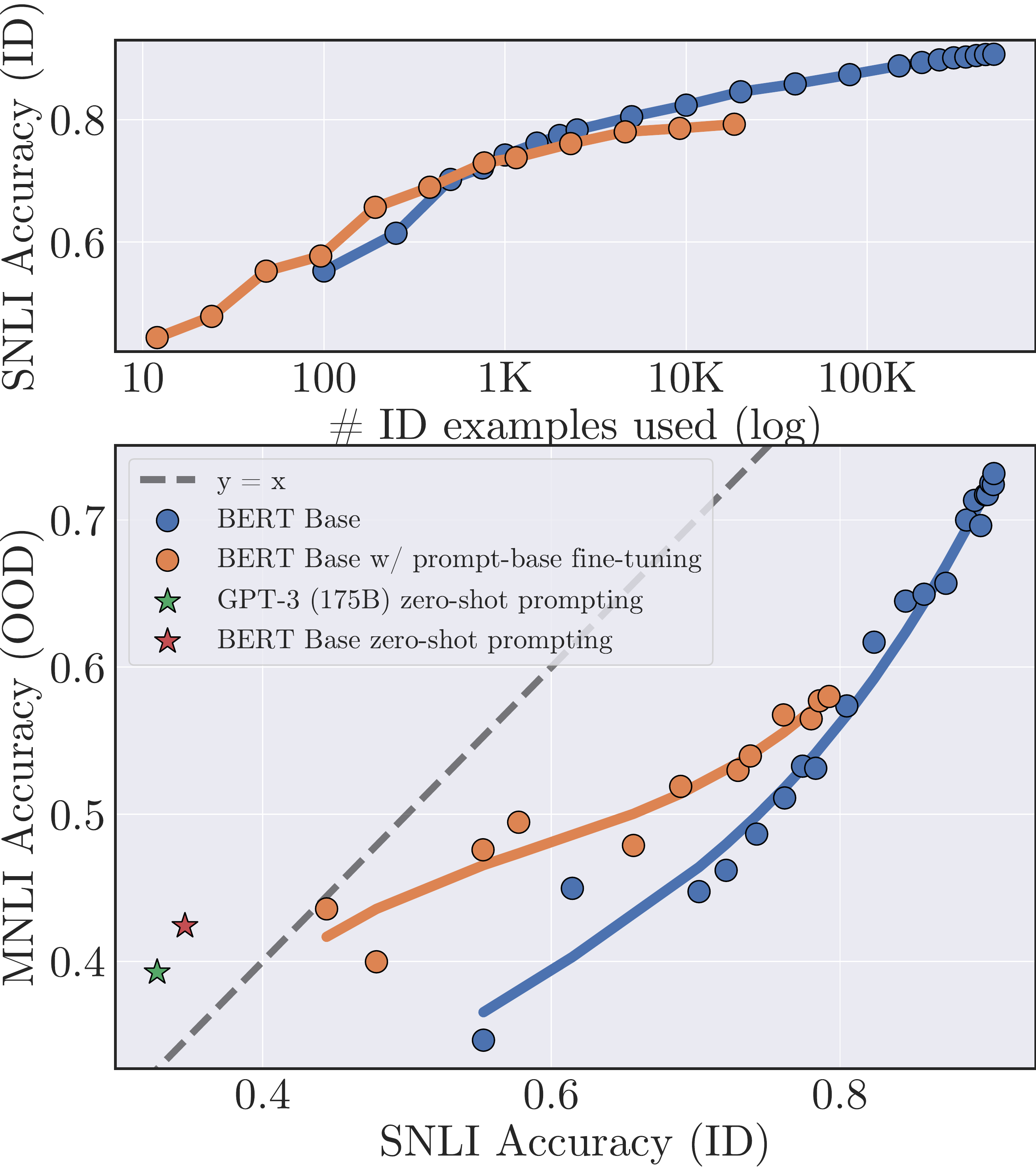}
            \caption{}
          \end{subfigure}
          \hfill
        \begin{subfigure}{0.32\linewidth}
            \centering
            \includegraphics[width=\linewidth]{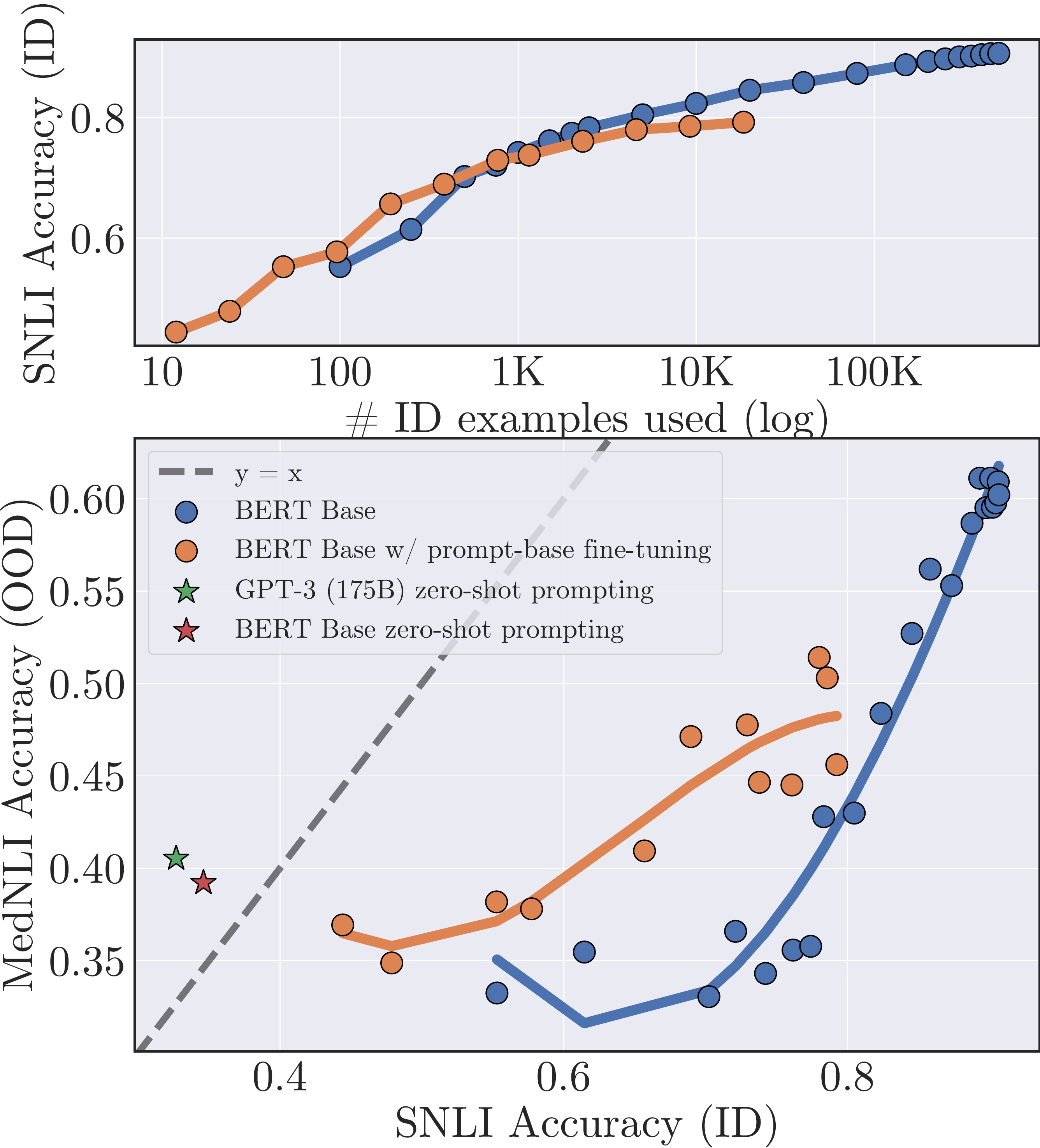}
            \caption{}
        \end{subfigure}
        \hfill
        \begin{subfigure}{0.32\linewidth}
            \centering
            \includegraphics[width=\linewidth]{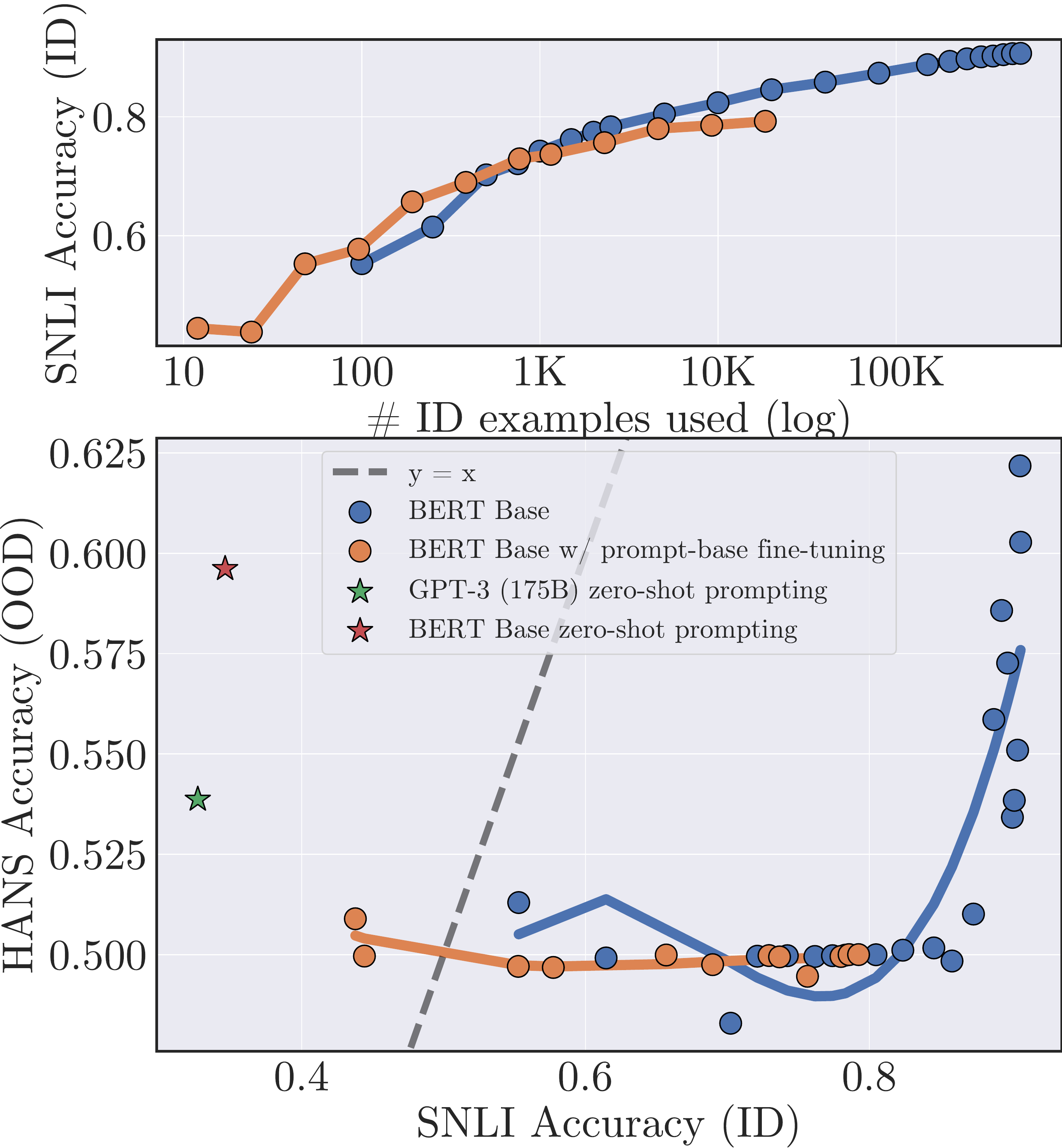}
            \caption{}
          \end{subfigure}
        \caption{Results on all NLI ID-OOD settings when comparing zero-shot prompting, prompt-based fine-tuning, and standard fine-tuning.}
    \end{figure*}

\begin{figure*}[!h]
        \centering
        \begin{subfigure}{0.32\linewidth}
            \centering
            \includegraphics[width=\linewidth]{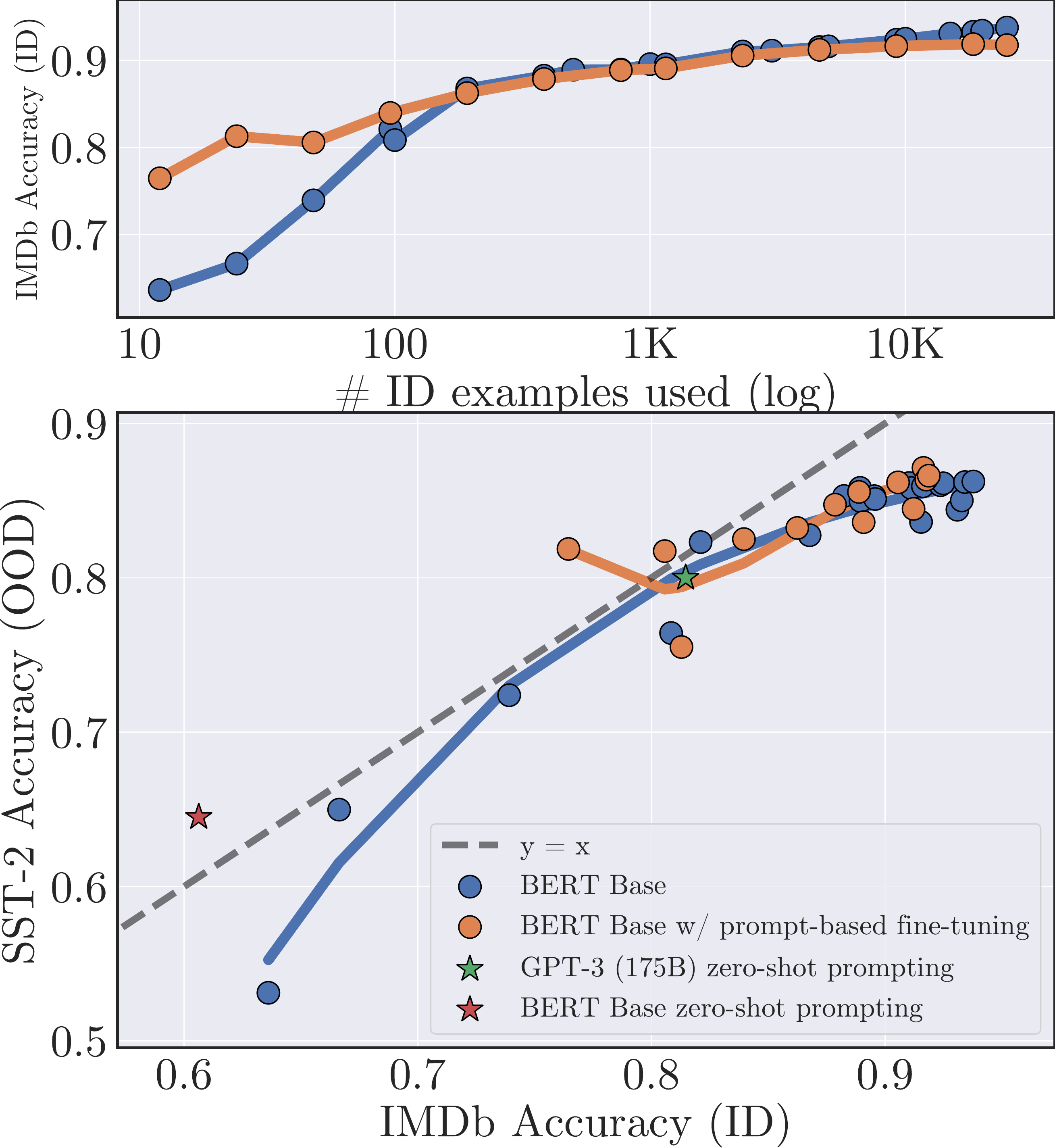}
            \caption{}
          \end{subfigure}
          \hfill
        \begin{subfigure}{0.32\linewidth}
            \centering
            \includegraphics[width=\linewidth]{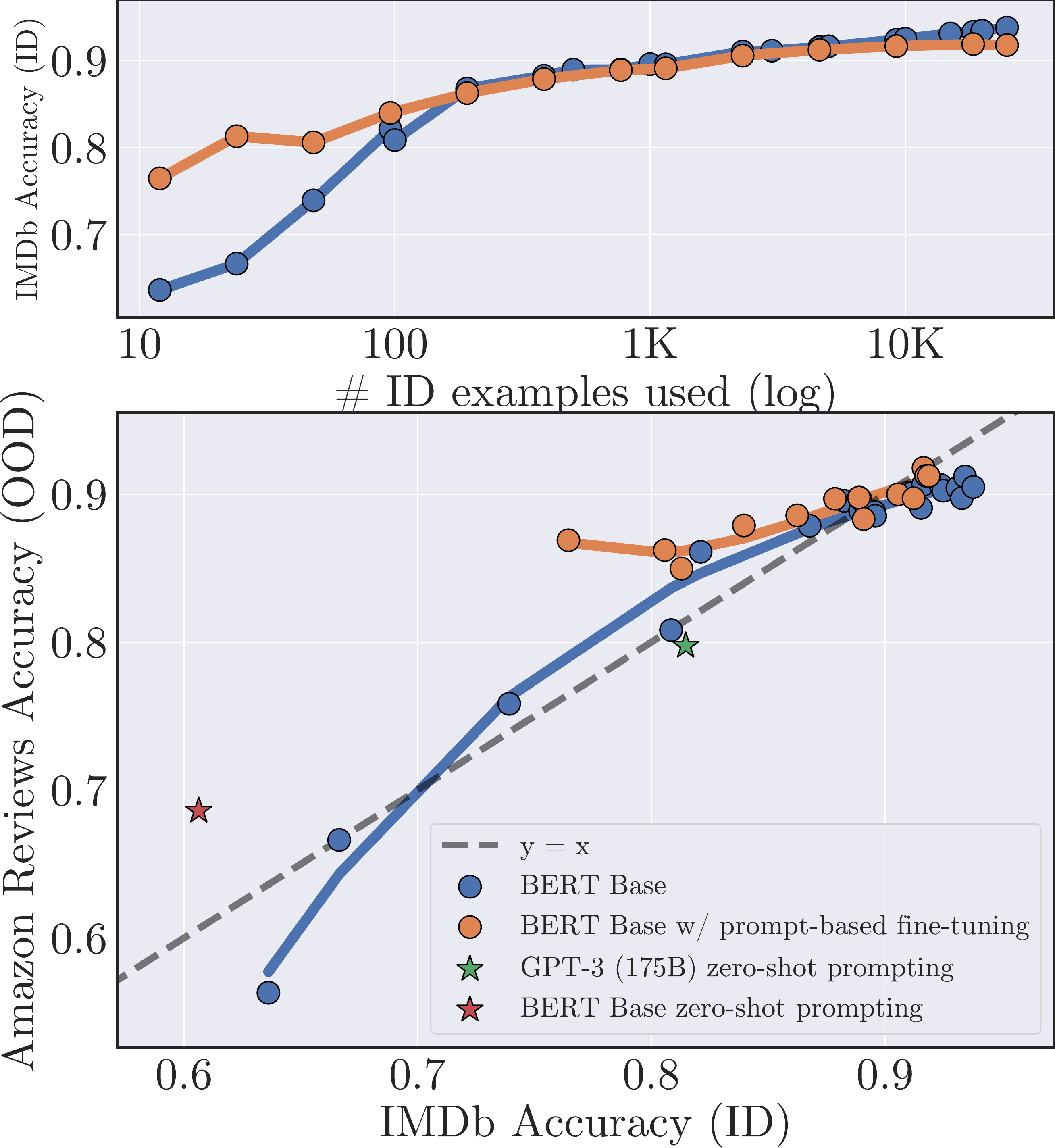}
            \caption{}
        \end{subfigure}
        \hfill
        \begin{subfigure}{0.32\linewidth}
            \centering
            \includegraphics[width=\linewidth]{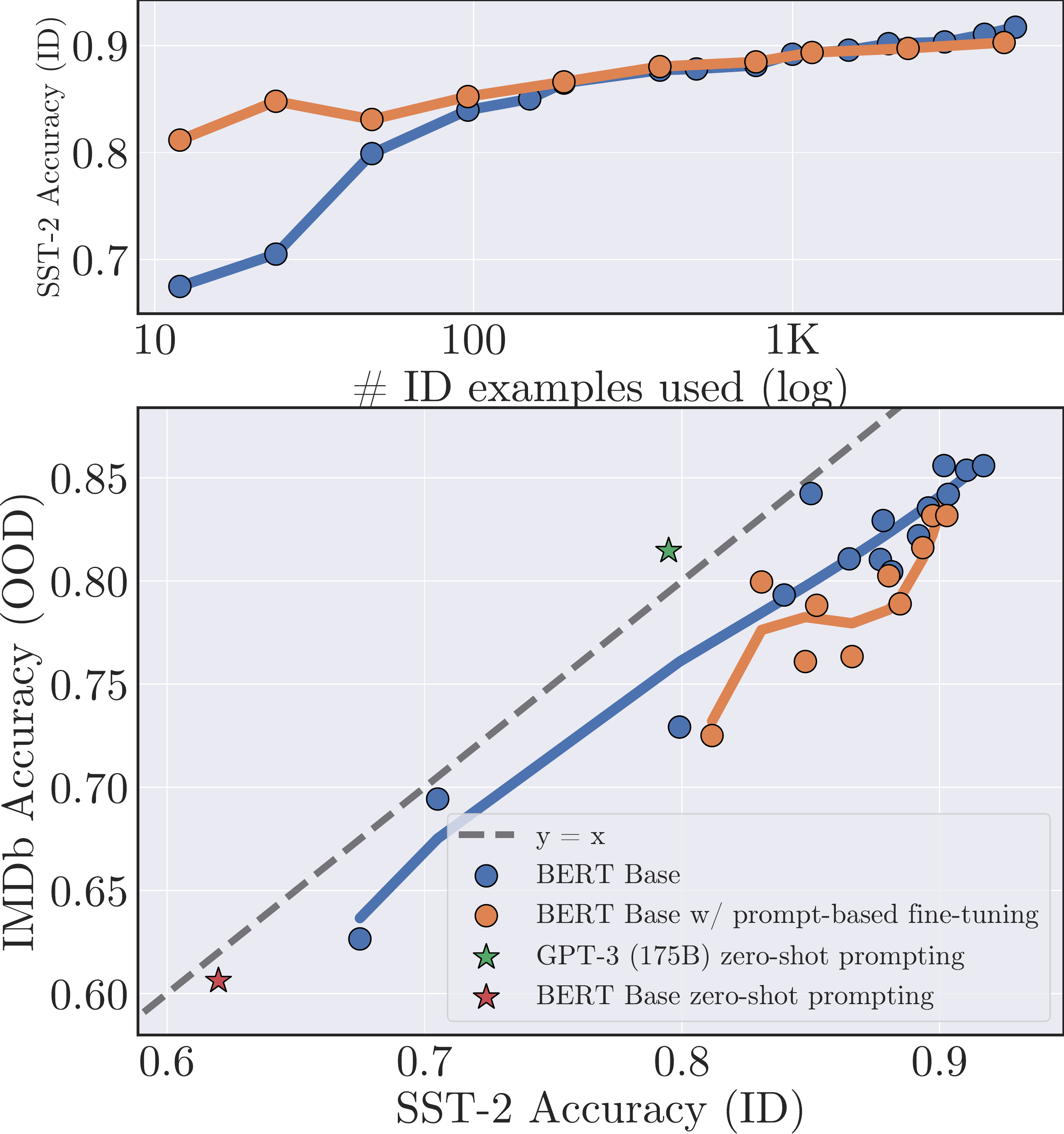}
            \caption{}
          \end{subfigure}
        \hfill
        \begin{subfigure}{0.32\linewidth}
            \centering
            \includegraphics[width=\linewidth]{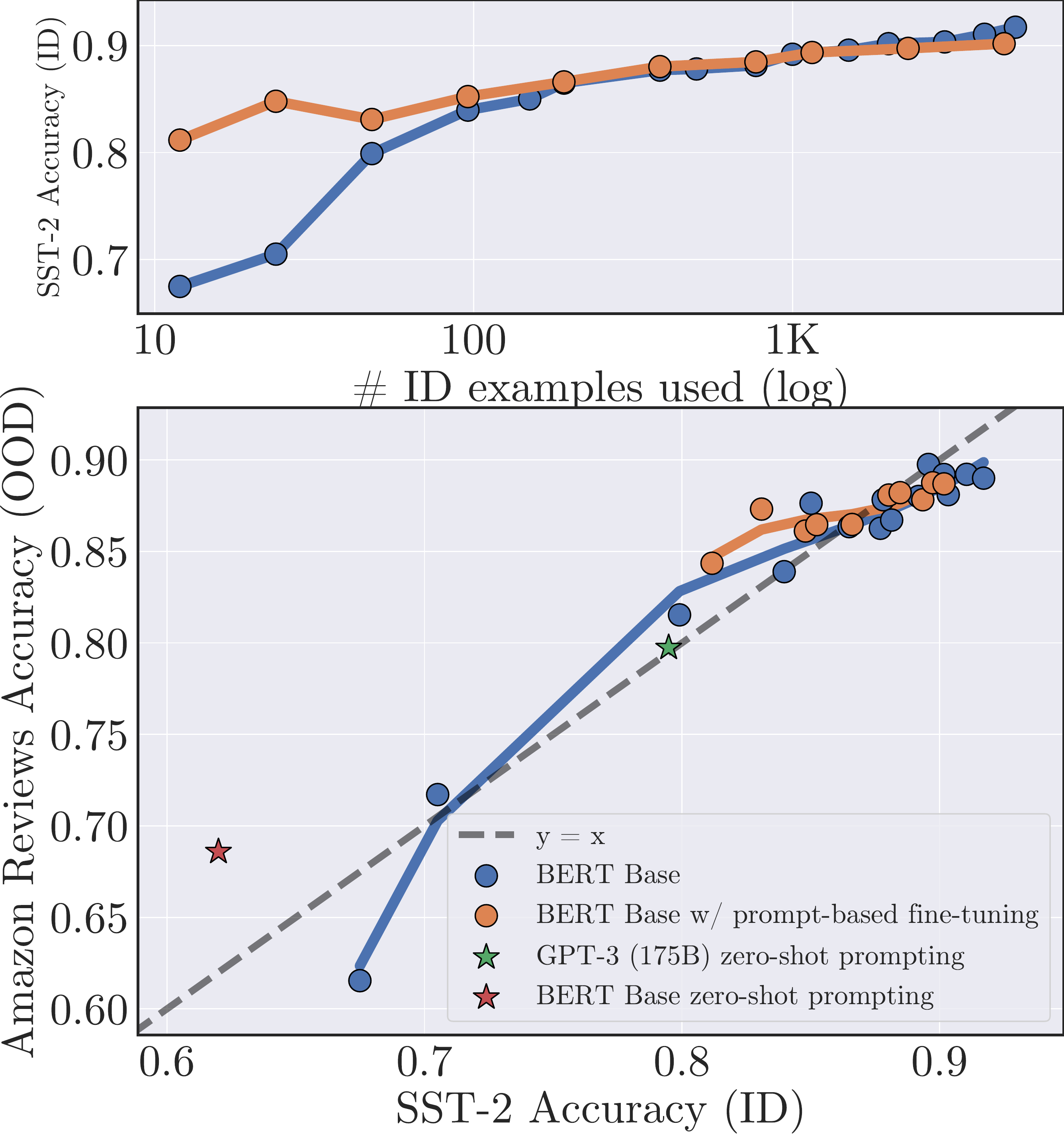}
            \caption{}
        \end{subfigure}
        \hspace*{\fill}
        \caption{Results on all sentiment analysis ID-OOD settings when comparing zero-shot prompting, prompt-based fine-tuning, and standard fine-tuning.}
    \end{figure*}

\clearpage
\subsection{Increasing Pre-Trained Model Size}\label{app:full_results_increasing_model_size}

\begin{figure*}[!h]
        \centering
        \begin{subfigure}{0.32\linewidth}
            \centering
            \includegraphics[width=\linewidth]{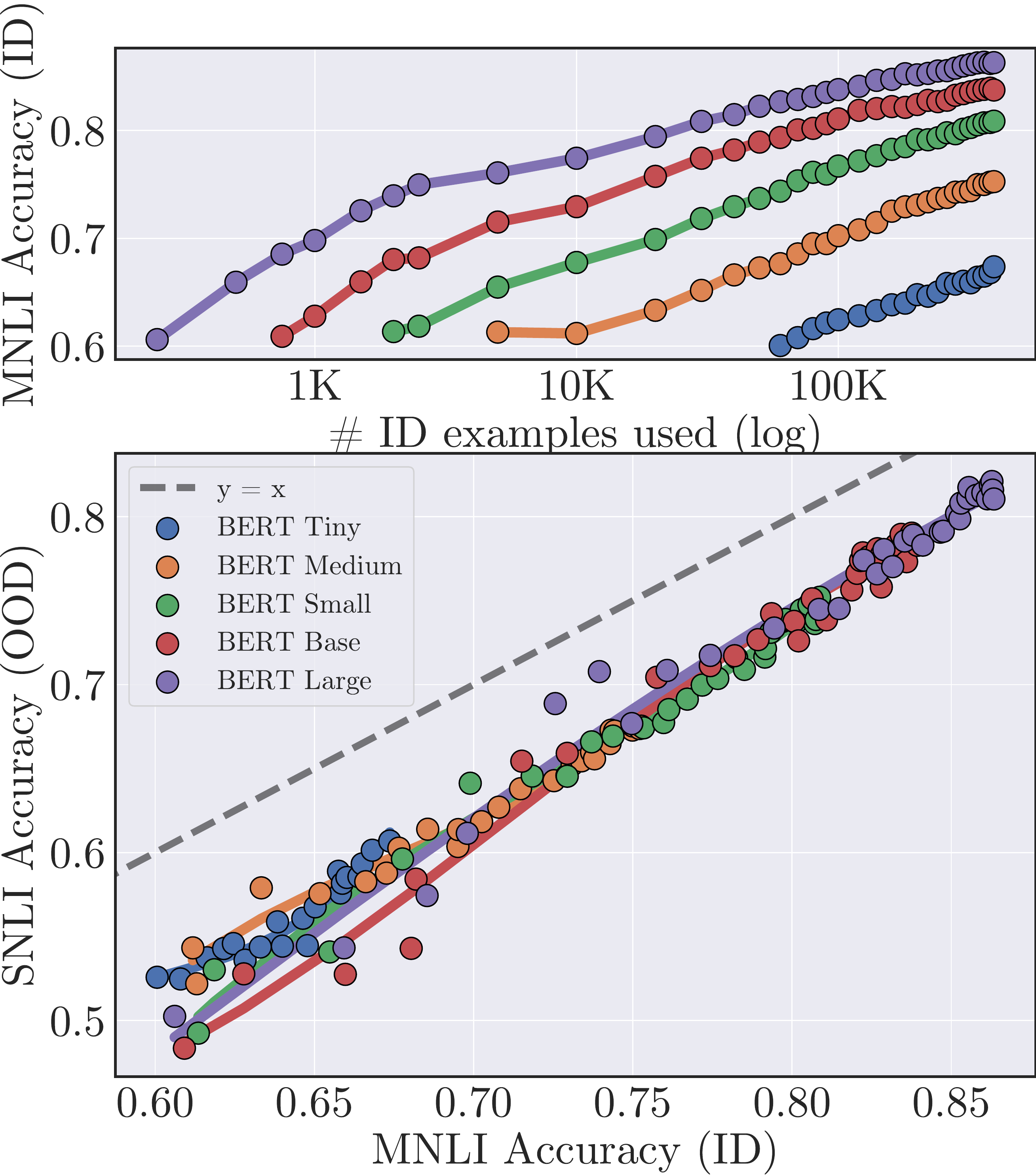}
            \caption{}
          \end{subfigure}
          \hfill
        \begin{subfigure}{0.32\linewidth}
            \centering
            \includegraphics[width=\linewidth]{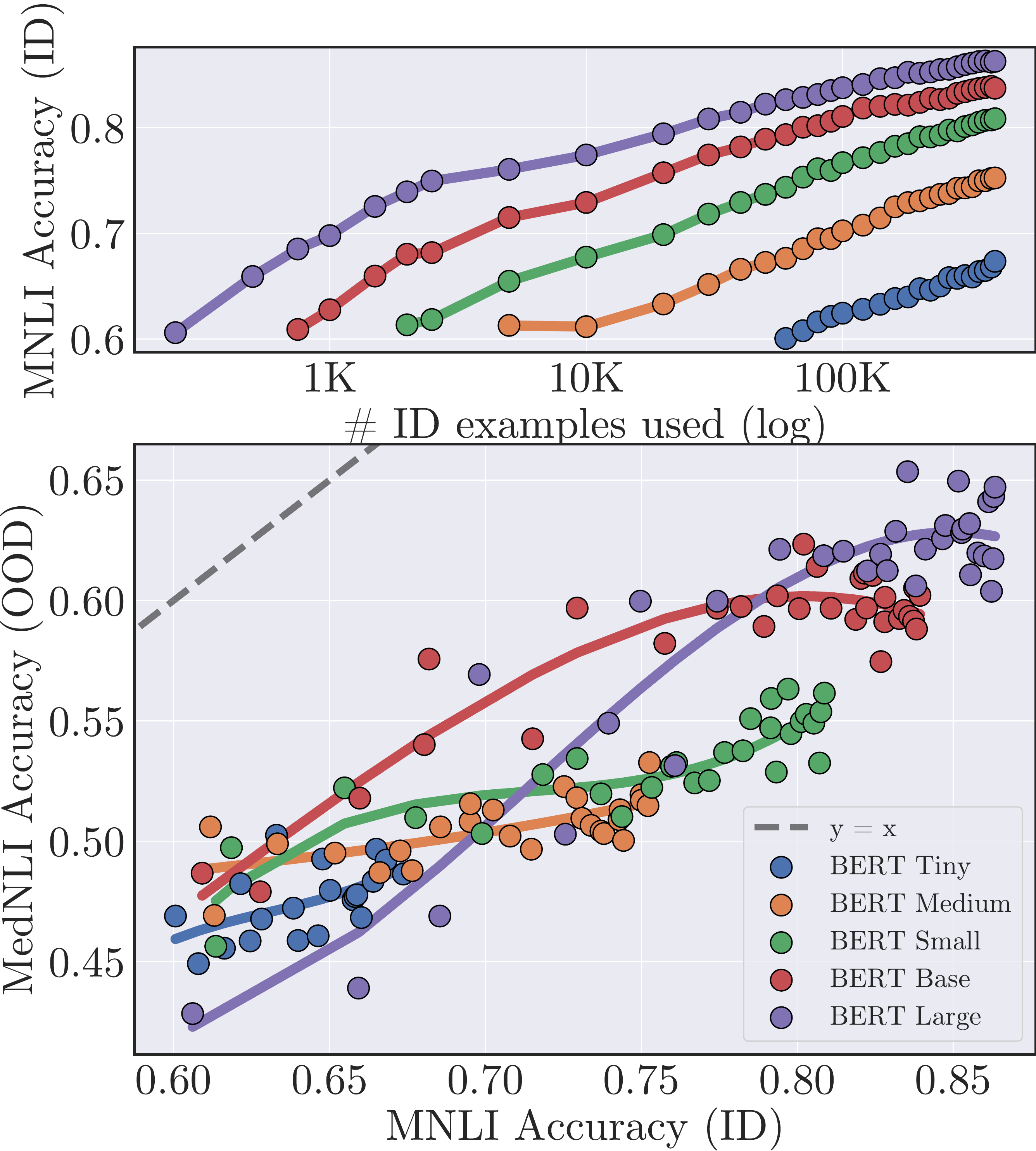}
            \caption{}
        \end{subfigure}
        \hfill
        \begin{subfigure}{0.32\linewidth}
            \centering
            \includegraphics[width=\linewidth]{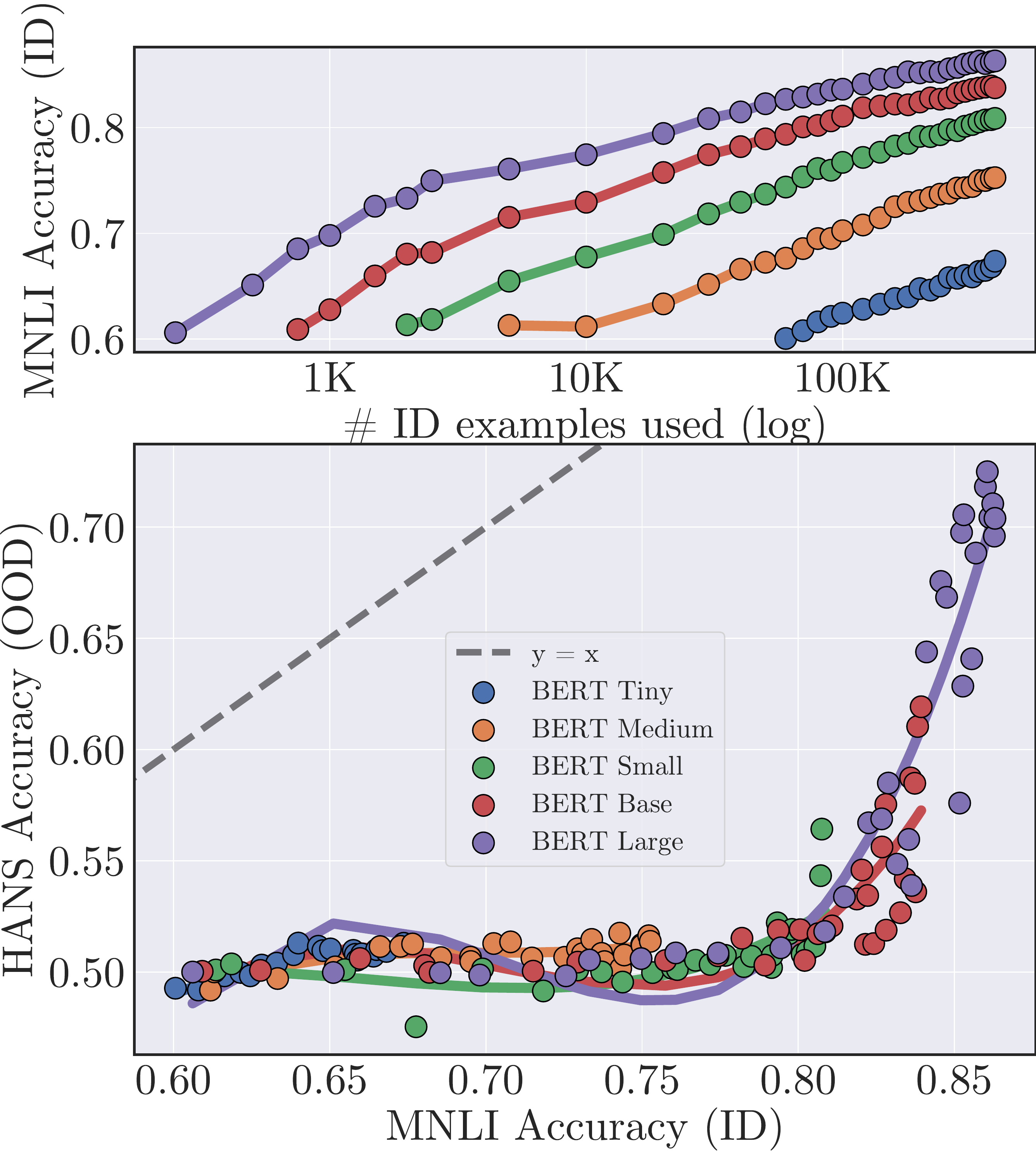}
            \caption{}
          \end{subfigure}
        \begin{subfigure}{0.32\linewidth}
            \centering
            \includegraphics[width=\linewidth]{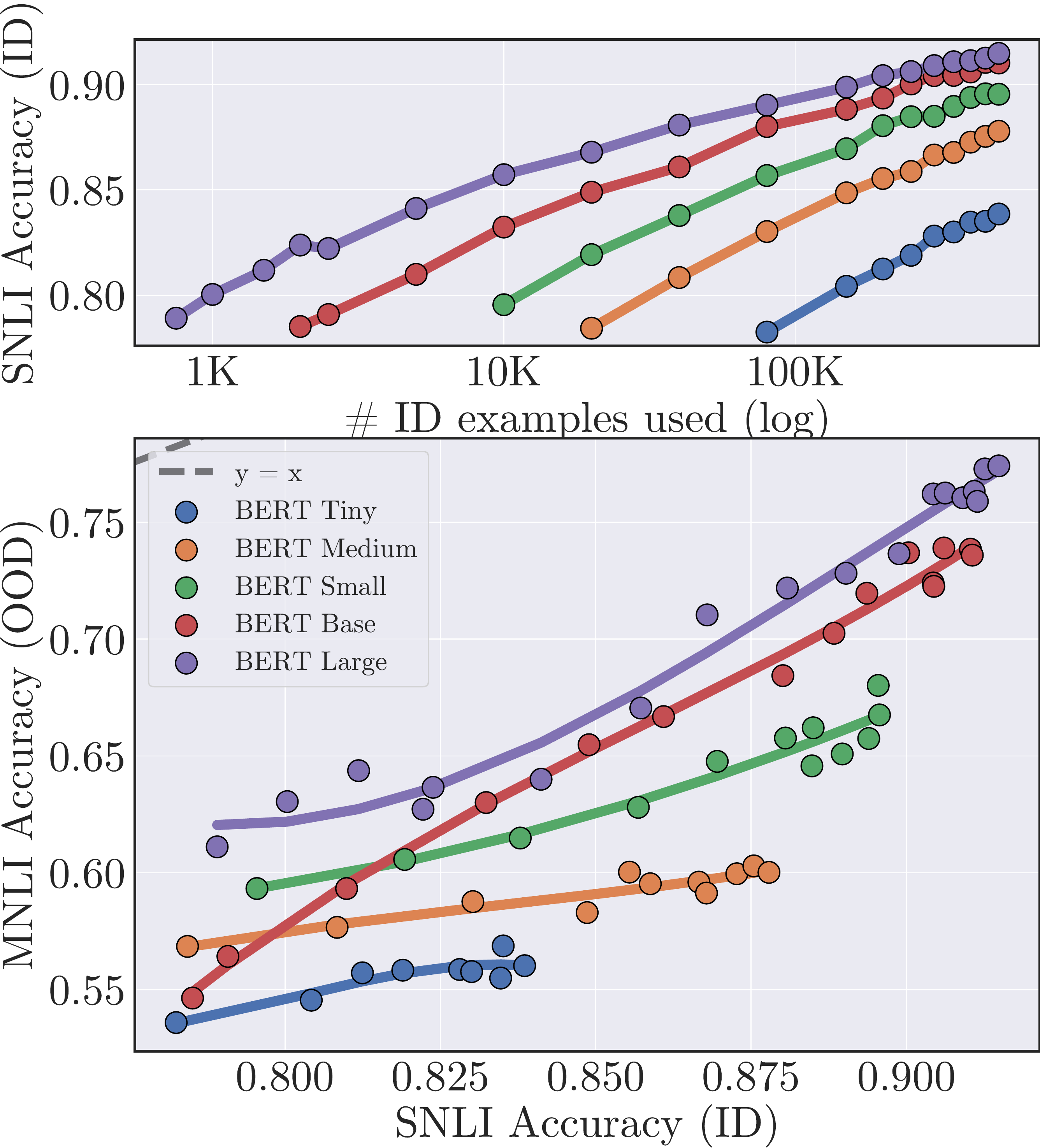}
            \caption{}
          \end{subfigure}
          \hfill
        \begin{subfigure}{0.32\linewidth}
            \centering
            \includegraphics[width=\linewidth]{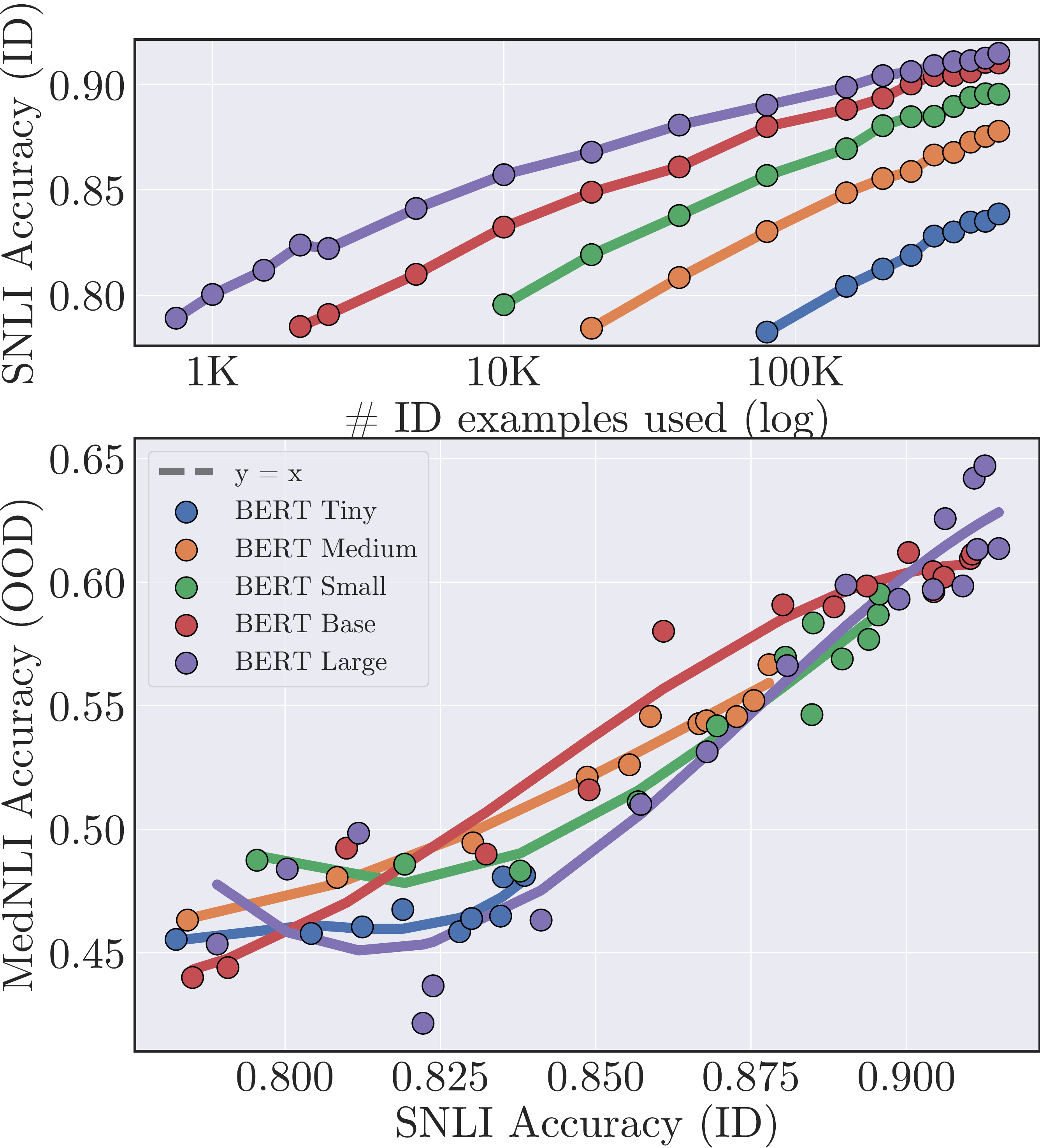}
            \caption{}
        \end{subfigure}
        \hfill
        \begin{subfigure}{0.32\linewidth}
            \centering
            \includegraphics[width=\linewidth]{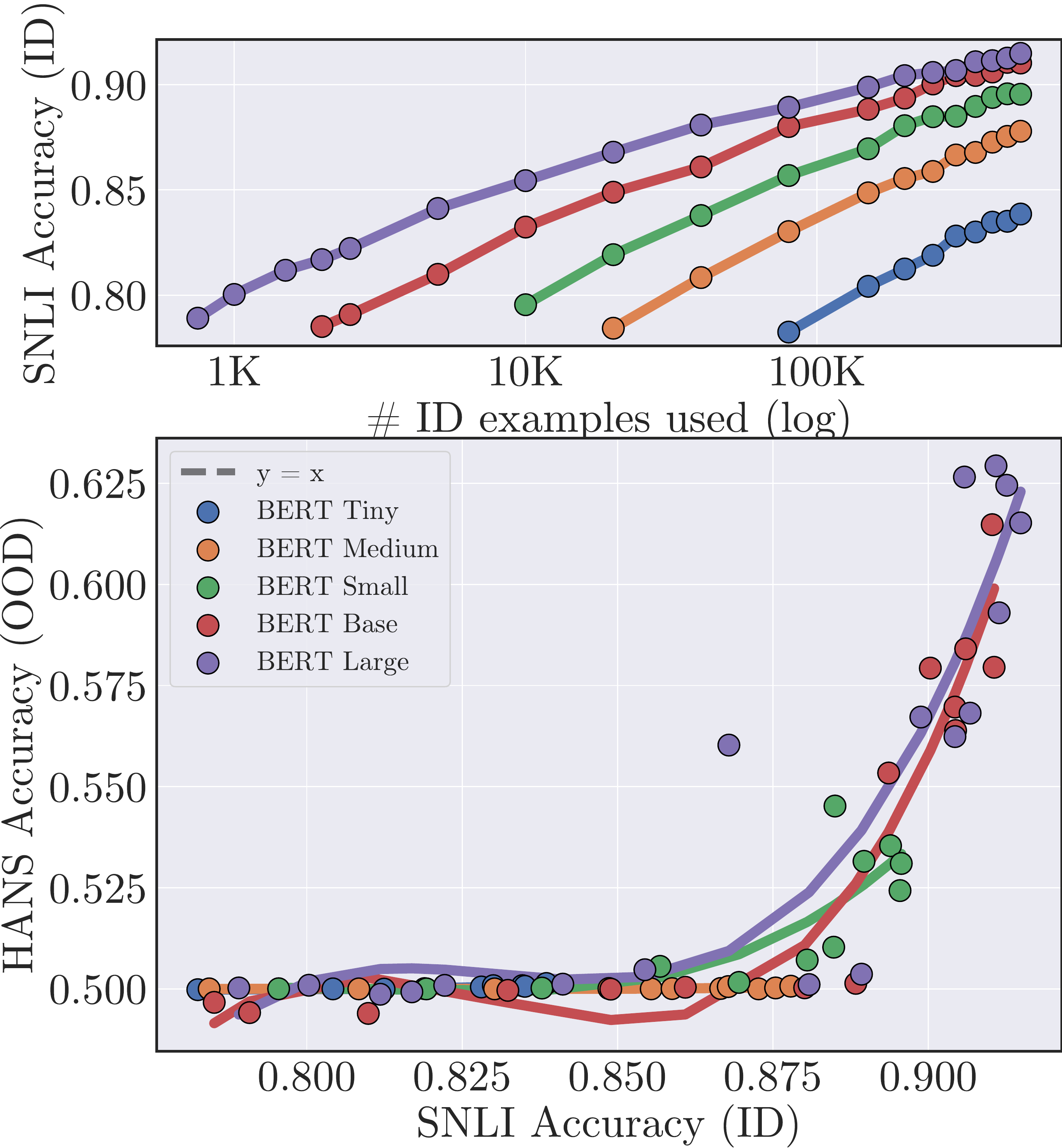}
            \caption{}
          \end{subfigure}
        \caption{Results on all NLI ID-OOD settings when increasing pre-trained model size.}
    \end{figure*}

\begin{figure*}[!h]
        \centering
        \begin{subfigure}{0.32\linewidth}
            \centering
            \includegraphics[width=\linewidth]{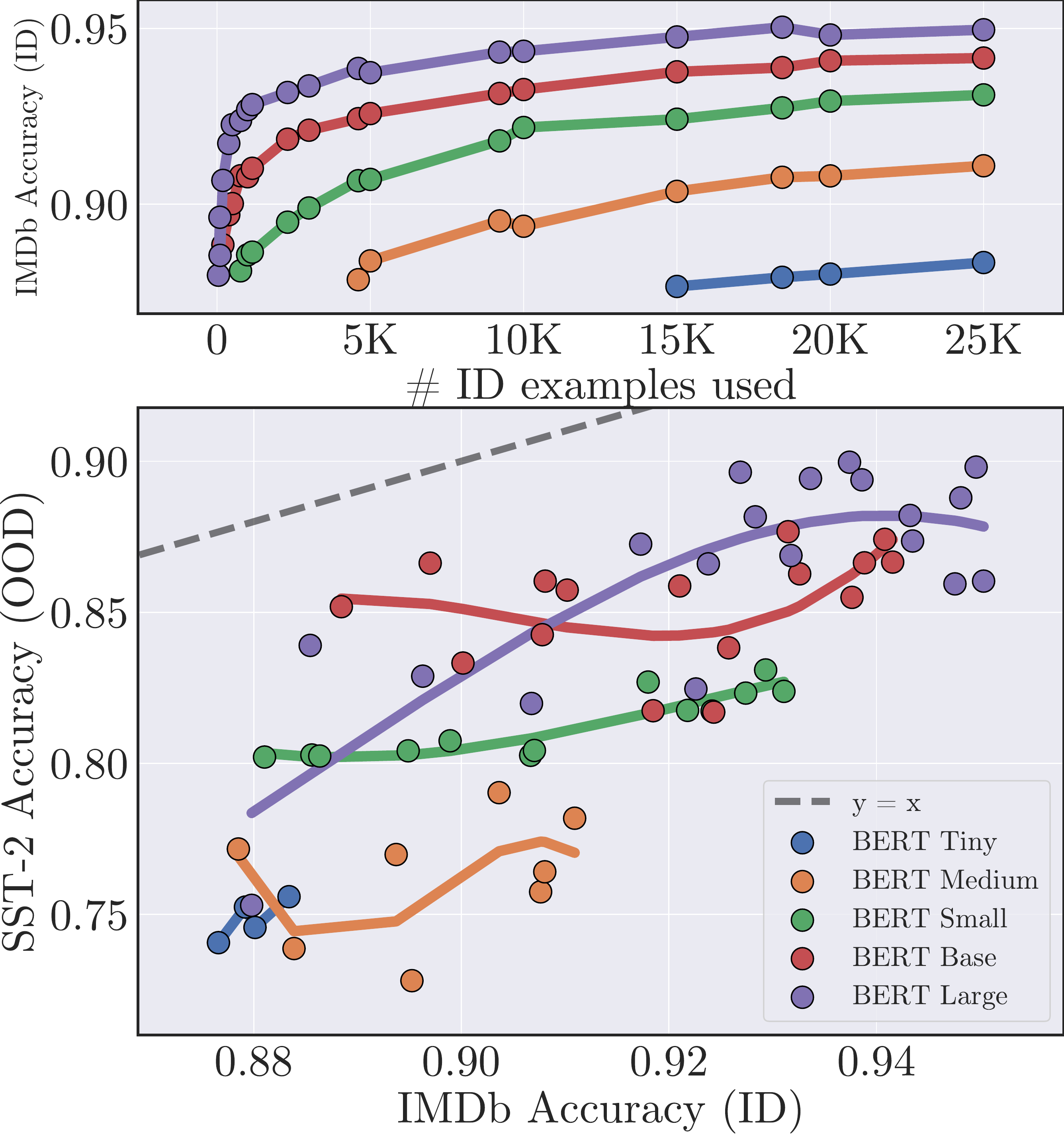}
            \caption{}
          \end{subfigure}
          \hfill
        \begin{subfigure}{0.32\linewidth}
            \centering
            \includegraphics[width=\linewidth]{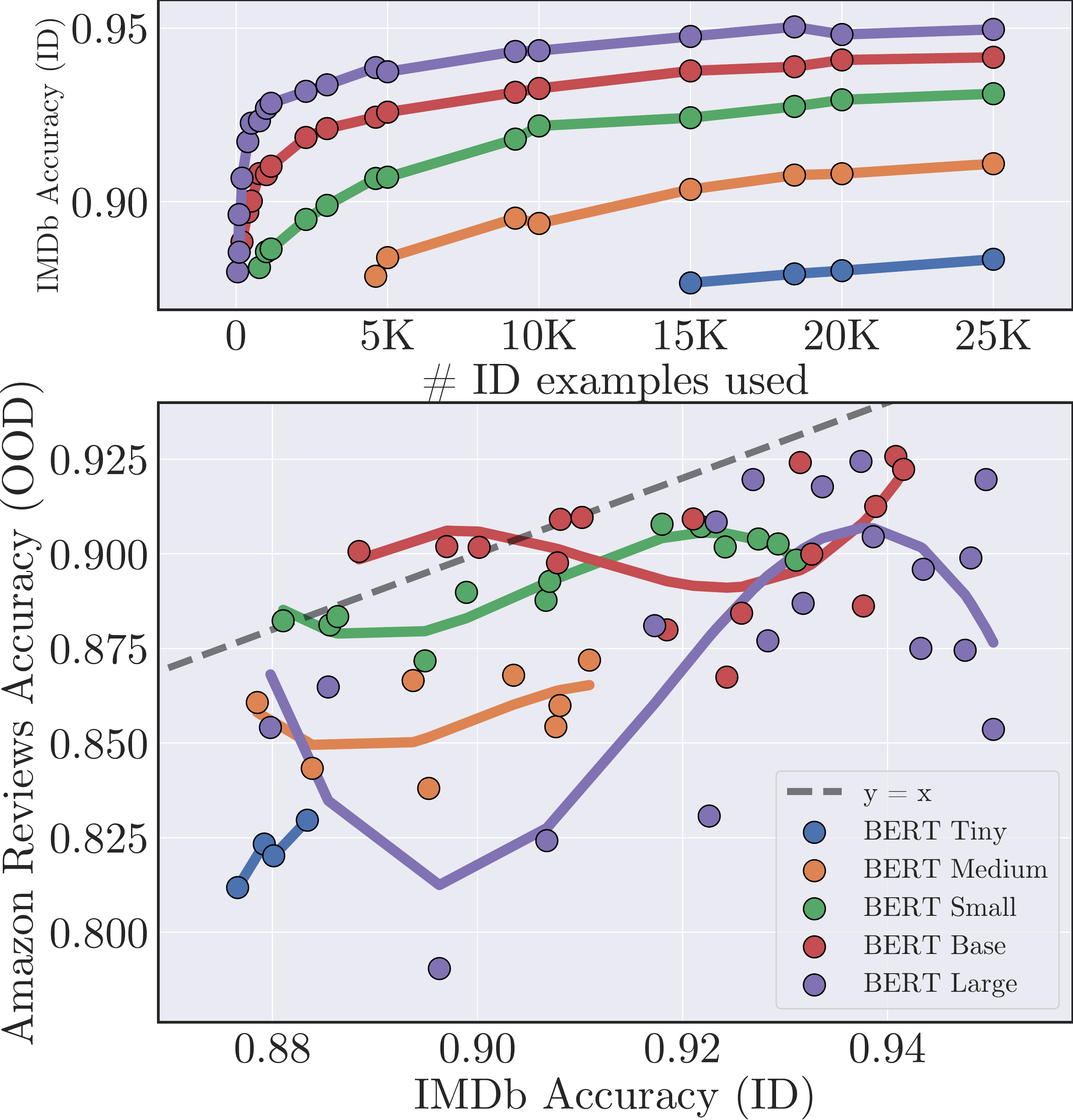}
            \caption{}
        \end{subfigure}
        \hfill
        \begin{subfigure}{0.32\linewidth}
            \centering
            \includegraphics[width=\linewidth]{figures/model_size_sst_test_vs_imdb_dev.pdf}
            \caption{}
          \end{subfigure}
        \begin{subfigure}{0.32\linewidth}
            \centering
            \includegraphics[width=\linewidth]{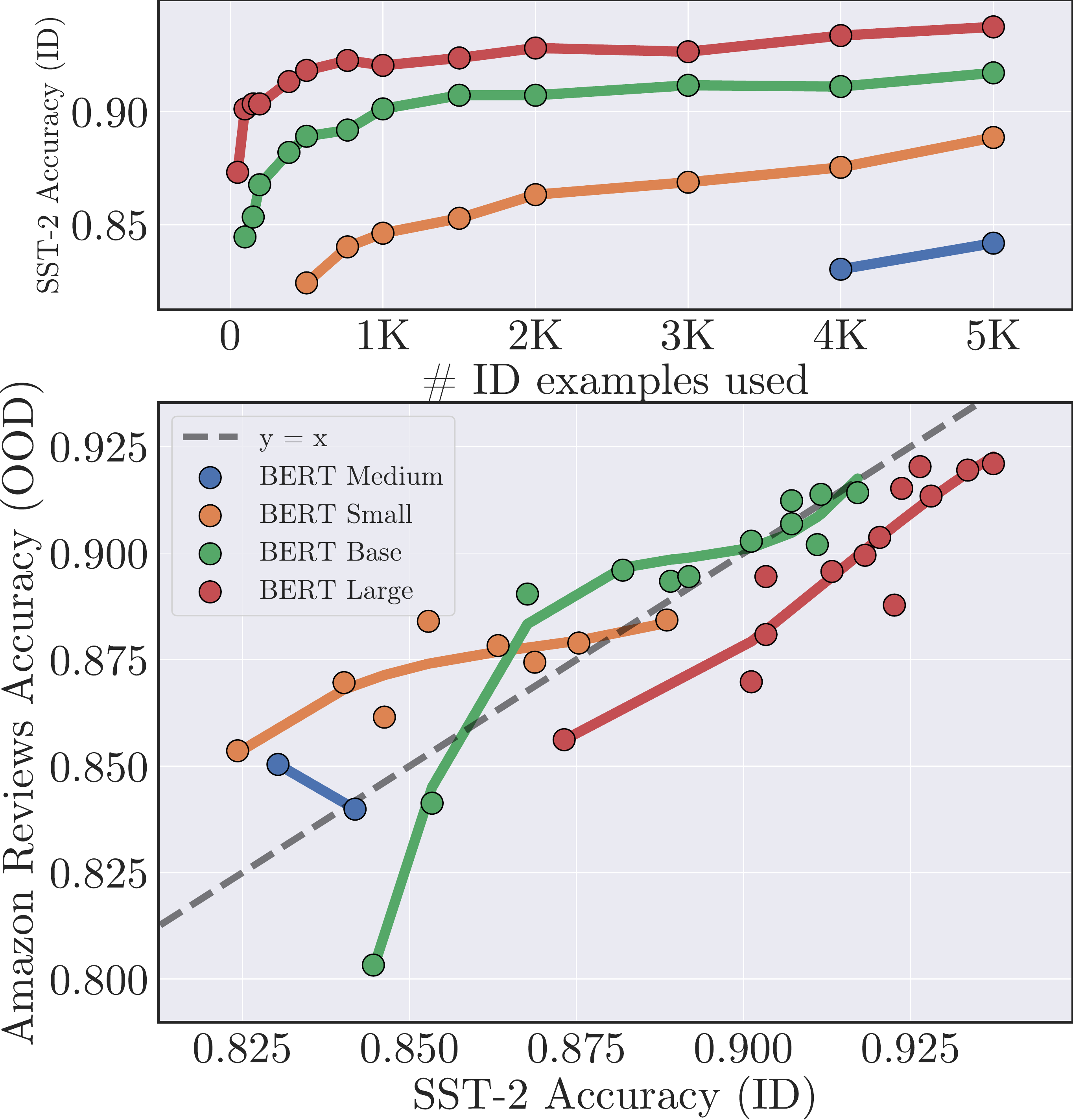}
            \caption{}
          \end{subfigure}
          \hspace*{\fill}
        \caption{Results on all sentiment analysis ID-OOD settings when increasing pre-trained model size.}
    \end{figure*}

  \begin{figure*}[!h]
        \centering
        \begin{subfigure}{0.32\linewidth}
            \centering
            \includegraphics[width=\linewidth]{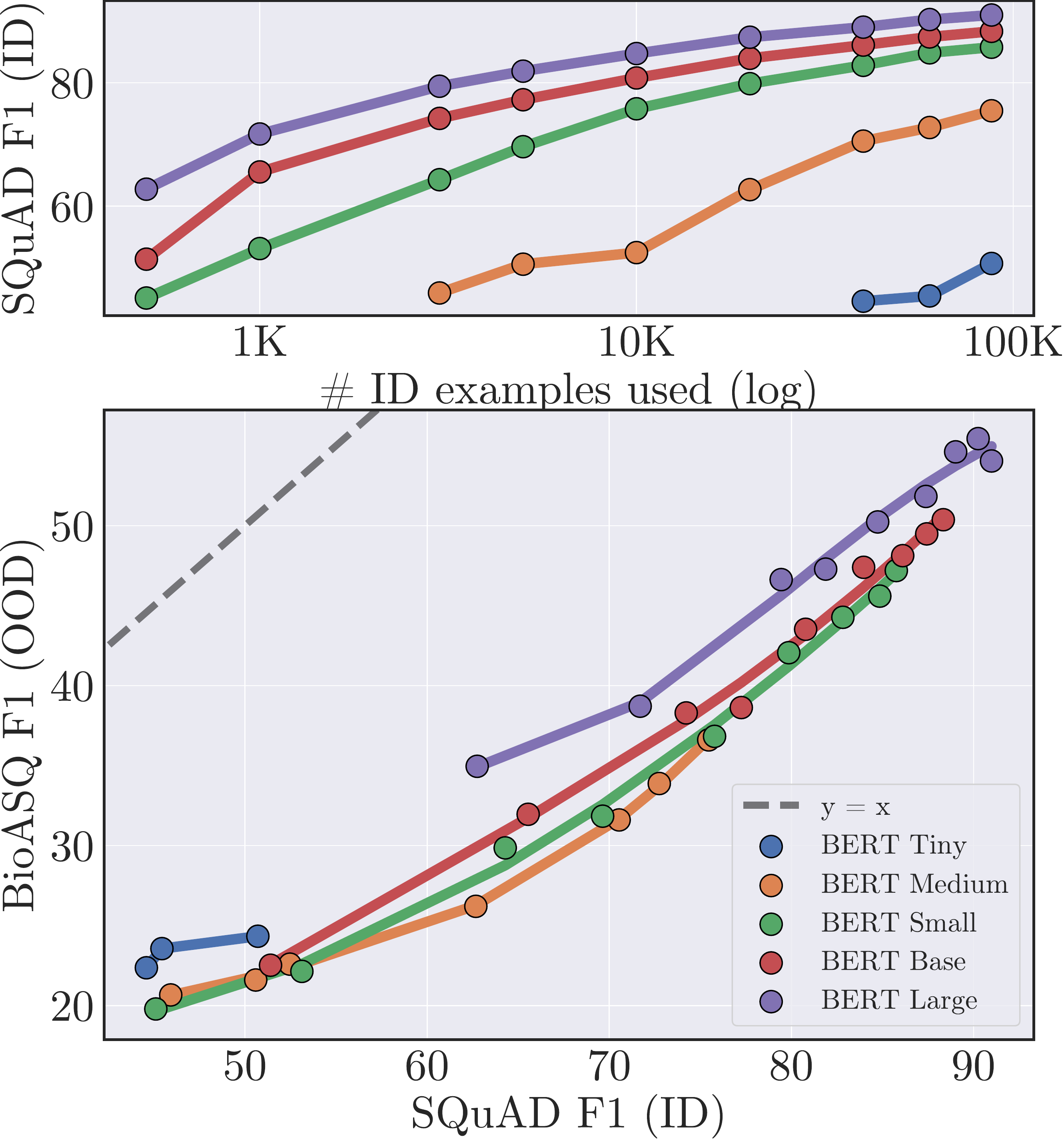}
            \caption{}
          \end{subfigure}
          \hfill
        \begin{subfigure}{0.32\linewidth}
            \centering
            \includegraphics[width=\linewidth]{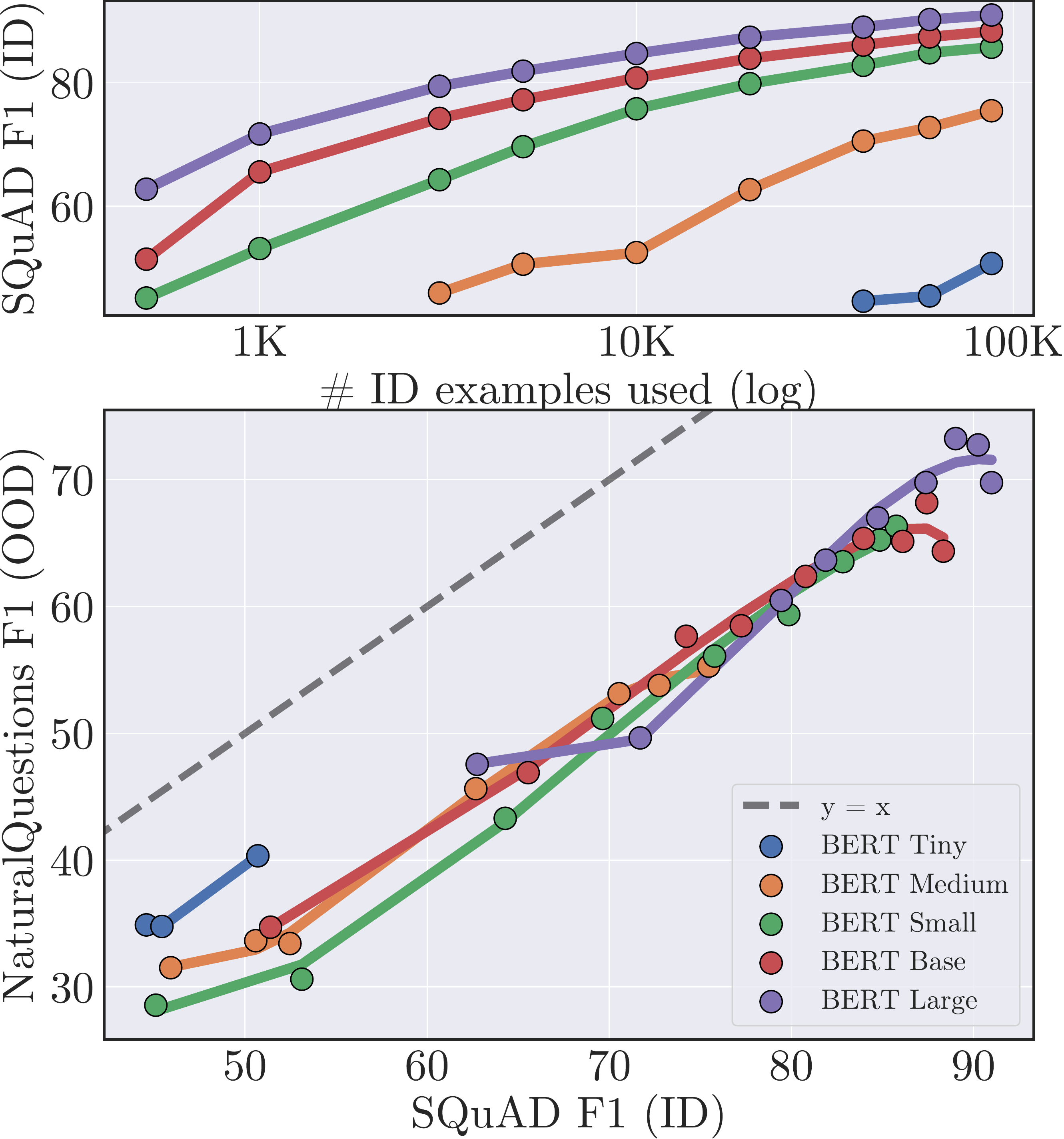}
            \caption{}
        \end{subfigure}
        \hfill
        \begin{subfigure}{0.32\linewidth}
            \centering
            \includegraphics[width=\linewidth]{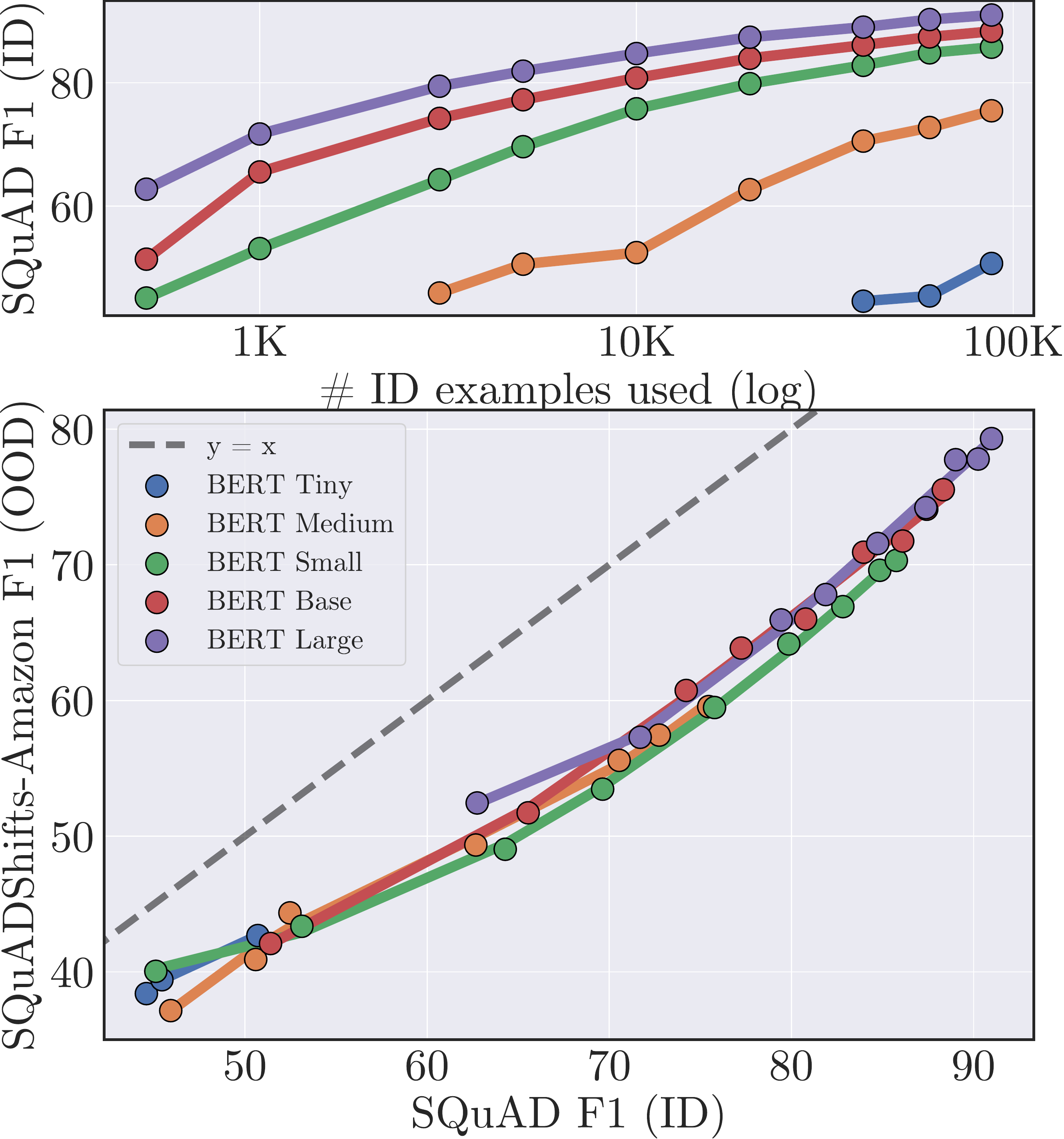}
            \caption{}
          \end{subfigure}
        \begin{subfigure}{0.32\linewidth}
            \centering
            \includegraphics[width=\linewidth]{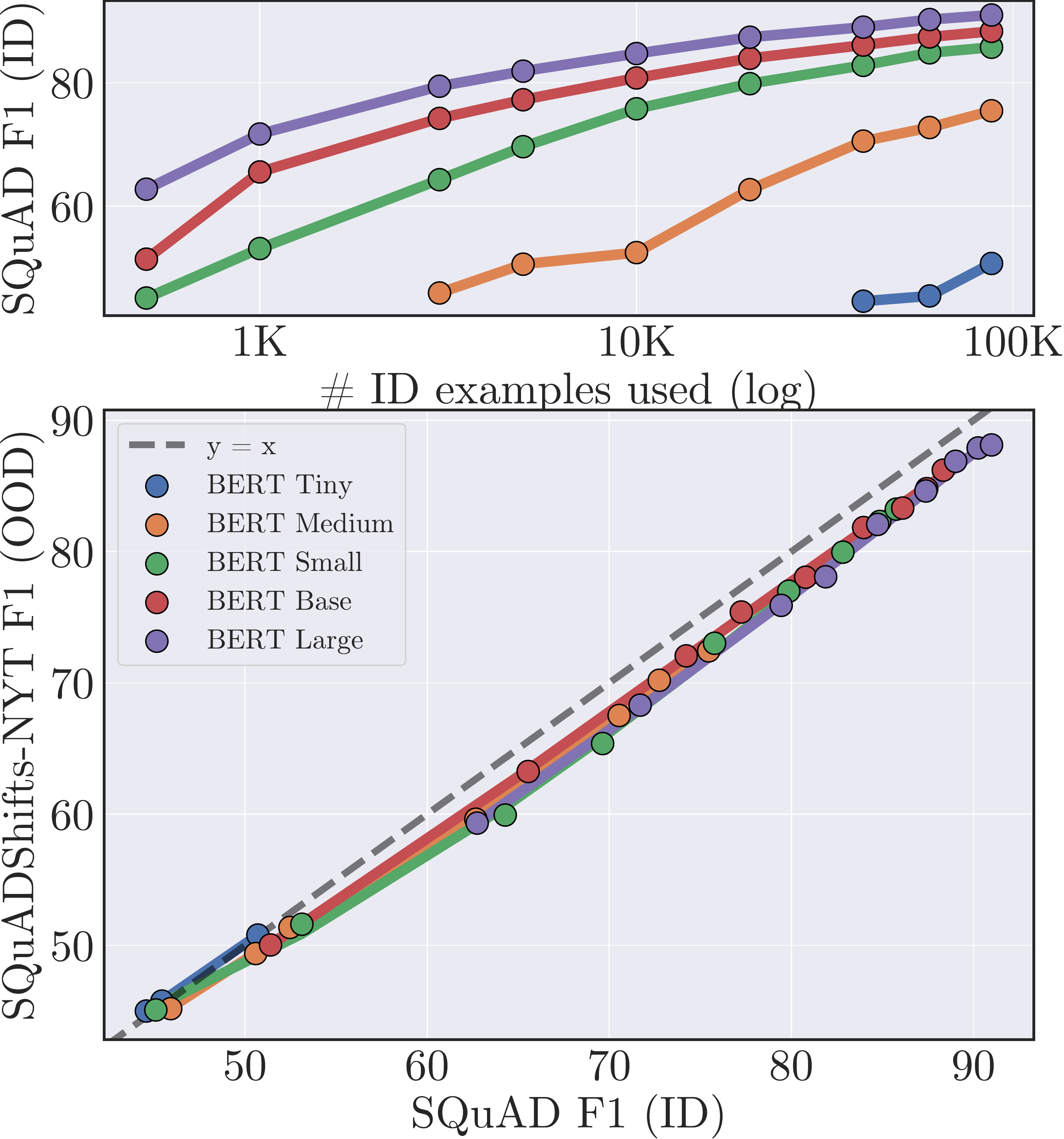}
            \caption{}
          \end{subfigure}
          \hfill
        \begin{subfigure}{0.32\linewidth}
            \centering
            \includegraphics[width=\linewidth]{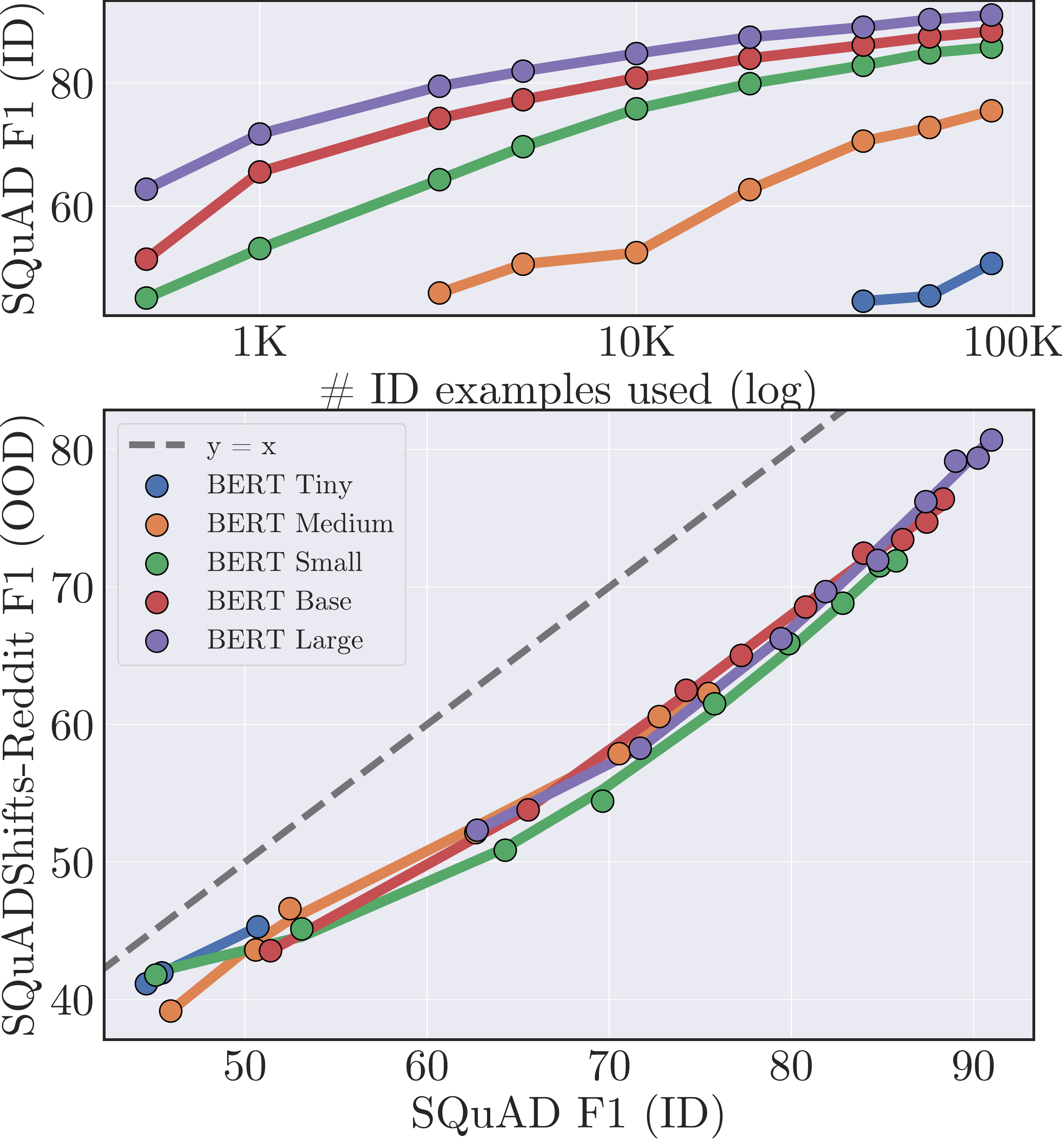}
            \caption{}
        \end{subfigure}
        \hfill
        \begin{subfigure}{0.32\linewidth}
            \centering
            \includegraphics[width=\linewidth]{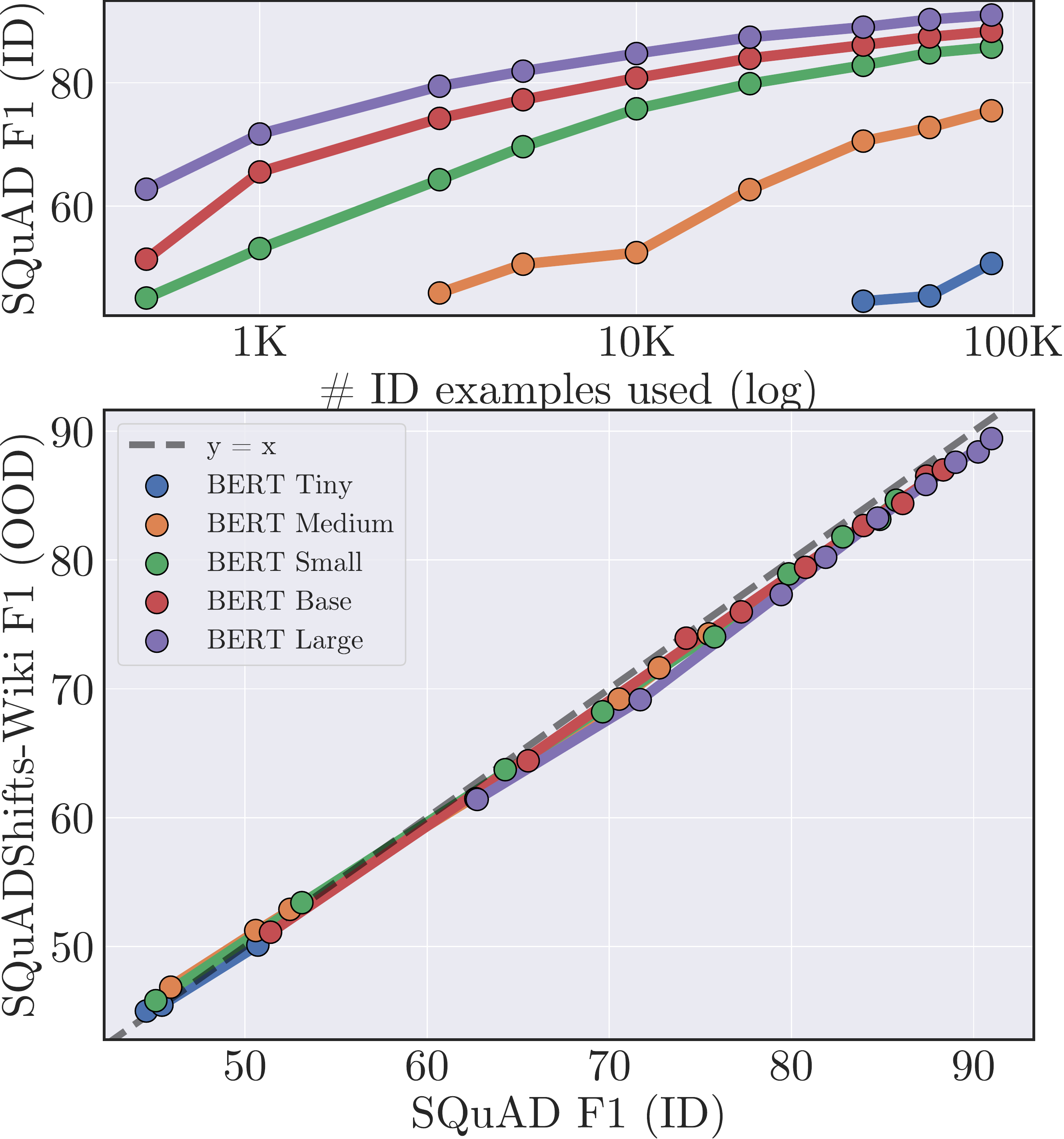}
            \caption{}
          \end{subfigure}
        \begin{subfigure}{0.32\linewidth}
            \centering
            \includegraphics[width=\linewidth]{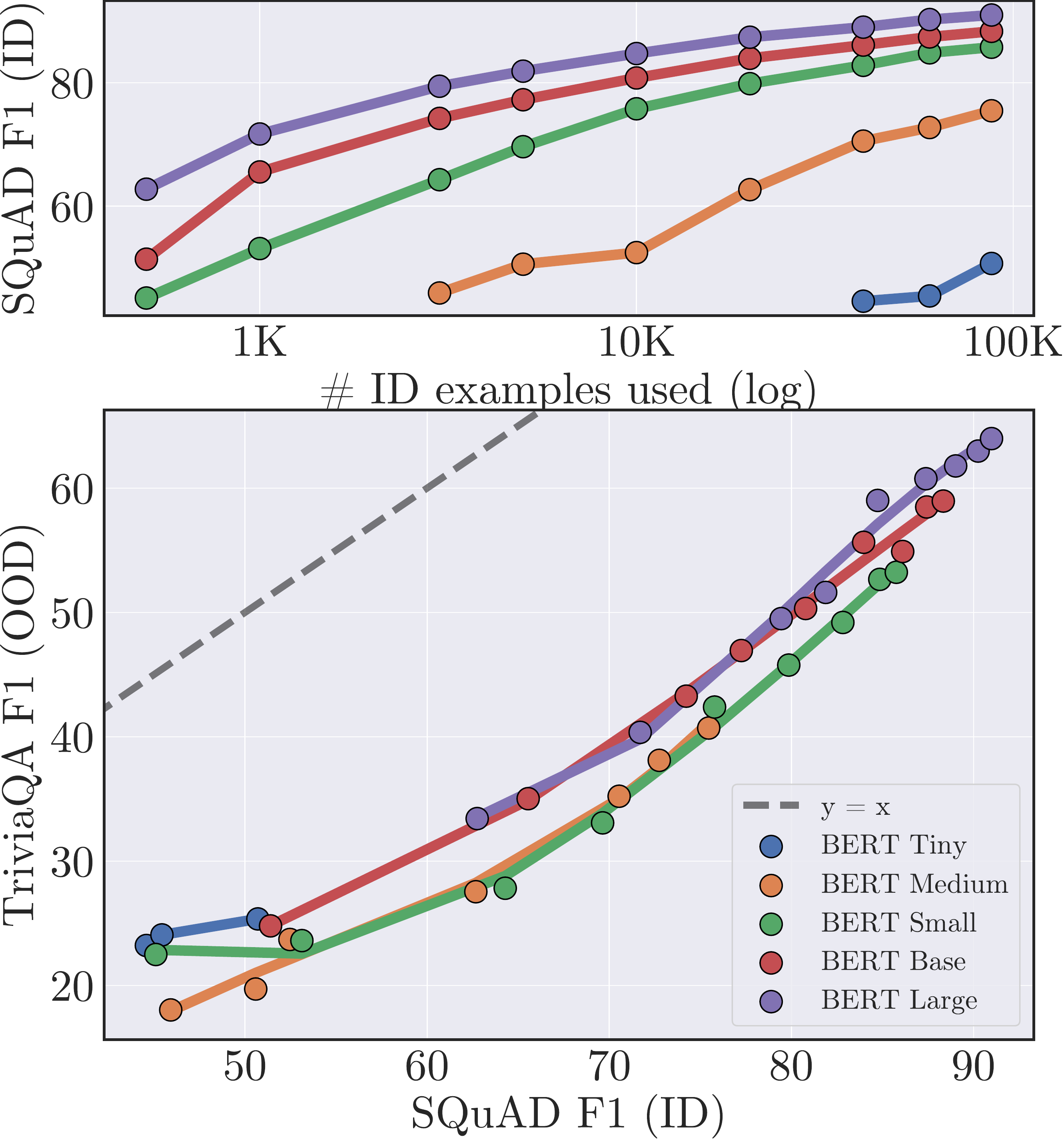}
            \caption{}
          \end{subfigure}
          \hspace*{\fill}
        \caption{Results on all extractive QA OOD settings when training on SQuAD with pre-trained models of increasing size.}
    \end{figure*}

  \begin{figure*}[!h]
        \centering
        \begin{subfigure}{0.32\linewidth}
            \centering
            \includegraphics[width=\linewidth]{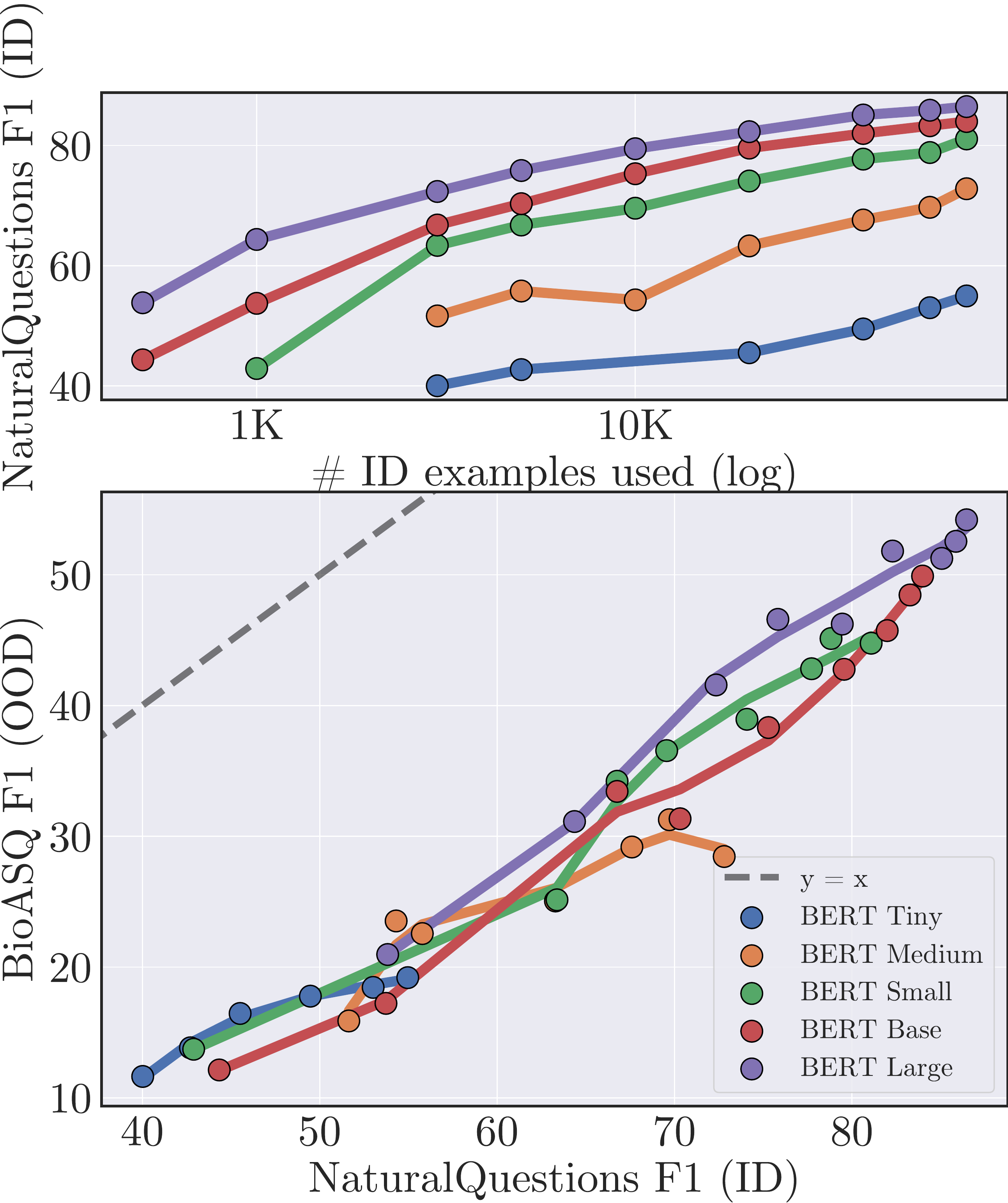}
            \caption{}
          \end{subfigure}
          \hfill
        \begin{subfigure}{0.32\linewidth}
            \centering
            \includegraphics[width=\linewidth]{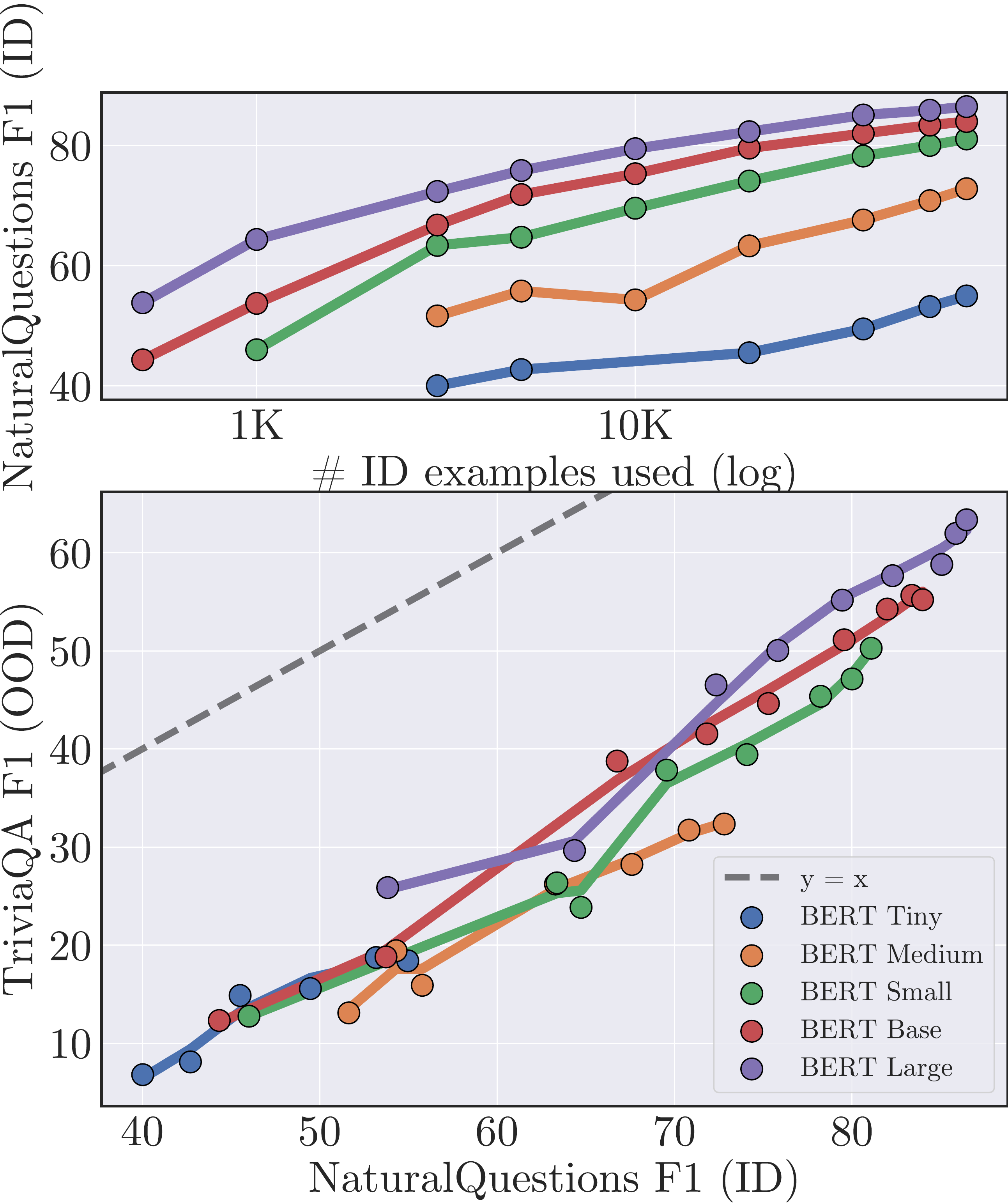}
            \caption{}
          \end{subfigure}
        \hfill
        \begin{subfigure}{0.32\linewidth}
            \centering
            \includegraphics[width=\linewidth]{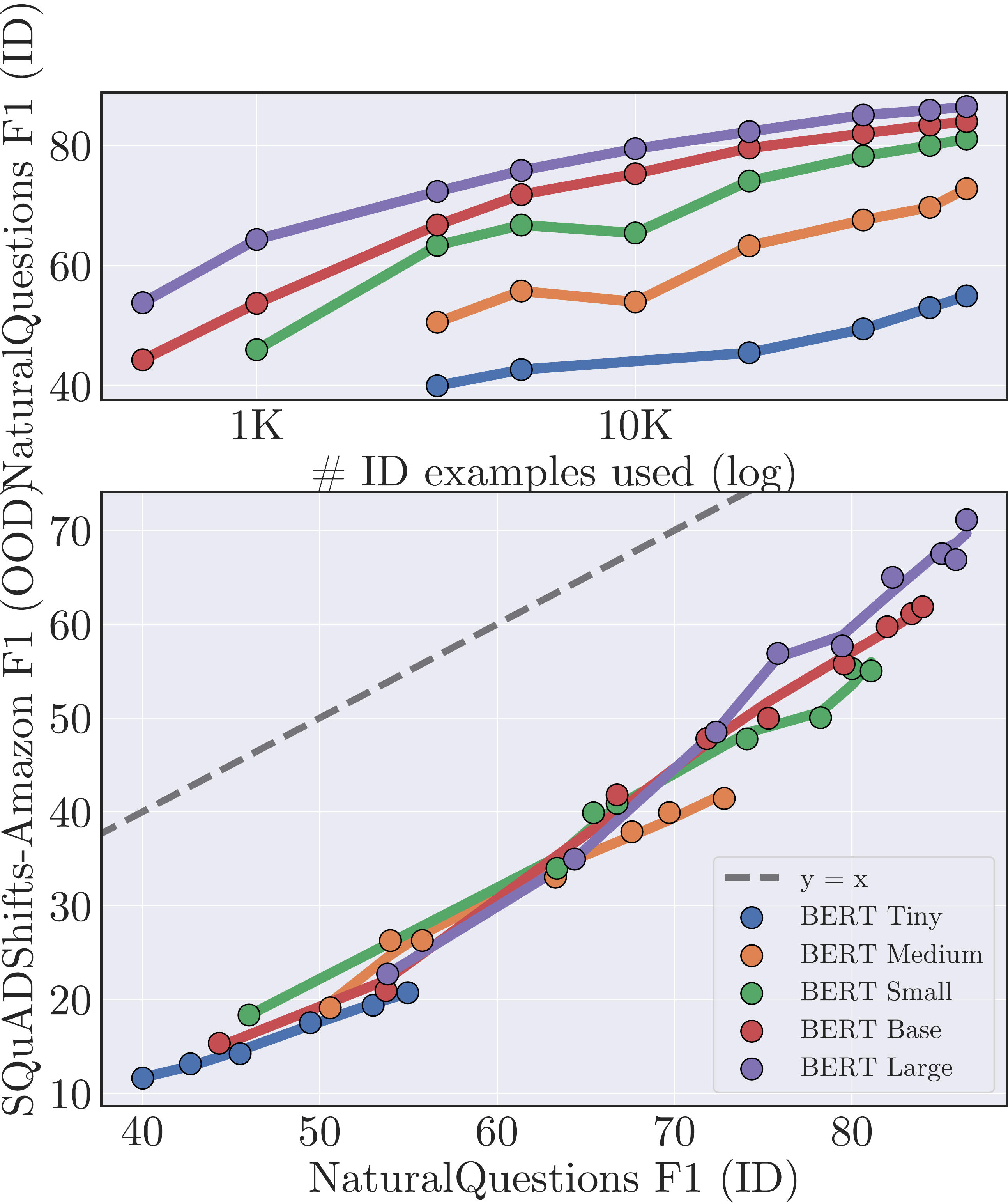}
            \caption{}
          \end{subfigure}
        \begin{subfigure}{0.32\linewidth}
            \centering
            \includegraphics[width=\linewidth]{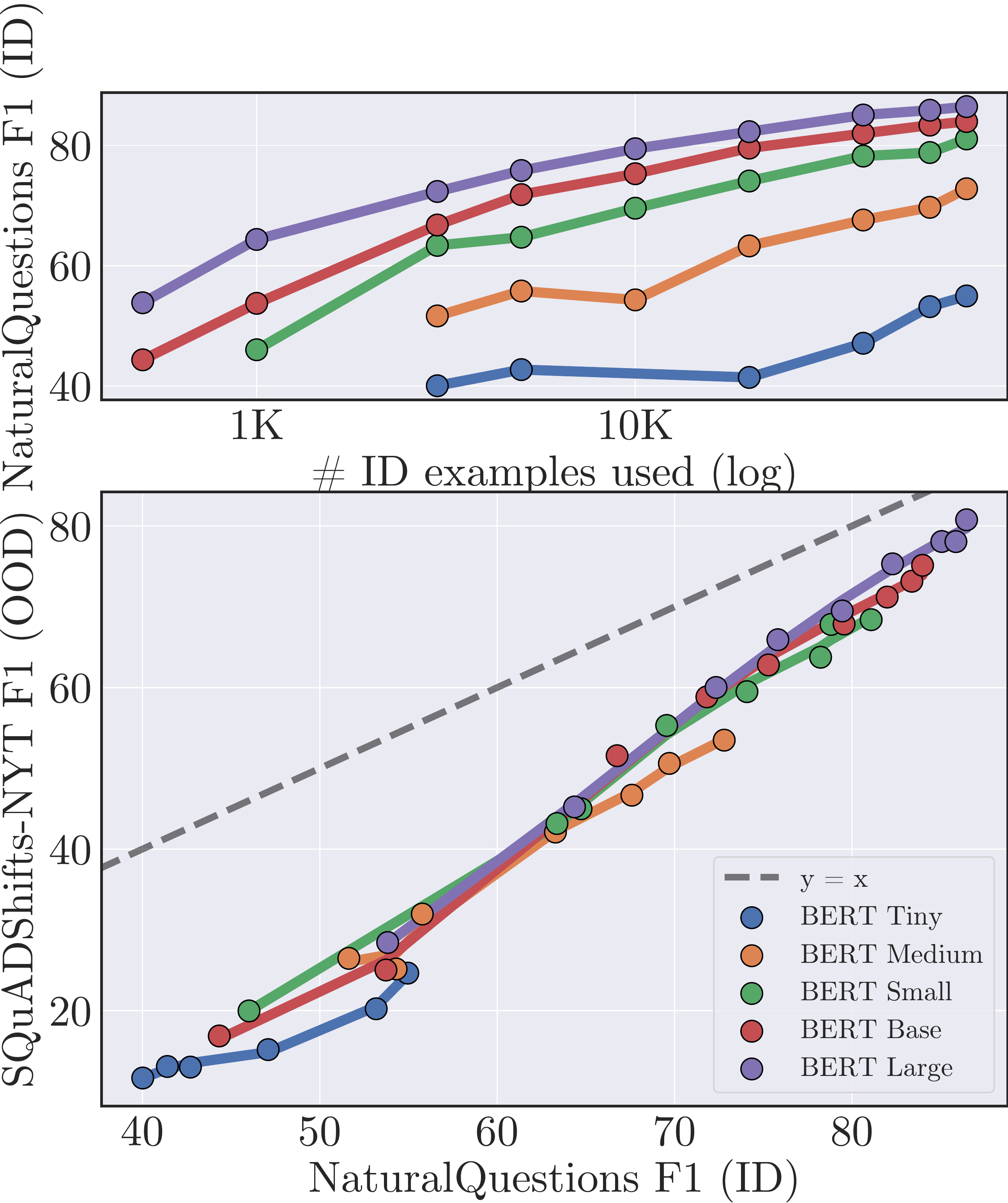}
            \caption{}
          \end{subfigure}
          \hfill
        \begin{subfigure}{0.32\linewidth}
            \centering
            \includegraphics[width=\linewidth]{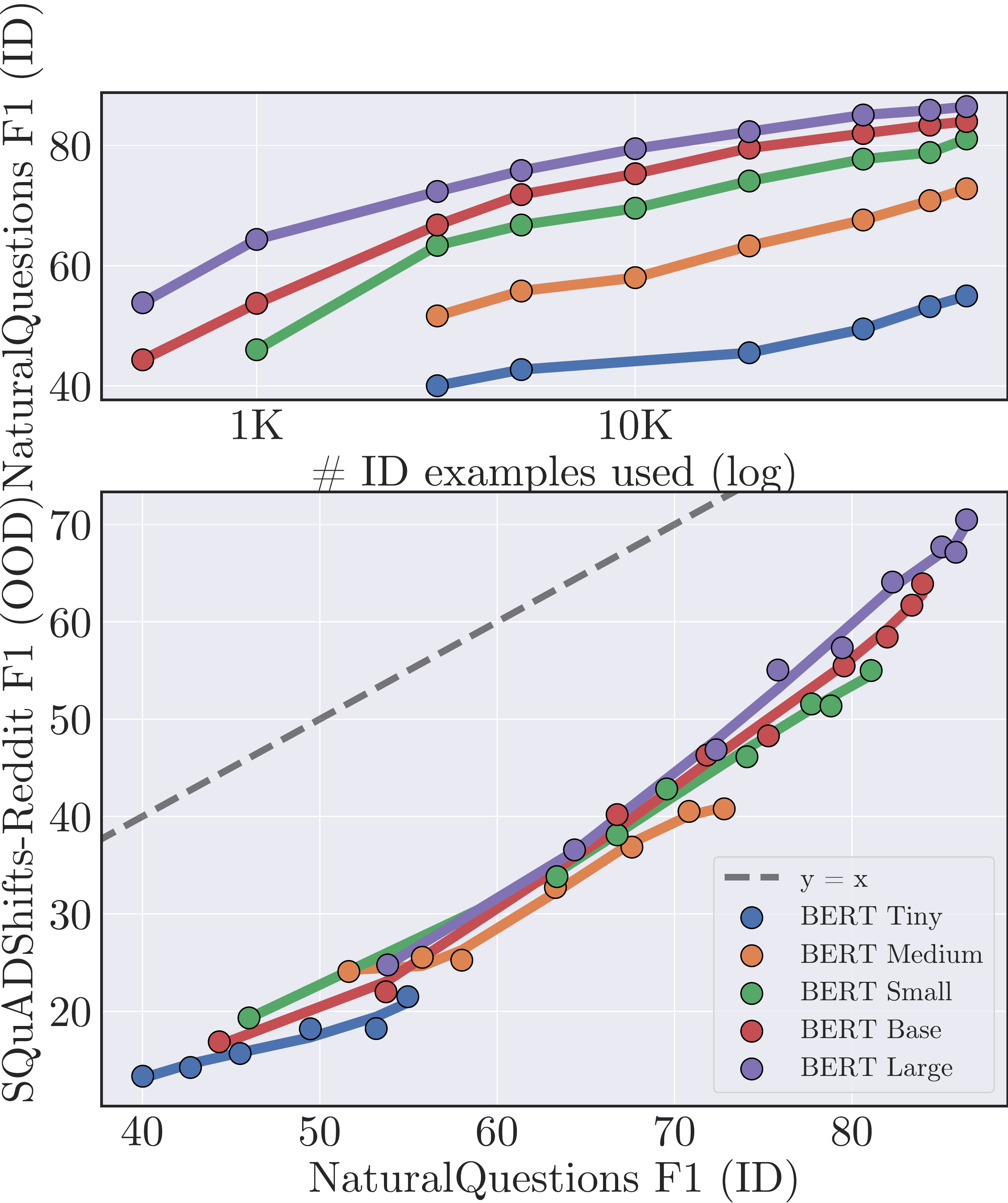}
            \caption{}
        \end{subfigure}
        \hfill
        \begin{subfigure}{0.32\linewidth}
            \centering
            \includegraphics[width=\linewidth]{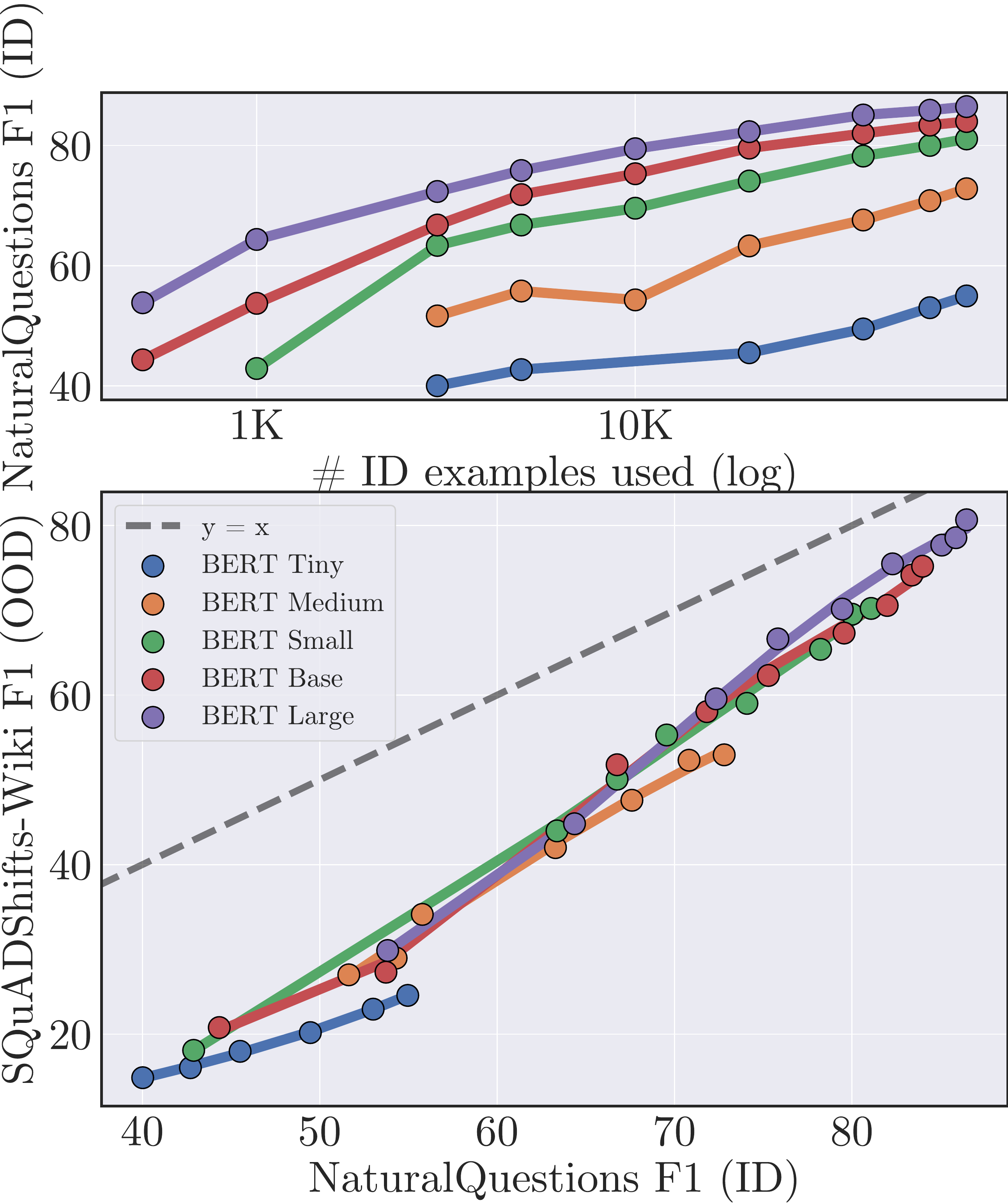}
            \caption{}
          \end{subfigure}

        \caption{Results on all extractive QA OOD settings when training on NaturalQuestions with pre-trained models of increasing size.}
    \end{figure*}

\clearpage
\subsection{Pre-Training on More Data}\label{app:full_results_pretraining_on_more_data}

\begin{figure*}[!h]
        \centering
        \begin{subfigure}{0.32\linewidth}
            \centering
            \includegraphics[width=\linewidth]{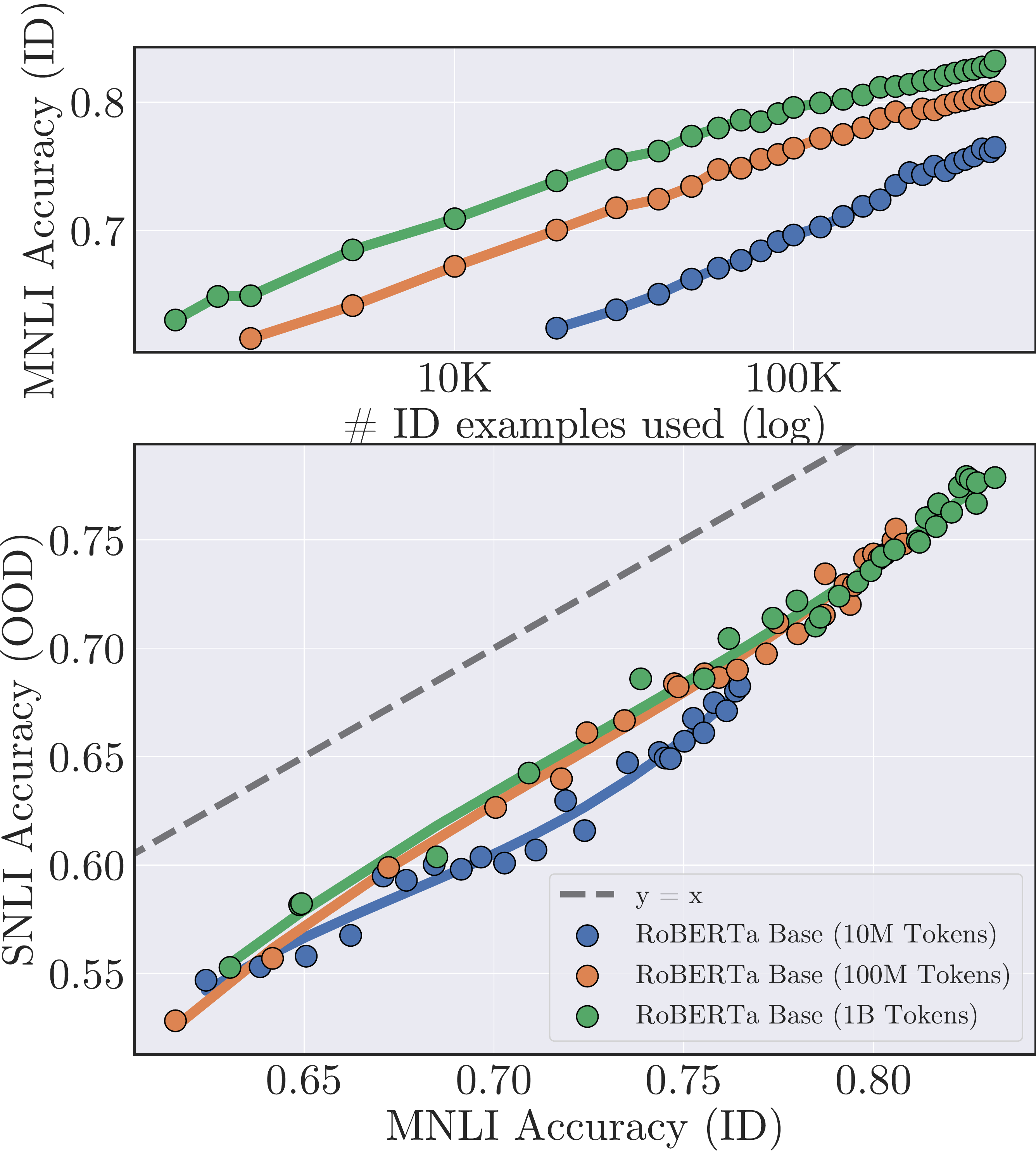}
            \caption{}
          \end{subfigure}
          \hfill
        \begin{subfigure}{0.32\linewidth}
            \centering
            \includegraphics[width=\linewidth]{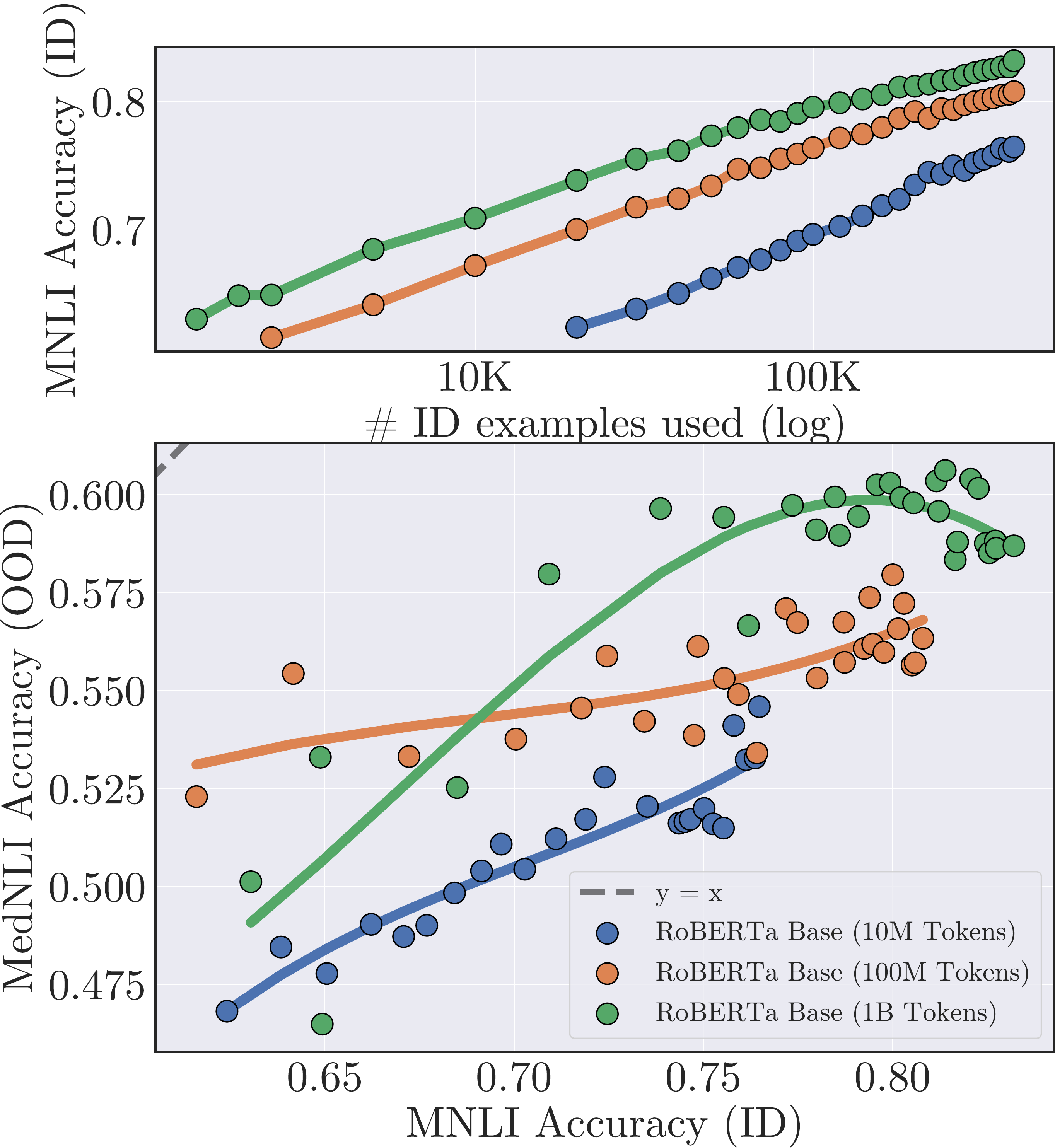}
            \caption{}
        \end{subfigure}
        \hfill
        \begin{subfigure}{0.32\linewidth}
            \centering
            \includegraphics[width=\linewidth]{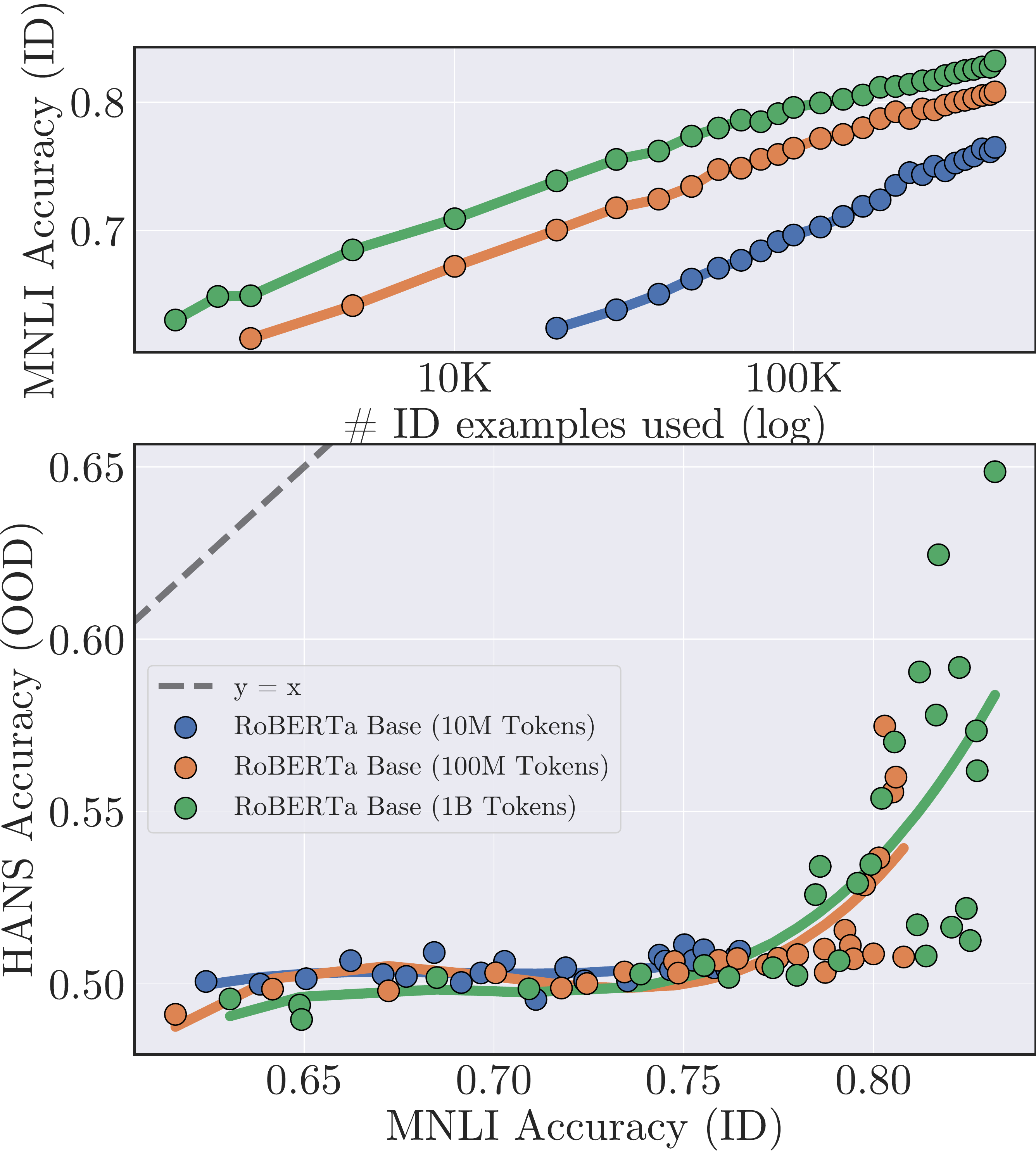}
            \caption{}
          \end{subfigure}
        \begin{subfigure}{0.32\linewidth}
            \centering
            \includegraphics[width=\linewidth]{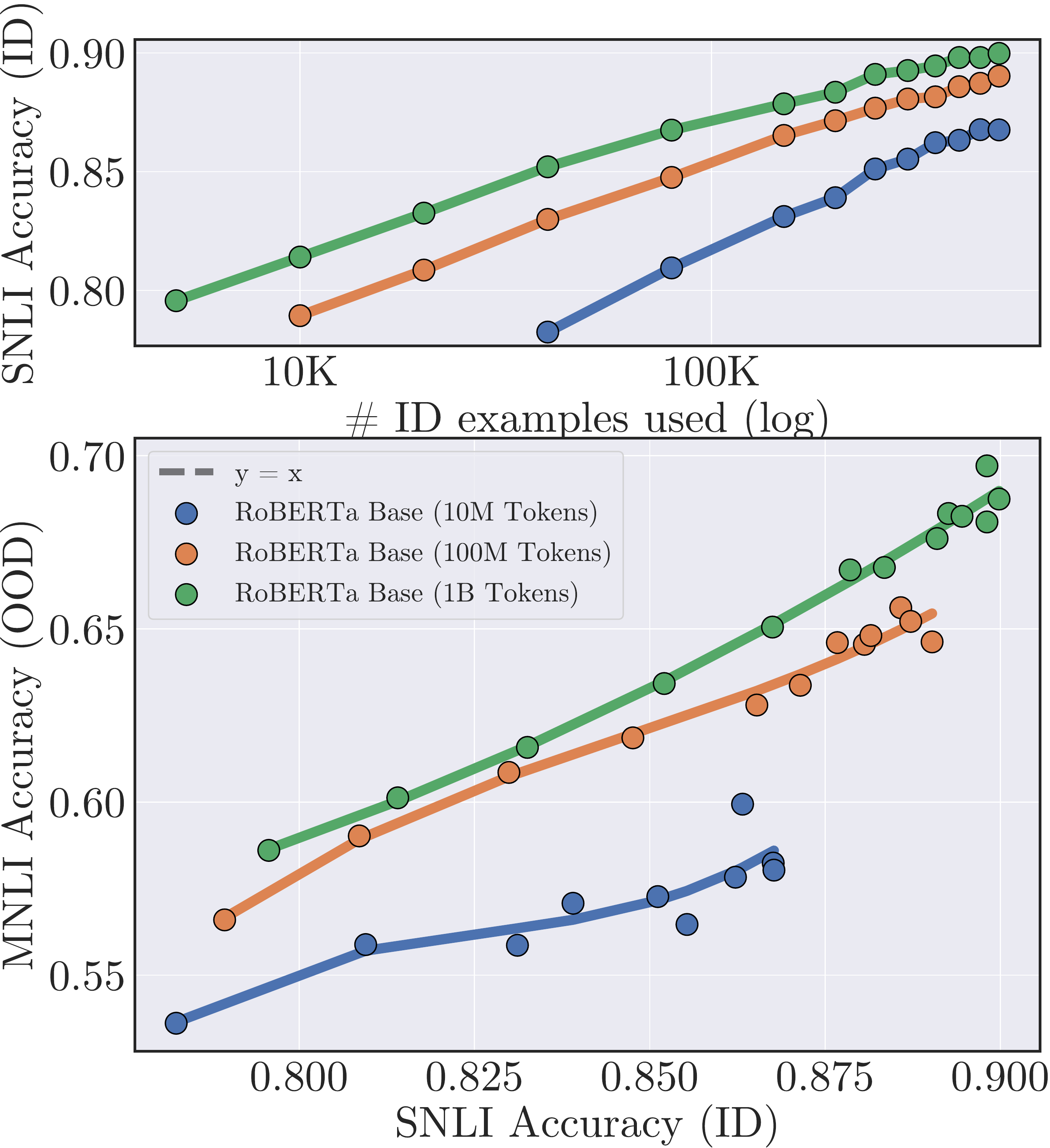}
            \caption{}
          \end{subfigure}
          \hfill
        \begin{subfigure}{0.32\linewidth}
            \centering
            \includegraphics[width=\linewidth]{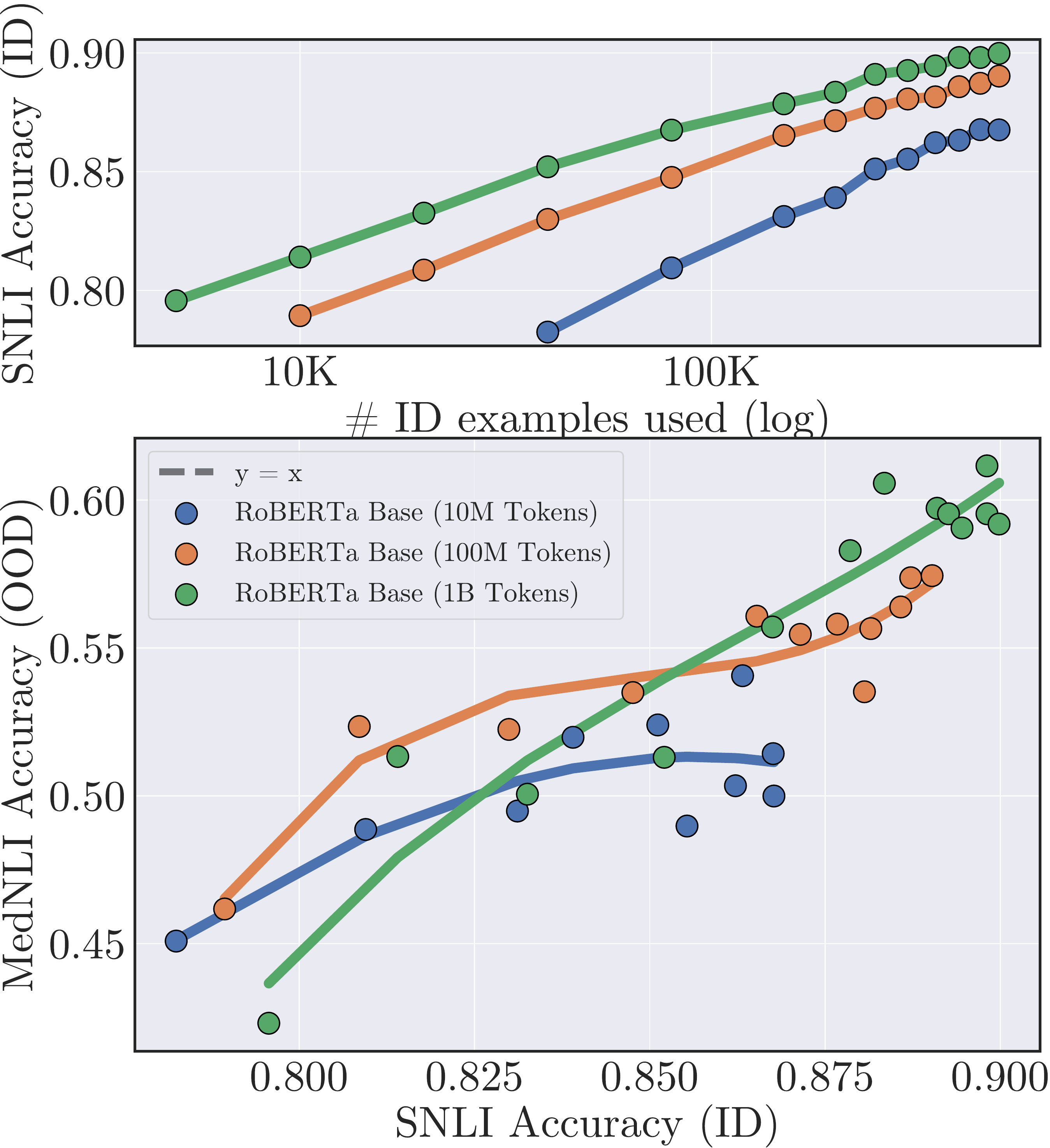}
            \caption{}
        \end{subfigure}
        \hfill
        \begin{subfigure}{0.32\linewidth}
            \centering
            \includegraphics[width=\linewidth]{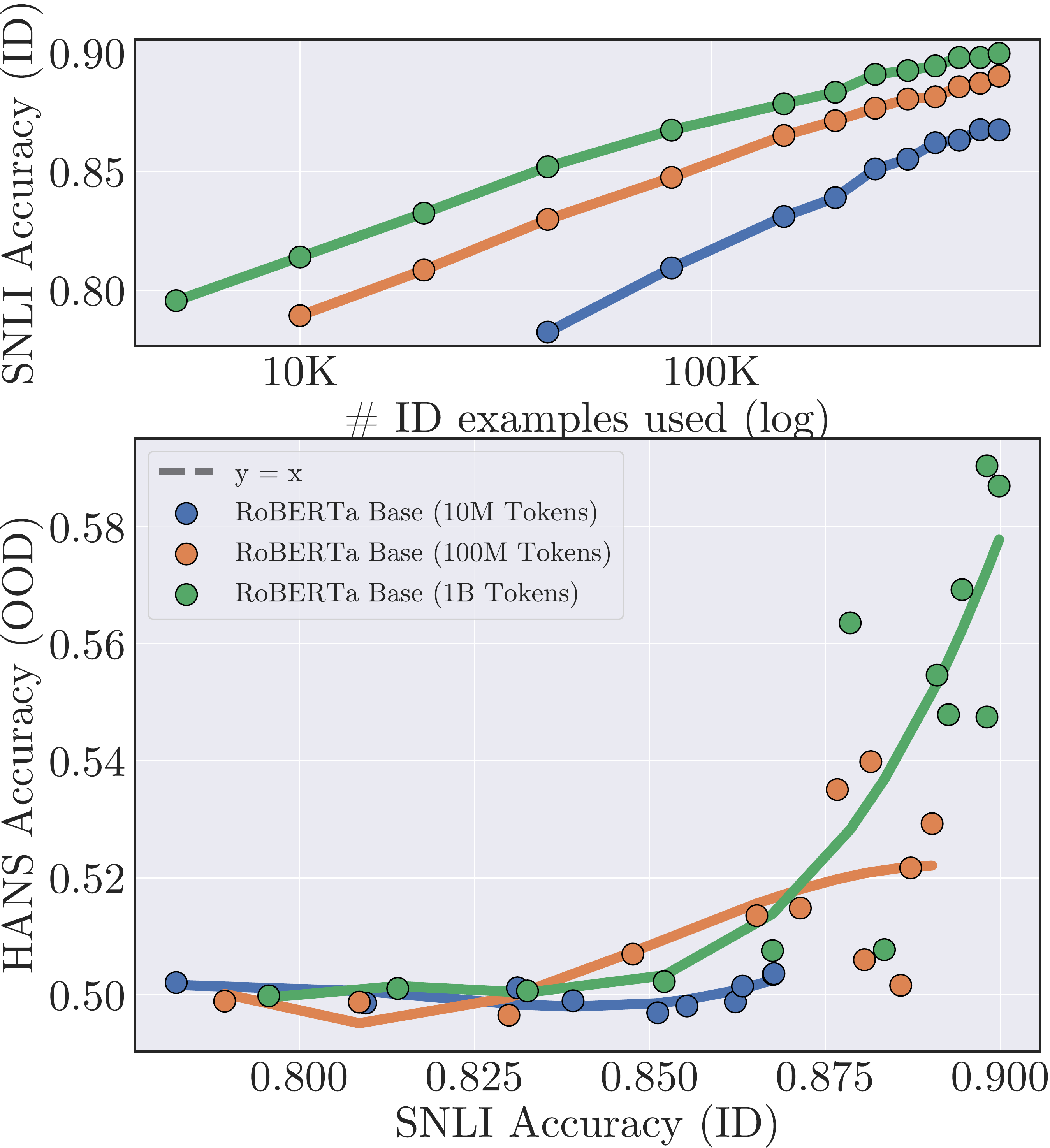}
            \caption{}
          \end{subfigure}
        \caption{Results on all NLI ID-OOD settings when increasing the amount of pre-training data.}
    \end{figure*}

\begin{figure*}[!h]
        \centering
        \begin{subfigure}{0.32\linewidth}
            \centering
            \includegraphics[width=\linewidth]{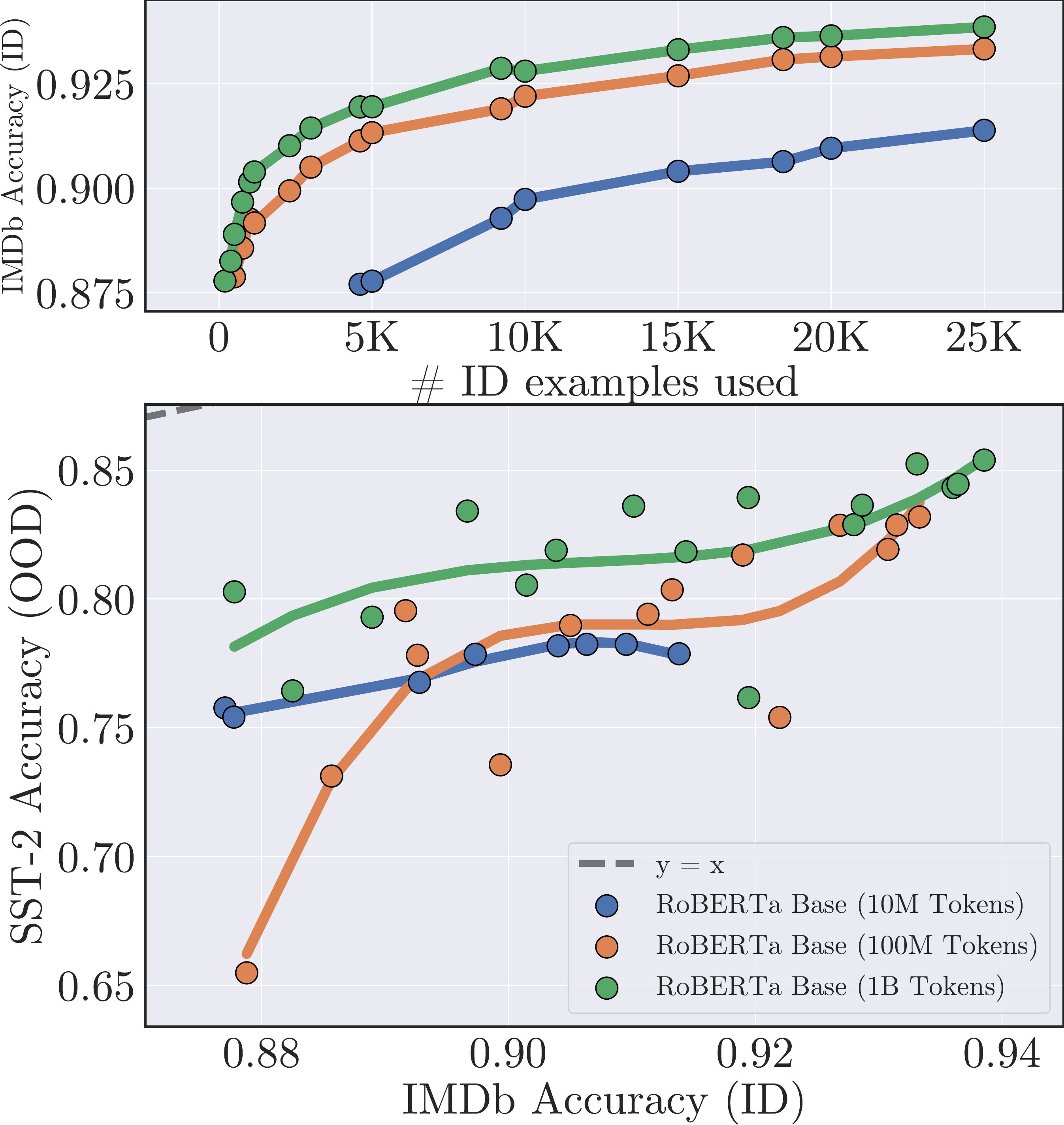}
            \caption{}
          \end{subfigure}
          \hfill
        \begin{subfigure}{0.32\linewidth}
            \centering
            \includegraphics[width=\linewidth]{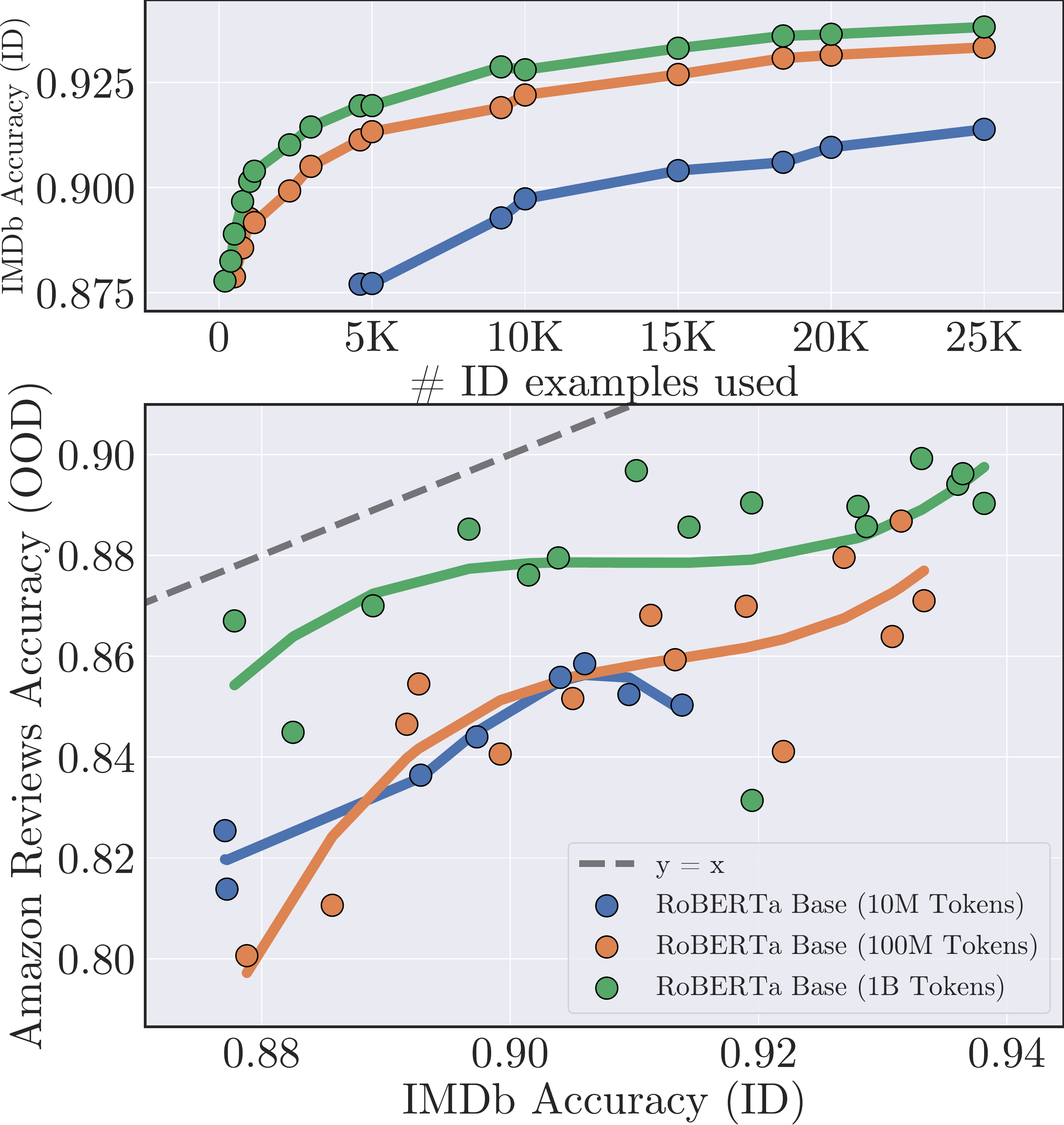}
            \caption{}
        \end{subfigure}
        \hfill
        \begin{subfigure}{0.32\linewidth}
            \centering
            \includegraphics[width=\linewidth]{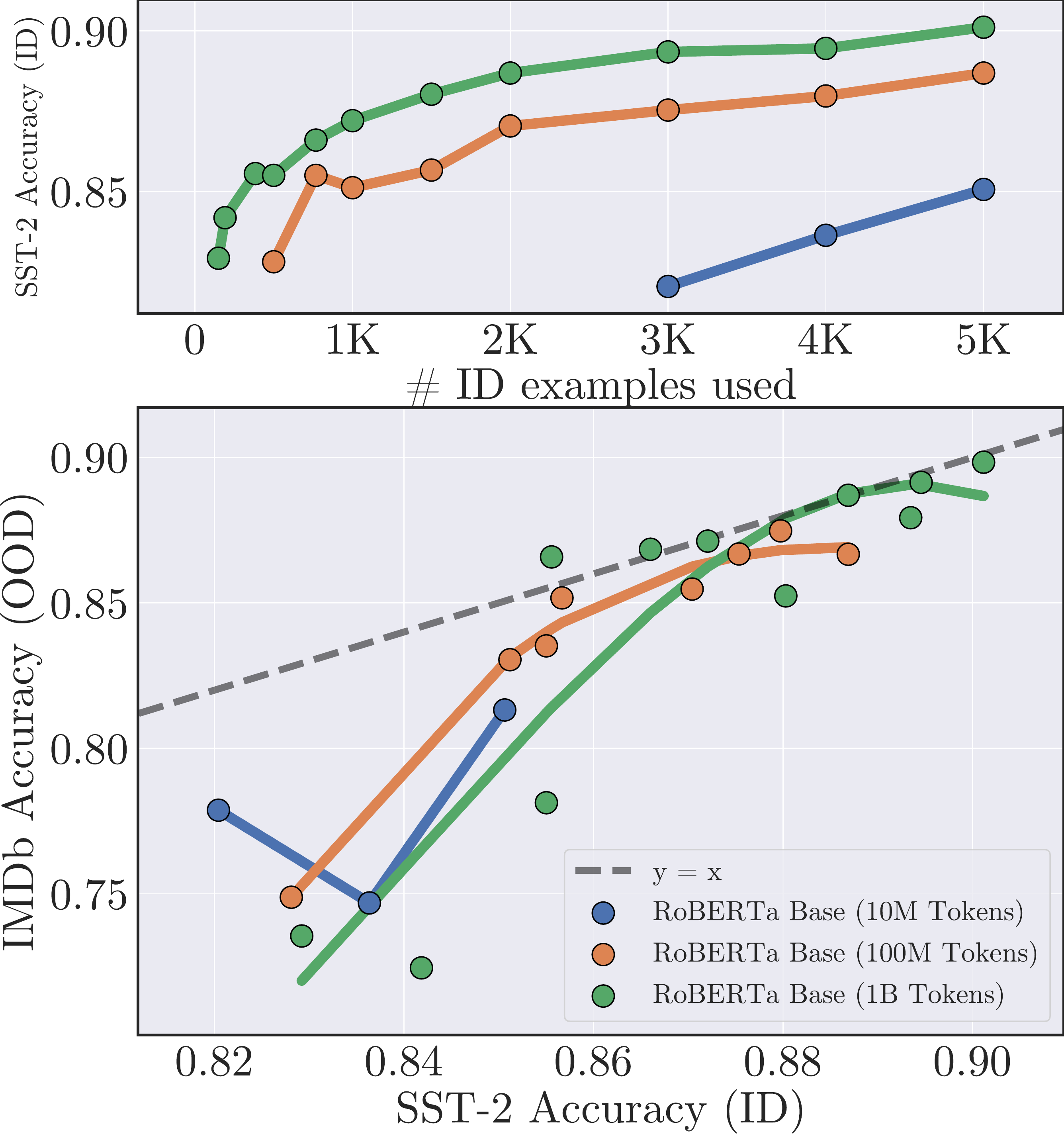}
            \caption{}
          \end{subfigure}
        \begin{subfigure}{0.32\linewidth}
            \centering
            \includegraphics[width=\linewidth]{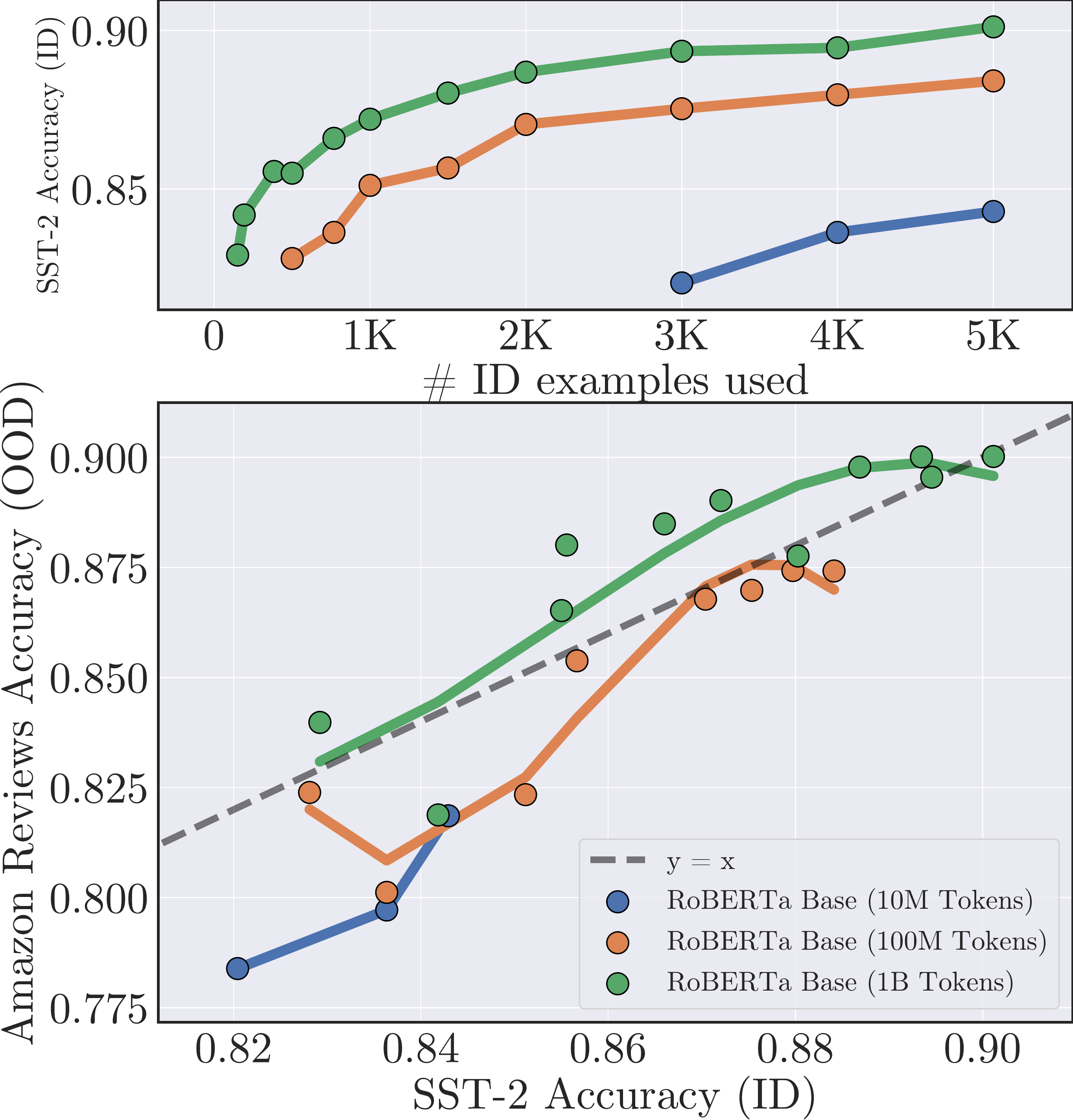}
            \caption{}
          \end{subfigure}
          \hspace*{\fill}
        \caption{Results on all sentiment analysis ID-OOD settings when increasing the amount of pre-training data.}
    \end{figure*}

  \begin{figure*}[!h]
        \centering
        \begin{subfigure}{0.32\linewidth}
            \centering
            \includegraphics[width=\linewidth]{figures/pretraining_data_squad_dev_vs_bioasq_train_dev_test.pdf}
            \caption{}
          \end{subfigure}
          \hfill
        \begin{subfigure}{0.32\linewidth}
            \centering
            \includegraphics[width=\linewidth]{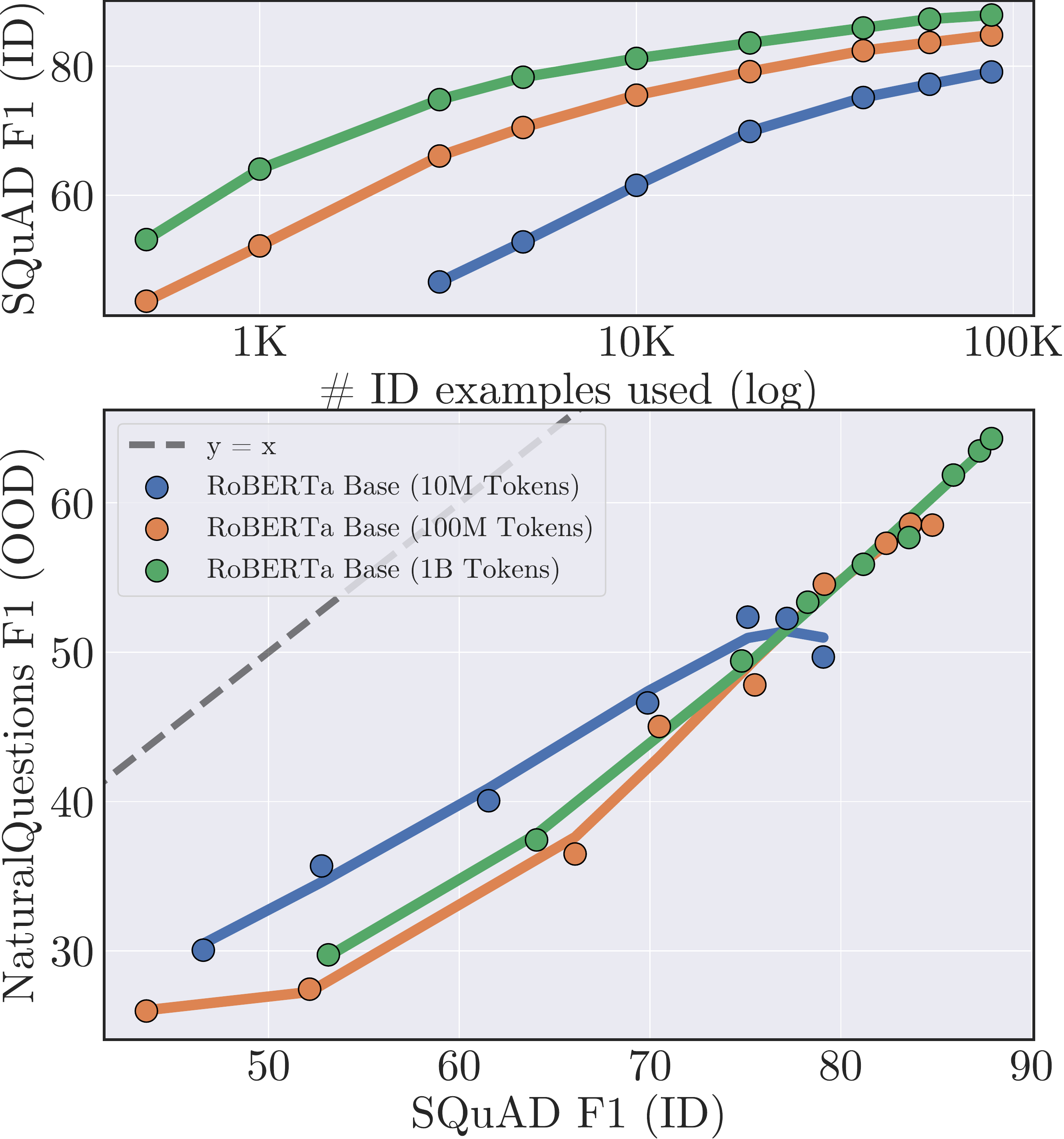}
            \caption{}
        \end{subfigure}
        \hfill
        \begin{subfigure}{0.32\linewidth}
            \centering
            \includegraphics[width=\linewidth]{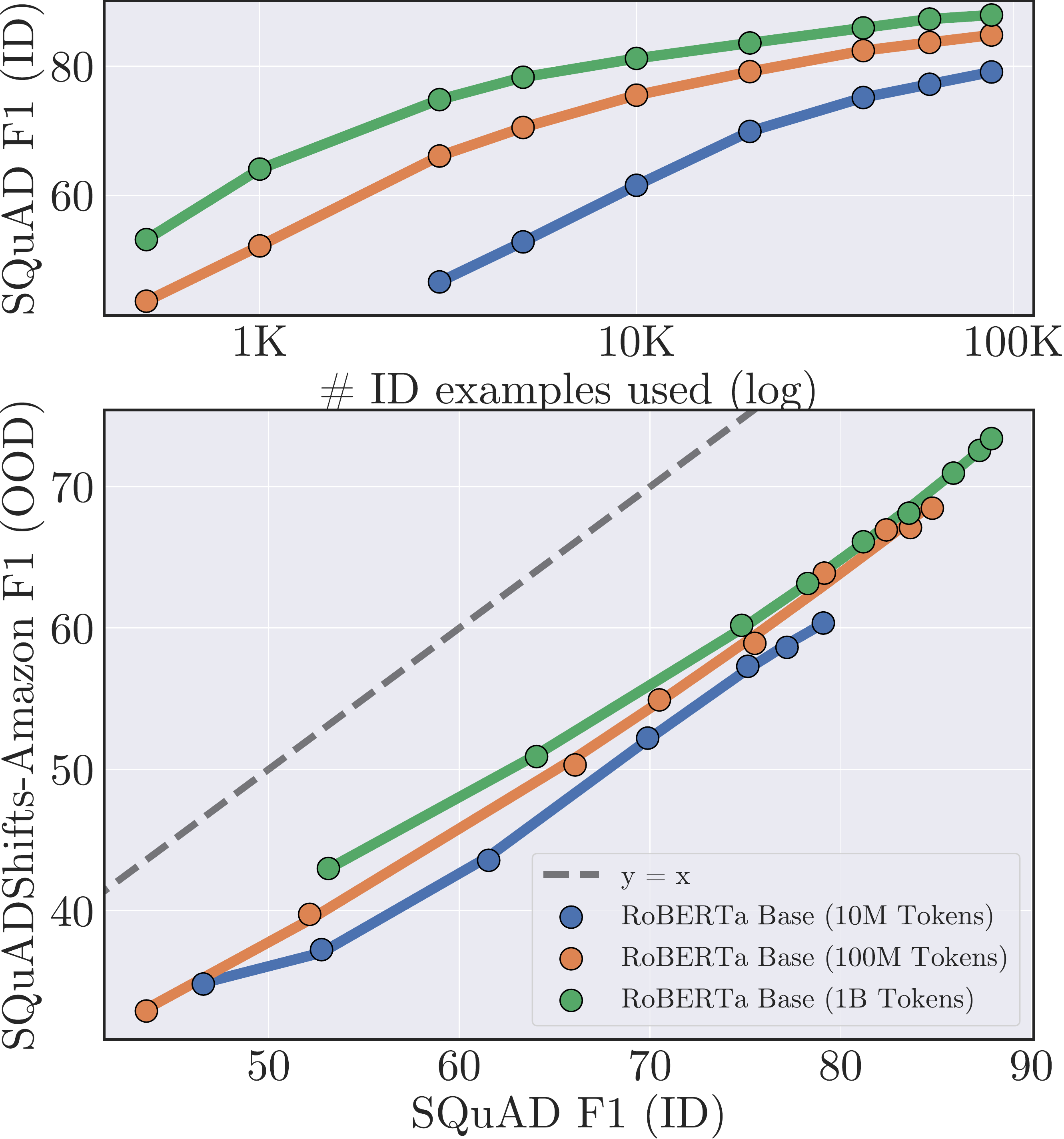}
            \caption{}
          \end{subfigure}
        \begin{subfigure}{0.32\linewidth}
            \centering
            \includegraphics[width=\linewidth]{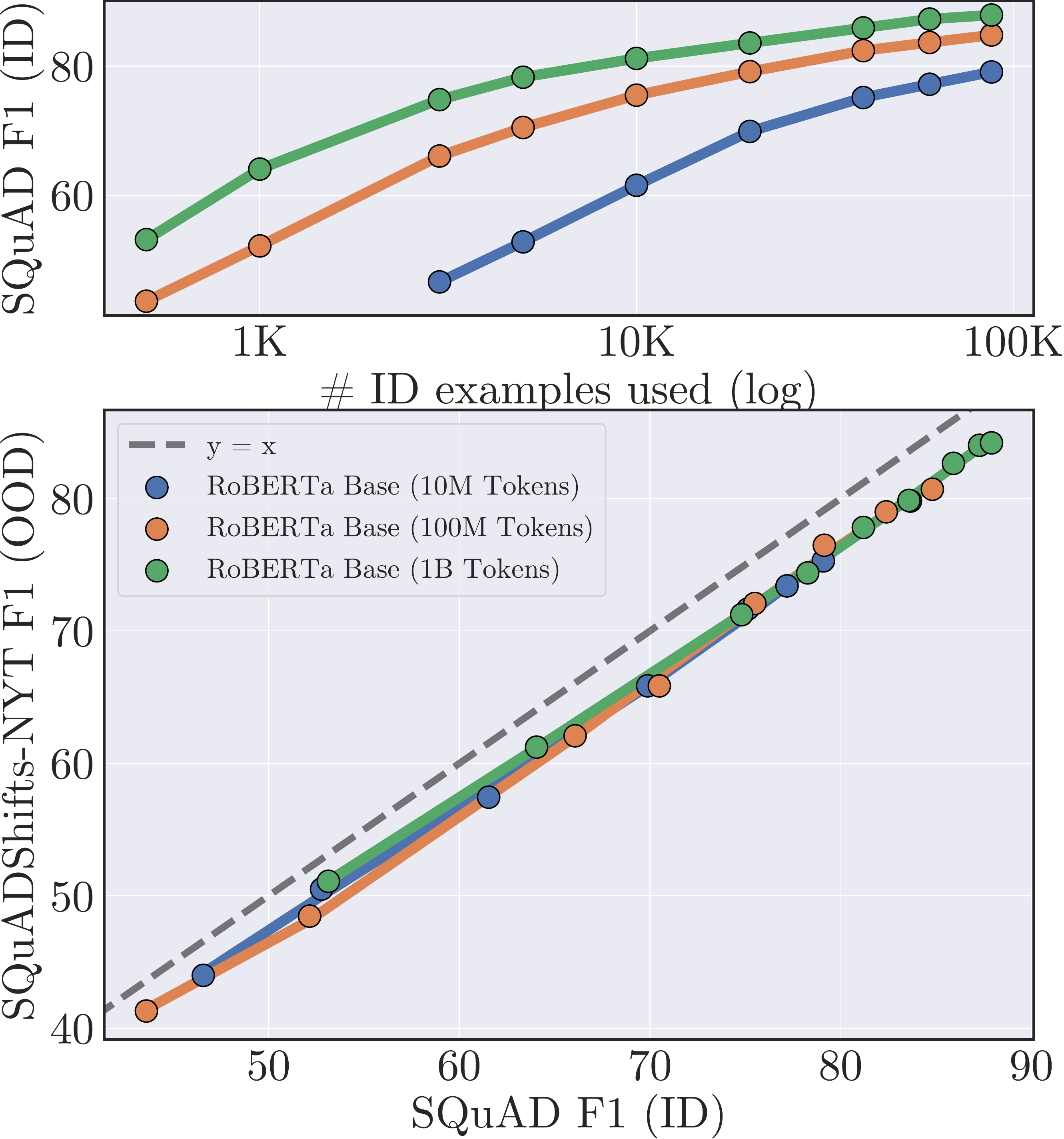}
            \caption{}
          \end{subfigure}
          \hfill
        \begin{subfigure}{0.32\linewidth}
            \centering
            \includegraphics[width=\linewidth]{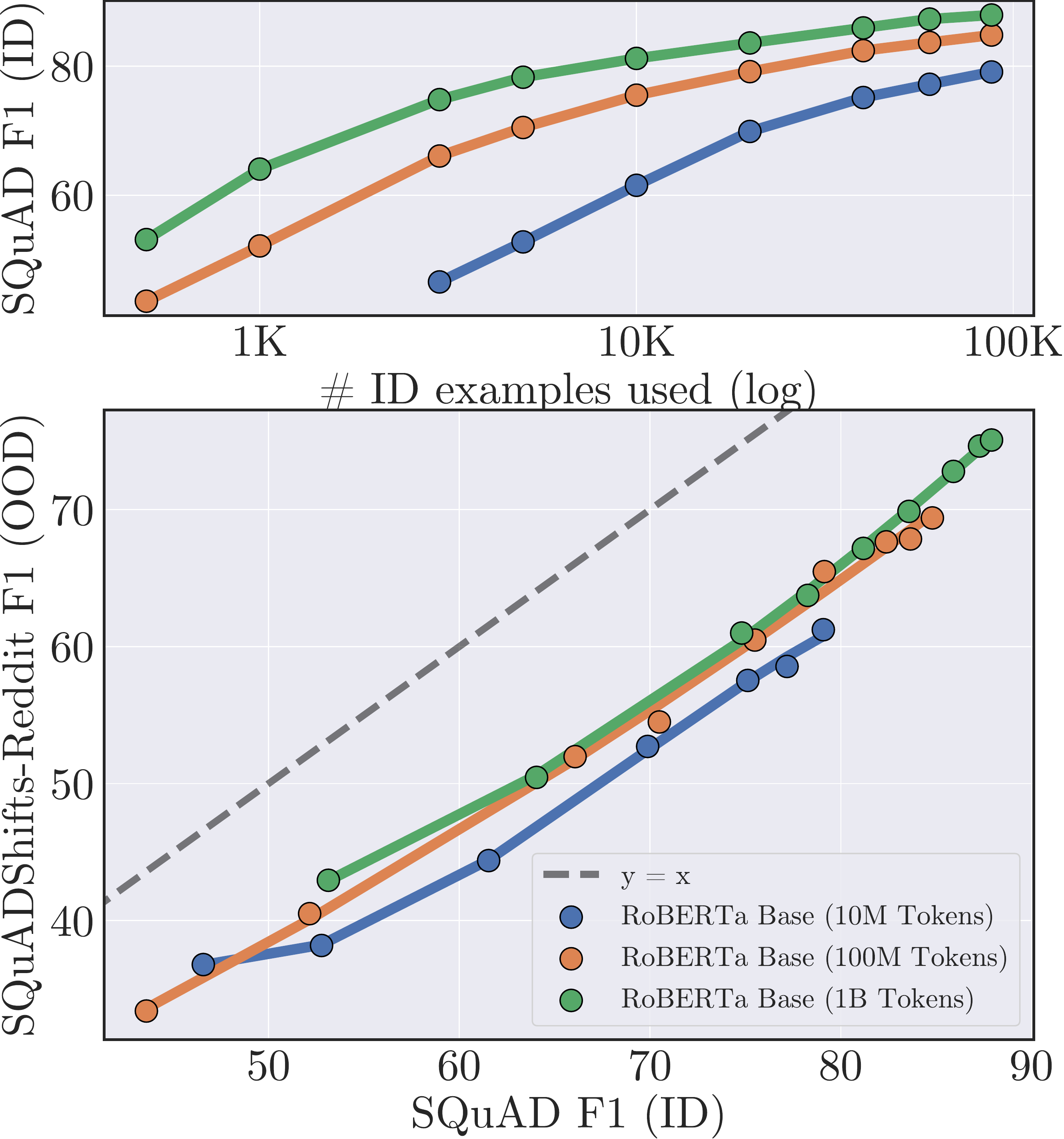}
            \caption{}
        \end{subfigure}
        \hfill
        \begin{subfigure}{0.32\linewidth}
            \centering
            \includegraphics[width=\linewidth]{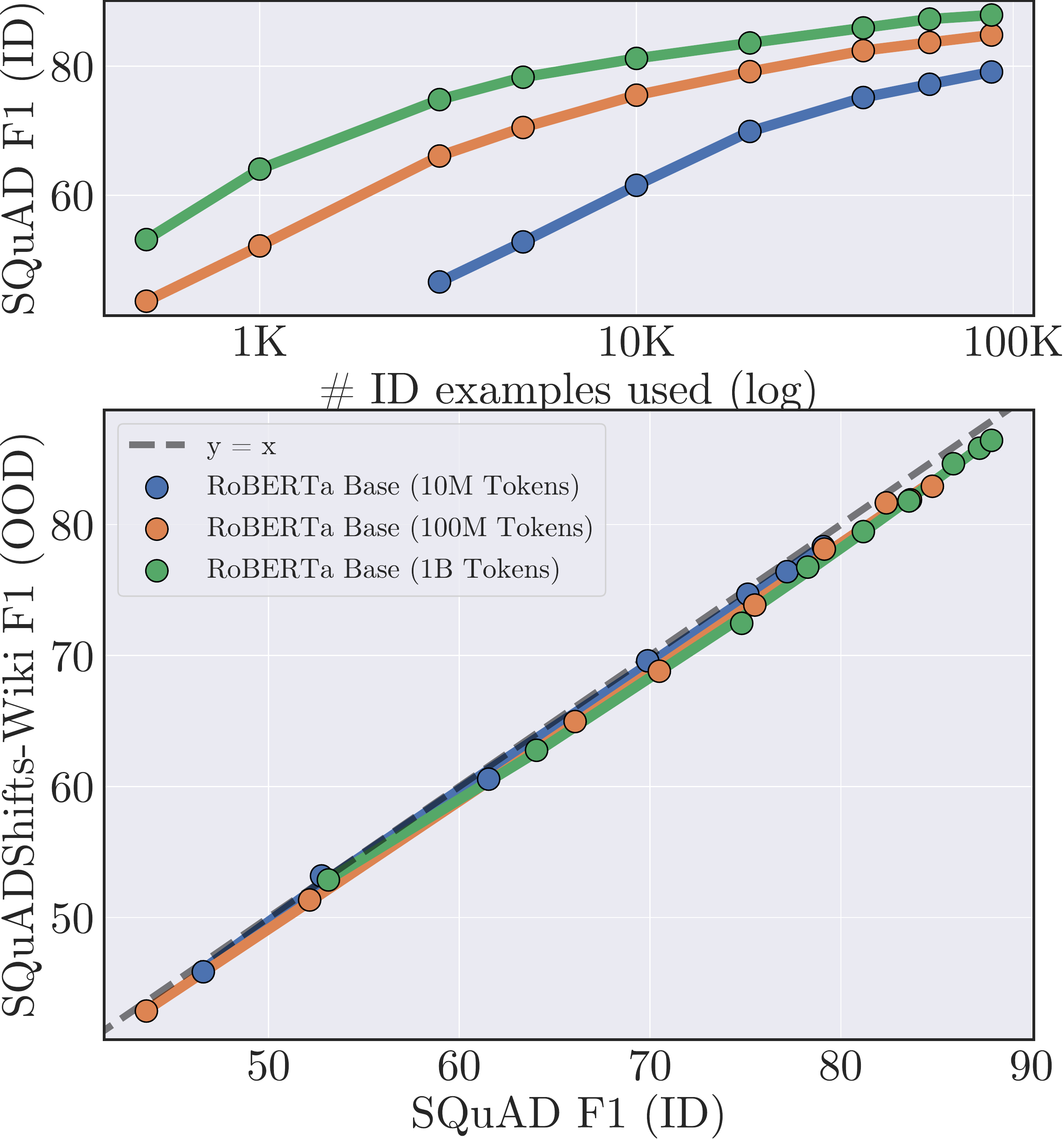}
            \caption{}
          \end{subfigure}
        \begin{subfigure}{0.32\linewidth}
            \centering
            \includegraphics[width=\linewidth]{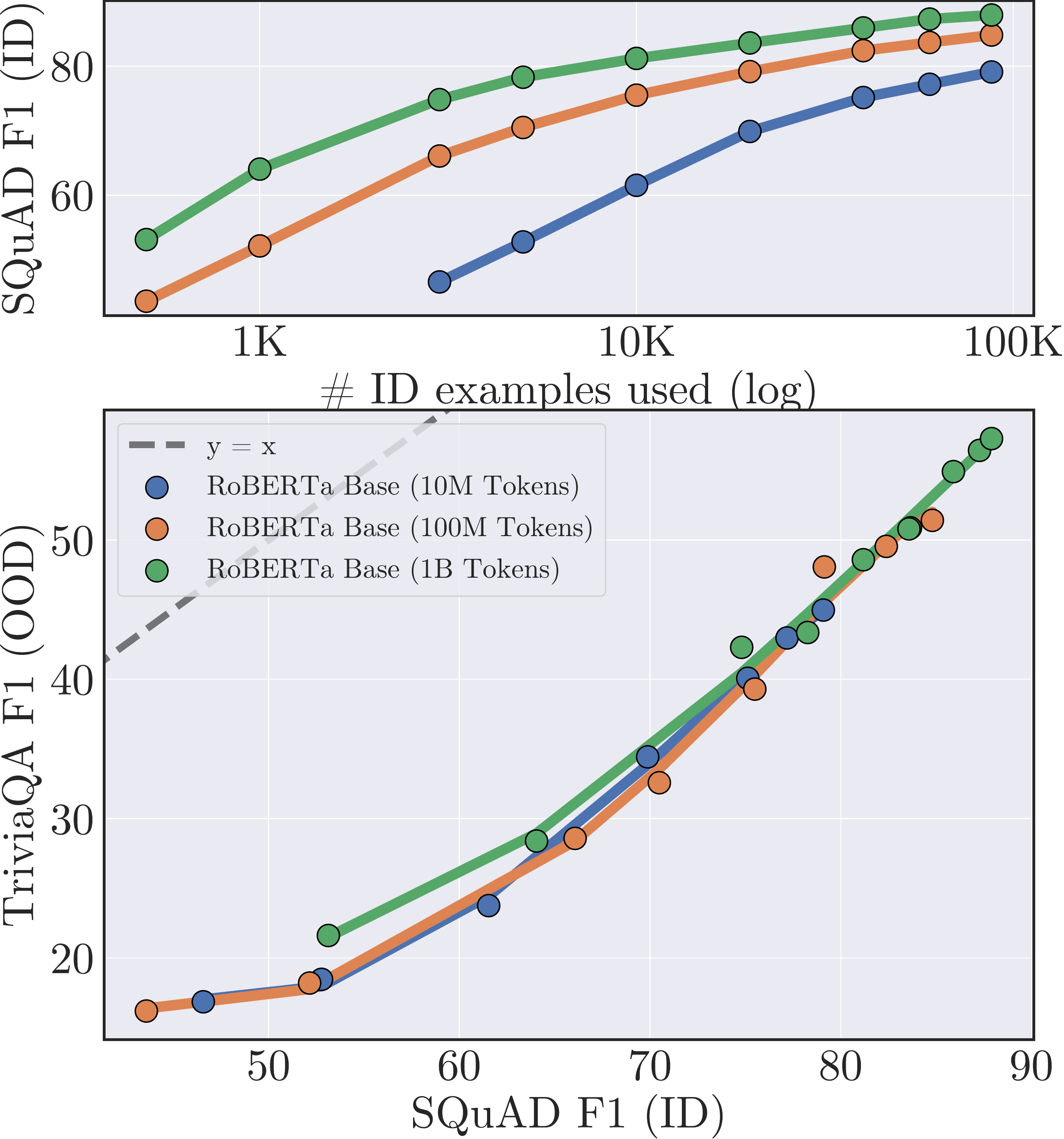}
            \caption{}
          \end{subfigure}
          \hspace*{\fill}
        \caption{Results on all extractive QA OOD settings when training on SQuAD with models pre-trained on varying amounts of data.}
    \end{figure*}

  \begin{figure*}[!h]
        \centering
        \begin{subfigure}{0.32\linewidth}
            \centering
            \includegraphics[width=\linewidth]{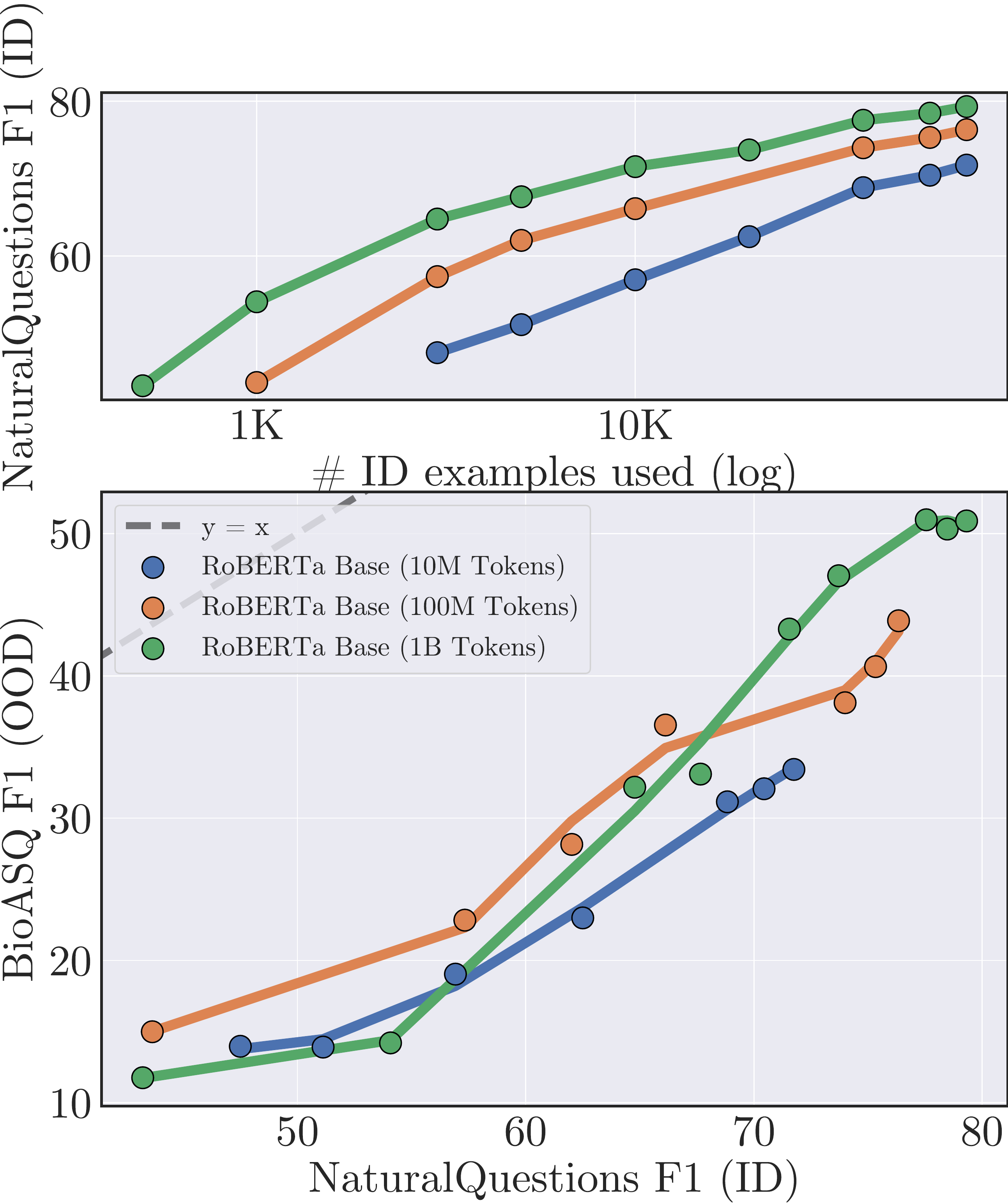}
            \caption{}
          \end{subfigure}
          \hfill
        \begin{subfigure}{0.32\linewidth}
            \centering
            \includegraphics[width=\linewidth]{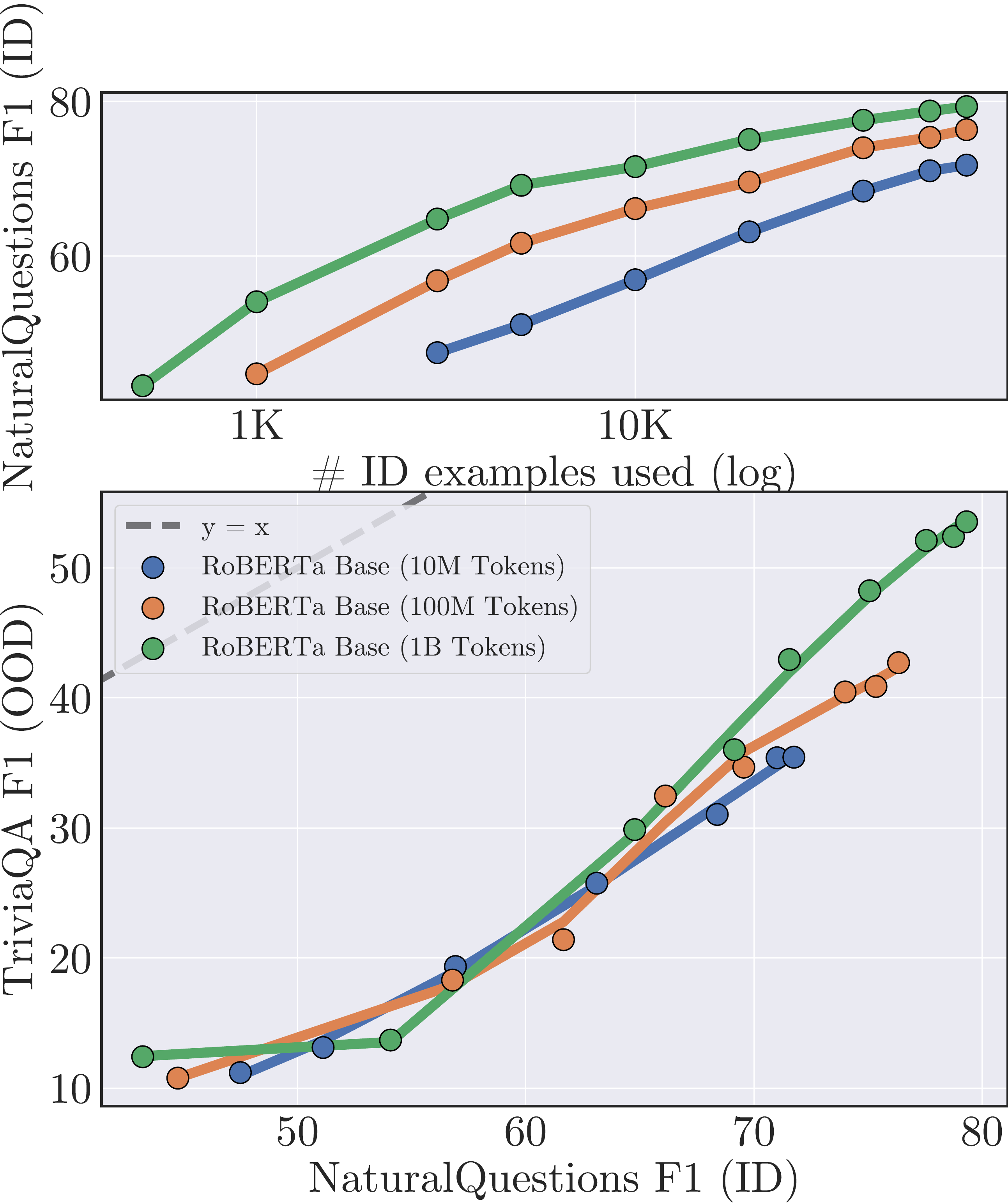}
            \caption{}
          \end{subfigure}
        \hfill
        \begin{subfigure}{0.32\linewidth}
            \centering
            \includegraphics[width=\linewidth]{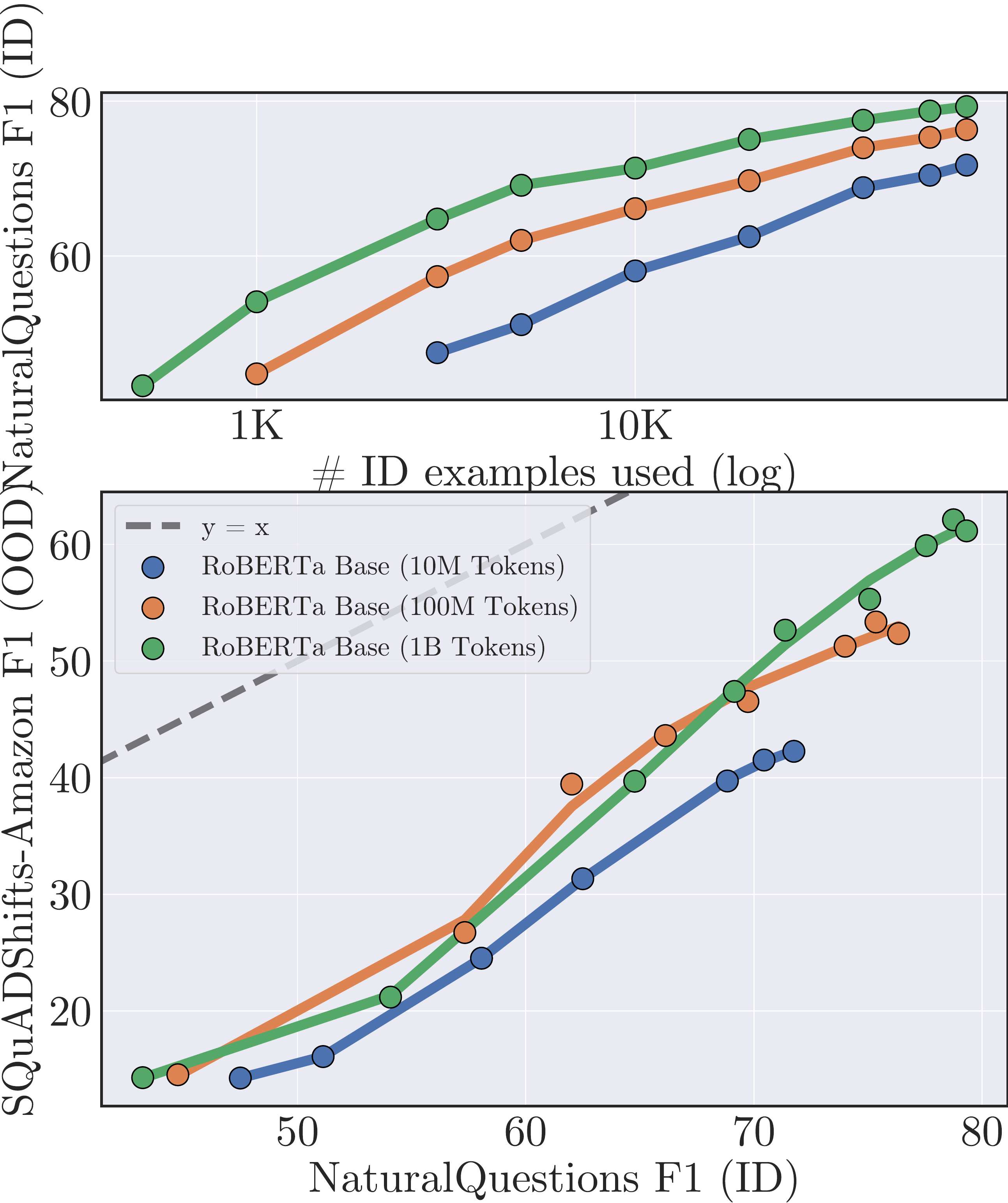}
            \caption{}
          \end{subfigure}
        \begin{subfigure}{0.32\linewidth}
            \centering
            \includegraphics[width=\linewidth]{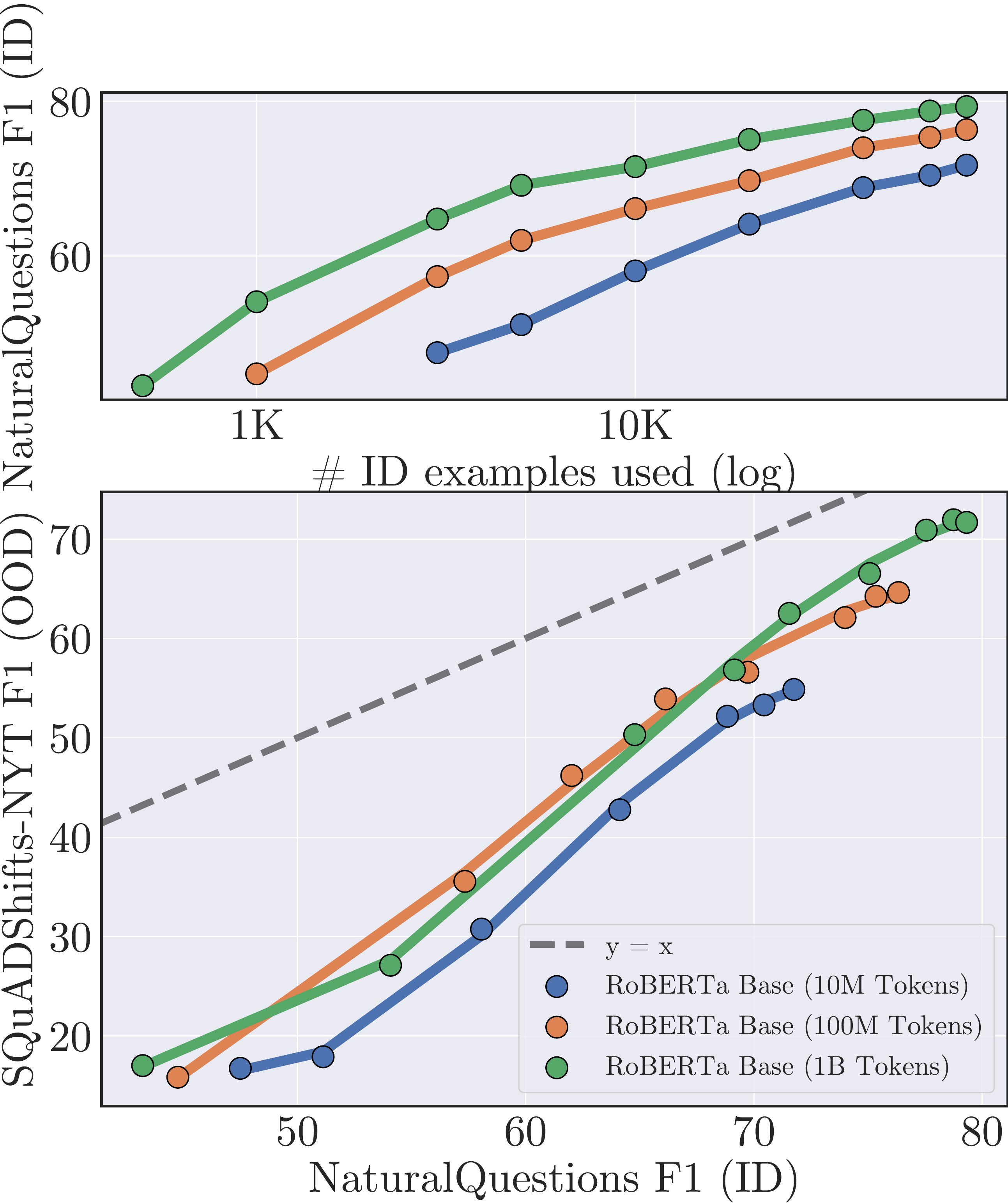}
            \caption{}
          \end{subfigure}
          \hfill
        \begin{subfigure}{0.32\linewidth}
            \centering
            \includegraphics[width=\linewidth]{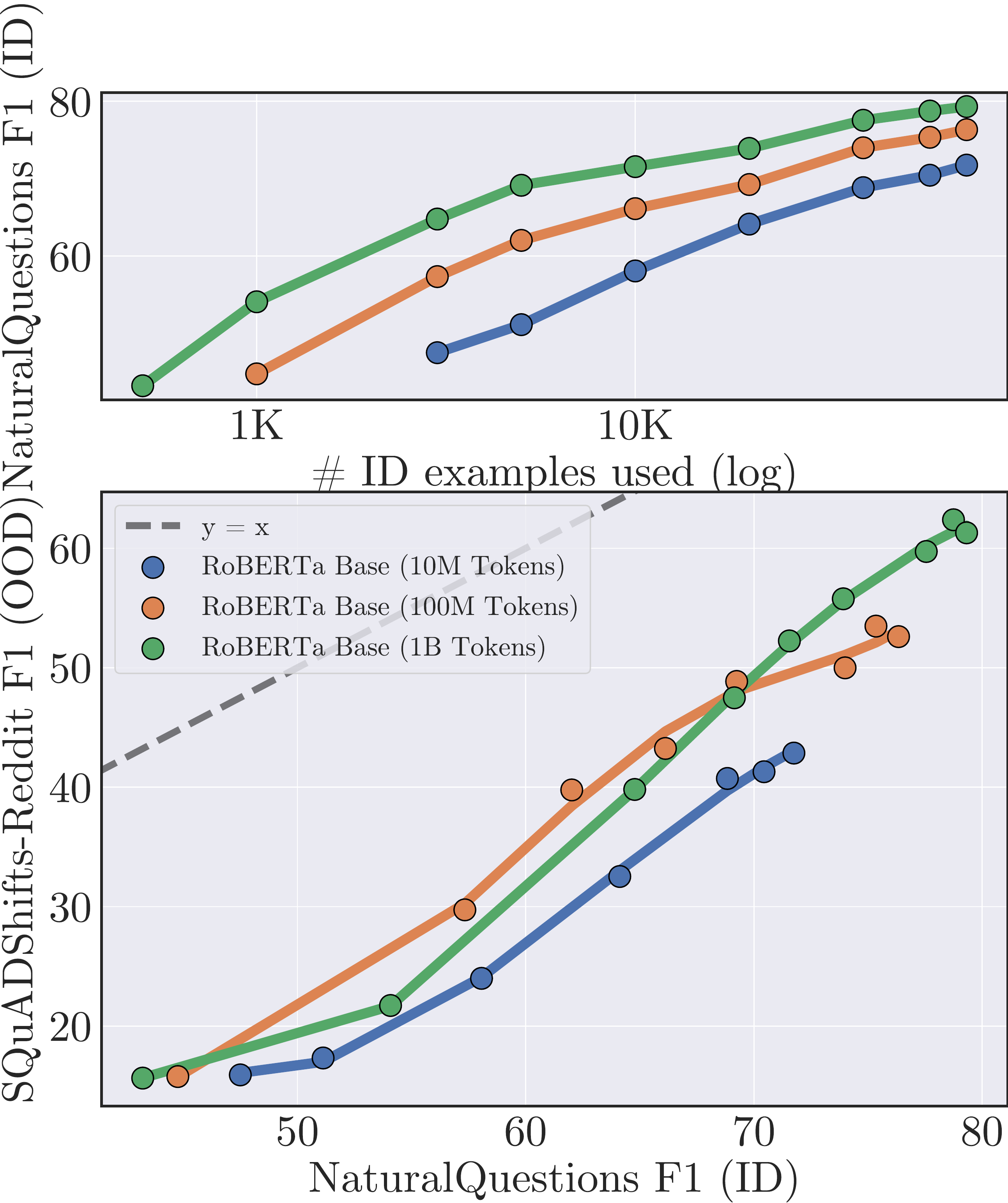}
            \caption{}
        \end{subfigure}
        \hfill
        \begin{subfigure}{0.32\linewidth}
            \centering
            \includegraphics[width=\linewidth]{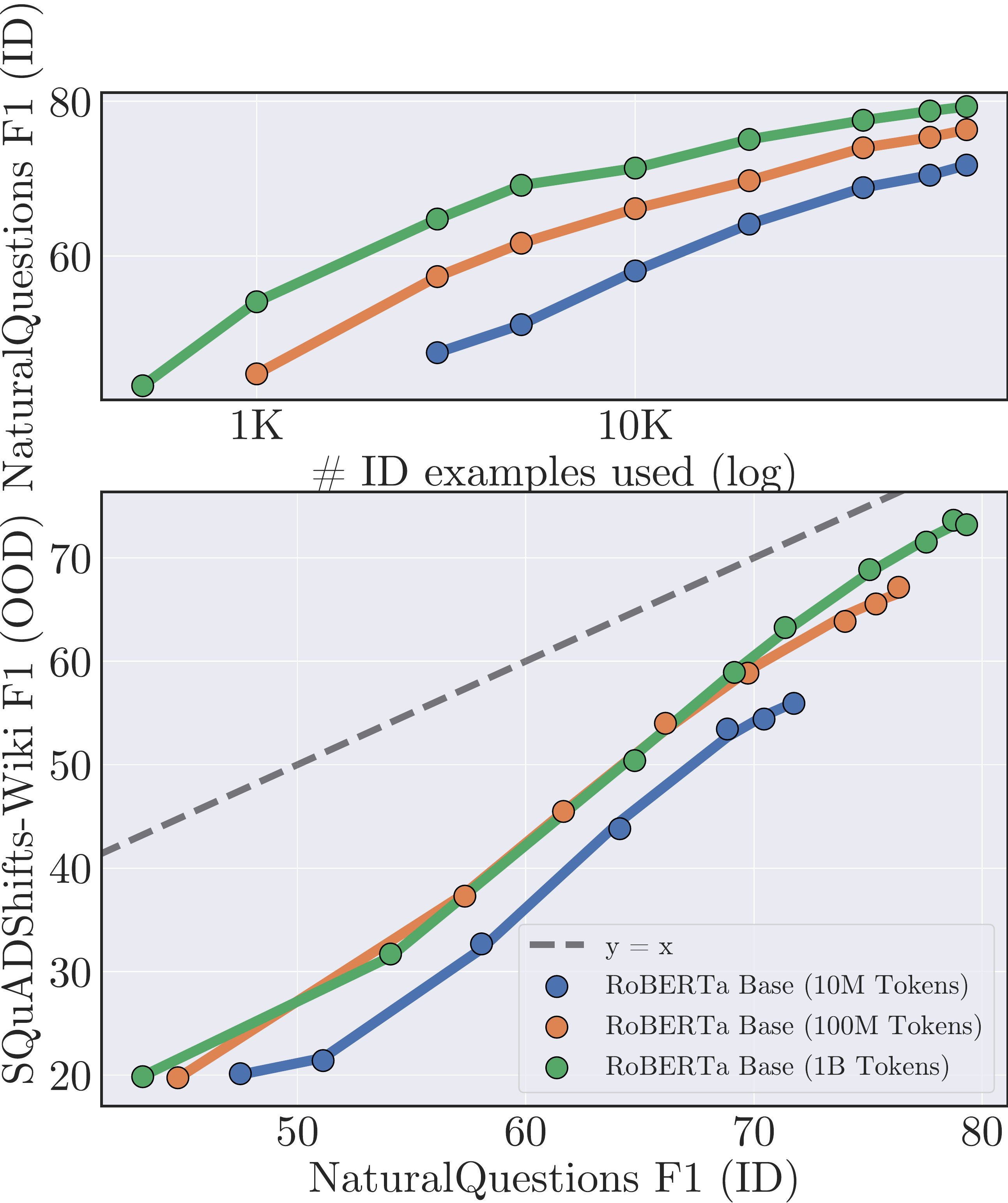}
            \caption{}
          \end{subfigure}
        \caption{Results on all extractive QA OOD settings when training on NaturalQuestions with models pre-trained on varying amounts of data.}
    \end{figure*}
\end{document}